\documentclass[preprint]{article}

\usepackage{aistats2027}
\usepackage{amsmath, amssymb, amsfonts, amsthm} 
\usepackage{mathtools, cases, empheq}
\usepackage{bm} 
\usepackage[utf8]{inputenc} 
\usepackage[T1]{fontenc}    
\usepackage{hyperref}       
\usepackage{url}            
\usepackage{booktabs}       
\usepackage{amsfonts}       
\usepackage{nicefrac}       
\usepackage{microtype}      
\usepackage[table]{xcolor} 
\usepackage{colortbl}
\definecolor{lightorange}{HTML}{FFE5CC} 
\definecolor{lightblue}{HTML}{E5F2FF}   
\usepackage{subcaption}
\usepackage{wrapfig}
\usepackage{multirow}
\usepackage{enumitem}
\usepackage{natbib}
\usepackage{rotating}
\usepackage{cleveref}

\newtheorem{lemma}{Lemma}
\newtheorem{prop}{Proposition}
\newtheorem{thrm}{Theorem}
\newtheorem{corollary}{Corollary}
\newtheorem{remark}{Remark}
\newtheorem{definition}{Definition}
\usepackage{algorithm, algpseudocodex, amsmath, bm}

\usepackage{booktabs}
\usepackage{graphicx}
\usepackage[table]{xcolor}
\definecolor{lightblue}{RGB}{219,234,254}
\newcommand{\ms}[2]{#1\,{\color{gray}\scriptsize$\pm$#2}}

\makeatletter
\renewcommand\paragraph{\@startsection{paragraph}{4}{\z@}%
 {0ex \@plus1ex \@minus.2ex}%
 {-1em}%
 {\normalfont\normalsize\bfseries}}
\makeatother

\crefname{prop}{Proposition}{Propositions}
\Crefname{prop}{Proposition}{Propositions}

\begin{document}

%

%
\runningauthor{Koshi Watanabe, Daichi Mochihashi, Hideitsu Hino, Takahiro Ogawa, Miki Haseyama}

\twocolumn[

\aistatstitle{
Helgason--Fourier Random Features \\ for Scalable Hyperbolic Latent Variables
}
\aistatsauthor{ Koshi Watanabe \\ \normalfont Hokkaido University
  \And Daichi Mochihashi \\ \normalfont The Institute of Statistical Mathematics
  \AND Hideitsu Hino \\ \normalfont The Institute of Statistical Mathematics \\ \normalfont Waseda University
  \And Takahiro Ogawa \\ \normalfont Hokkaido University
  \And Miki Haseyama \\ \normalfont Hokkaido University }
\aistatsaddress{} ]

\begin{abstract}
    Hyperbolic geometry is well suited to embedding hierarchical data, capturing tree-like structure faithfully even in the low dimensions where Euclidean dimensionality reduction is severely constrained.
    We introduce Random Helgason--Fourier Features (RHFF), an explicit finite-dimensional feature map for kernels on hyperbolic space.
    We characterize the spectral density from which RHFF frequencies are drawn, allowing kernels such as the Mat\'ern family to be specified directly in the frequency domain.
    We also prove a uniform approximation error bound for RHFF.
    Building on this representation, we develop a hyperbolic latent variable model
    called RHFF-LVM, which requires no numerical quadrature, variational bounds or auxiliary variables unlike previous hyperbolic GP-LVMs.
    Our RHFF-LVM recovers the metric of a synthetic tree more faithfully than all baselines and attains the best mean rank, driven by large gains on tree-structured atlases at $Q\!=\!2$.
    Furthermore, it fits $5\times10^{5}$ points in $505$ seconds, making it the first non-parametric hyperbolic LVM to reach this scale.
\end{abstract}

\section{Introduction}
\label{Section:Introduction}
Dimensionality reduction is widely used in data analysis and as a preprocessing step for downstream tasks.
However, compression to very low dimensions (e.g., for visualization with t-SNE~\cite{van2008visualizing}) is inherently lossy, as no embedding can retain all the structure of the original data. 
Every method therefore makes an implicit choice: preserve \emph{local} relations, which reside in the neighborhood of each point \cite{mcinnes2018umap,van2008visualizing}; preserve \emph{global} relations, which reside in pairwise distances \cite{kruskal1964multidimensional,lawrence2005probabilistic,titsias2010bayesian}; or balance the two \cite{kury2025dreams,moon2019visualizing,wang2021understanding}.
Every choice sacrifices something.
\par
Altering the geometry of the low-dimensional embedding space mitigates the inherent limitations of dimensionality reduction.
For hierarchical data, hyperbolic space provides far more capacity than Euclidean space.
The volume of a ball in hyperbolic space grows exponentially with its radius, matching the exponential branching inherent in hierarchies~\cite{cannon1997hyperbolic,nickel2017poincare,nickel2018learning}.
Indeed, trees can be embedded into the hyperbolic plane with arbitrarily low distortion~\cite{sala2018representation,sarkar2011low}, a guarantee that no finite-dimensional Euclidean space offers.
\par
In hierarchical structures, information regarding the branching of edges and the depth at which each point is located is of critical importance.
As epitomized by the reconstruction of phylogenetic trees from pairwise distance matrices~\cite{saitou1987neighbor}, this information is carried primarily by the full matrix of pairwise distances rather than by local neighborhoods alone.
However, hyperbolic dimensionality reduction at scale has so far been realized only by neighbor embeddings~\cite{skrodzki2024accelerating}, which leave precisely the global structure outside their guarantees.
\par
The Gaussian process latent variable model (GP-LVM)~\cite{lalchand2022generalised,lawrence2005probabilistic,titsias2010bayesian} is particularly well suited to dimensionality reduction that preserves global structure, as it matches the Gram matrix induced by the kernel on the latent points to that of the observations.
Unlike stress-based embeddings~\cite{kruskal1964multidimensional,sala2018representation}, this generative formulation models the observations themselves and treats noise explicitly.
Yet, hyperbolic GP-LVM has remained impractical for three reasons.
First, practically relevant hyperbolic kernels (e.g., the Mat\'ern family) lack closed forms, so each Gram entry must itself be approximated numerically~\cite{azangulov2024stationary,jaquier2024bringing}, incurring a heavy computational burden. 
Second, the variational objective of the sparse GP-LVM~\cite{lalchand2022generalised,titsias2010bayesian} is a lower bound of the likelihood whose tightness depends on auxiliary variables optimized jointly with the latents, an interaction known to destabilize training~\cite{li2024preventing,moreno2022revisiting}.
Indeed, the existing hyperbolic GP-LVM fixes its inducing inputs rather than optimizing them~\cite{pmlr-v258-watanabe25a}.
The third difficulty stems from the geometry itself: Riemannian updates of the latent variables accumulate floating-point error that grows with distance from the origin, destabilizing training regardless of the model~\cite{mishne2023numerical,nickel2018learning,yu2019numerically}.
\par
\begin{figure}[t]
    \centering
    \includegraphics[width=\linewidth]{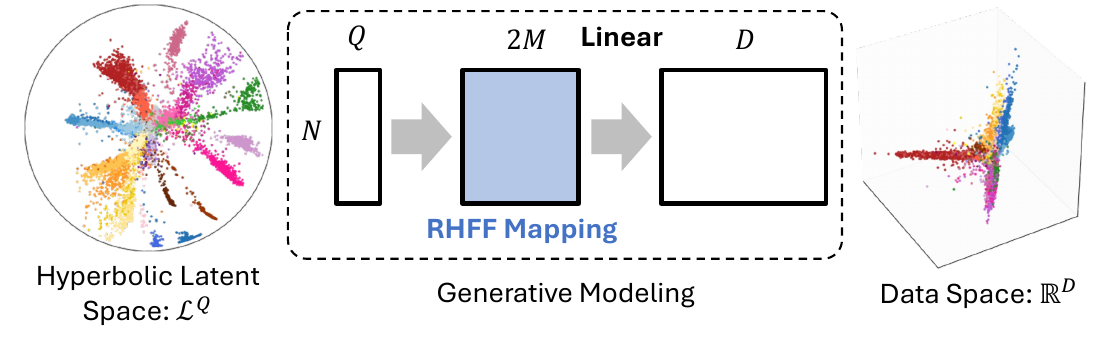}
    \vspace{-1.5em}
    \caption{Overview of our method. $Q$-dimensional latent variables in hyperbolic space $\mathcal{L}^{Q}$ are mapped by RHFF to explicit $2M$-dimensional closed-form features (\Cref{Eq:RHFF-Mapping}). Then, the observations follow a Bayesian linear model whose marginal likelihood is evaluated exactly in $\mathcal{O}(NM^{2}\!+\!M^{3})$ time without inducing inputs.}
    \label{Fig:Abstract}
    \vspace{-1em}
\end{figure}
We make the hyperbolic GP-LVM practical through a latent variable model built on a single choice of representation: Random Helgason--Fourier Features (RHFF; \Cref{Fig:Abstract}).
RHFF approximates hyperbolic kernels, including the Mat\'ern family, by an explicit inner product of $2M$-dimensional features derived from the Helgason--Fourier transform~\cite{helgason2022groups}. 
This choice resolves each difficulty in turn.
First, the features are closed-form expressions in terms of \textit{Busemann functions}, eliminating numerical kernel evaluation and reducing computational cost. 
Second, by modeling the observations as generated from RHFF projections of the latents, our objective eliminates the need for variational bounds and inducing inputs~\cite{gundersen2021latent,li2024preventing}, and stabilizes the optimization at $\mathcal{O}(NM^{2}+M^{3})$ time complexity. 
Third, we reparameterize the latents as unconstrained Euclidean variables and optimize them by natural gradient descent, eliminating projection and retraction errors and thereby reducing the accumulation of floating-point errors. 
\par
Theoretically, we prove a uniform bound on the error with which RHFF approximates the target hyperbolic kernel (Proposition~\ref{Prop:Error-Bound}).
The same analysis reveals that the approximation error grows exponentially with the latent dimension $Q$ and the diameter $R$ of the latent domain, a practical limit that we verify empirically.
We further show that HyLa~\cite{yu2023random}, the existing hyperbolic random feature, is a special case of RHFF that implicitly fixes one particular spectrum.
\par
Empirically, RHFF embeds hierarchical structure faithfully.
On a synthetic tree, its distance correlation with the true tree metric reaches $0.843$, versus $0.749$ for its Euclidean counterpart and $0.824$ for the strongest hyperbolic baseline.
On eight real-world datasets, including single-cell atlases of up to $292$k cells, RHFF-LVM (heat) attains the best mean rank on global-structure metrics at $Q=2$ ($2.55$ vs.\ $3.35$ for PCA, a strong linear reference for distance preservation).
The choice of spectral density further controls the balance between global and local structure: with the heavy-tailed Mat\'ern spectrum ($\nu=0.5$), RHFF achieves the best combined rank under equal weighting of the two criteria ($4.08$).
Moreover, it fits $N=5\times10^{5}$ points in $505$ seconds, whereas existing hyperbolic latent variable models become infeasible beyond $10^{4}$--$2\times10^{4}$ points.

\section{Related Work}
\label{Section:Related-Work}

\paragraph{Gaussian process latent variable models.}
The GP-LVM~\cite{lawrence2005probabilistic,titsias2010bayesian,lalchand2022generalised} with a distance-based kernel matches the covariance induced by the kernel on latent distances to that of the observations---a mechanism known as \emph{dissimilarity preservation}~\cite{lawrence2005probabilistic,lawrence2006local}.
In Euclidean space, random-feature LVMs~\cite{gundersen2021latent,li2024preventing} improve robustness precisely because no auxiliary variables are optimized jointly with the latents.
We extend this line of work to hyperbolic
space. 
However, translation invariance is lost, and Bochner's theorem no longer applies, which requires the Helgason--Fourier machinery, Busemann-based features, and a covering analysis on the product space (\S~\ref{Section:Error-Bound}).
The optimization further requires a retraction-free pullback formulation (\S~\ref{Section:RHFF-LVM}).

\paragraph{Hyperbolic dimensionality reduction.}
Hyperbolic neighbor embeddings~\cite{klimovskaia2020poincare,skrodzki2024accelerating} optimize local objectives and leave global distances outside their guarantees.
Hyperbolic VAEs~\cite{cho2022rotated,nagano2019wrapped} are likelihood-based but parametric, so the latents inherit the encoder's inductive bias rather than being fit per observation.
Hyperbolic GP-LVMs~\cite{jaquier2024bringing,pmlr-v258-watanabe25a} are non-parametric and free of this constraint, but also rely on variational objectives whose joint optimization over auxiliary variables and latents is unstable.
We instead maximize an exact likelihood with an approximate kernel, which stabilizes latent-variable optimization.
Stress-based hyperbolic MDS~\cite{sala2018representation} targets global distances but requires a reliable precomputed distance matrix.
HoroPCA~\cite{chami2021horopca} and Co-SNE~\cite{guo2022co} address dimensionality reduction \emph{within} hyperbolic space, whereas we embed Euclidean observations \emph{into} hyperbolic space.

\paragraph{Hyperbolic kernels and random features.}
The spectral theory of kernels on non-compact symmetric spaces~\cite{azangulov2024stationary} provides the spectral densities from which we sample.
Our contribution is to turn these densities into explicit finite-dimensional features with a uniform approximation guarantee.
HyLa~\cite{yu2023random} is closest to RHFF, but it draws its frequencies from a normal distribution, whereas RHFF admits any spectrum, including those of the heat and Mat\'ern kernels.
HyLa is thus a special case of our construction (\S~\ref{Subsec:RHFF}).

\paragraph{Numerical stability of hyperbolic learning.}
The numerical fragility of hyperbolic training is well documented.
Optimization instabilities in the Poincar\'e ball motivated the move to the Lorentz model~\cite{nickel2018learning}.
Floating-point error grows without bound with distance from the origin~\cite{yu2019numerically}, and the precision inherently required by hyperbolic representations has been characterized~\cite{sala2018representation}.
Proposed remedies include tiling-based and multi-component float representations~\cite{yu2019numerically,yu2021representing} and Euclidean parameterizations of hyperbolic space~\cite{mishne2023numerical}.
Our formulation follows the latter insight and additionally equips the parameterization with a closed-form pullback natural gradient (\S~\ref{Section:RHFF-LVM}).

\section{Background}
\label{Section:Background}
\subsection{Hyperbolic Space}
\begin{figure}[t]
    \centering
    \begin{subfigure}{0.15\textwidth}
        \centering
        \includegraphics[width=0.7\textwidth]{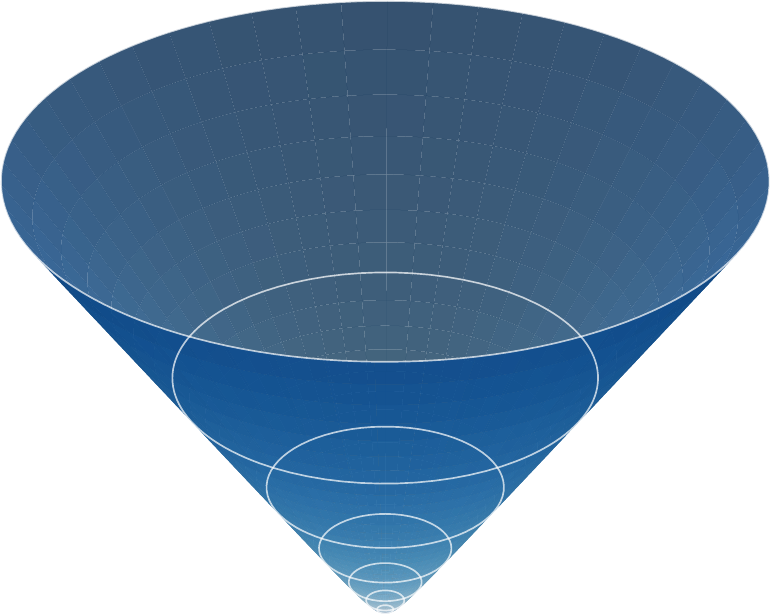}
        \caption{Lorentz model}
        \label{Fig:Lorentz-Model}
    \end{subfigure}
    \begin{subfigure}{0.15\textwidth}
        \centering
        \includegraphics[width=0.6\textwidth]{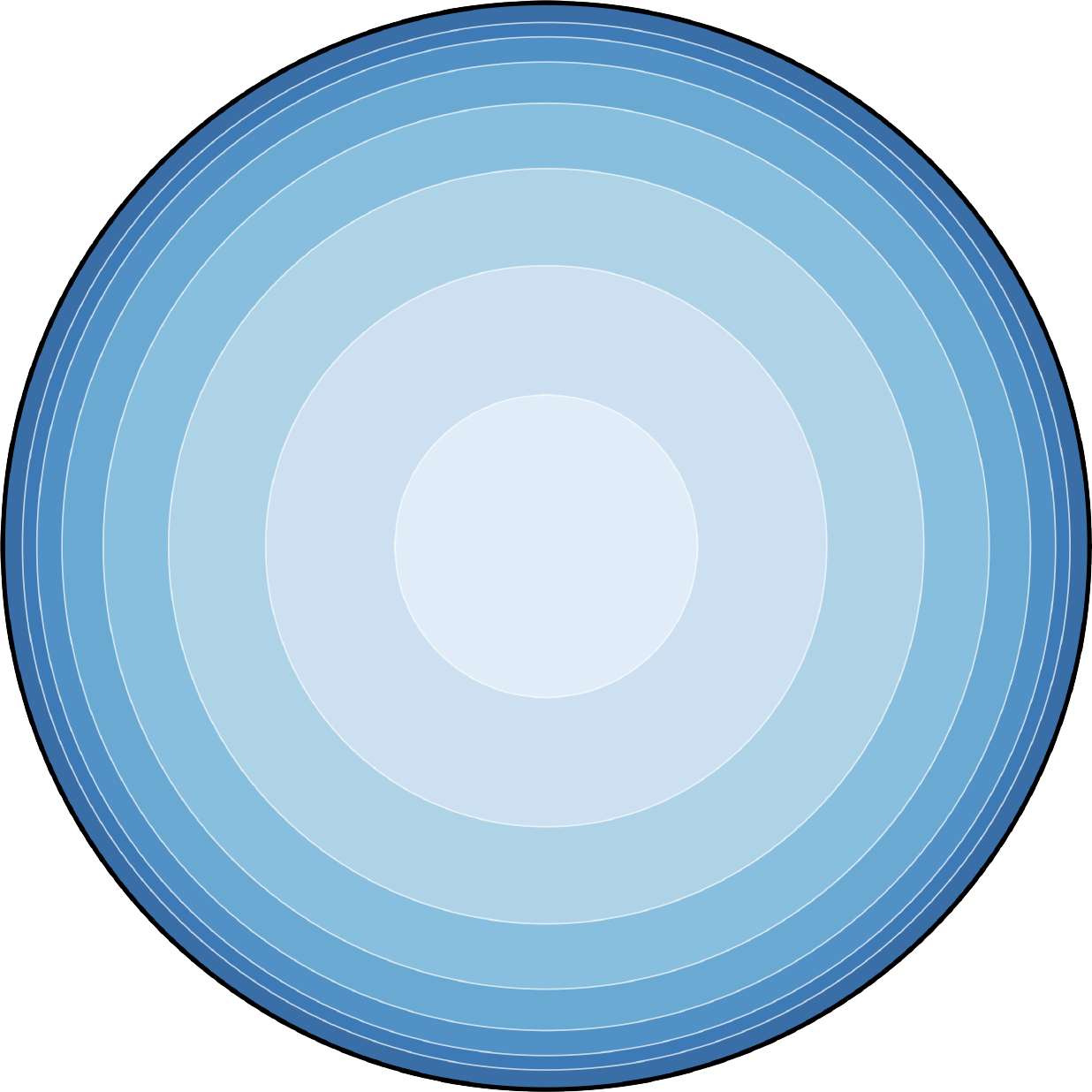}
        \caption{Poincar\'e ball}
        \label{Fig:Poincare-ball-model}
    \end{subfigure}
    \begin{subfigure}{0.16\textwidth}
        \centering
        \includegraphics[width=0.7\textwidth]{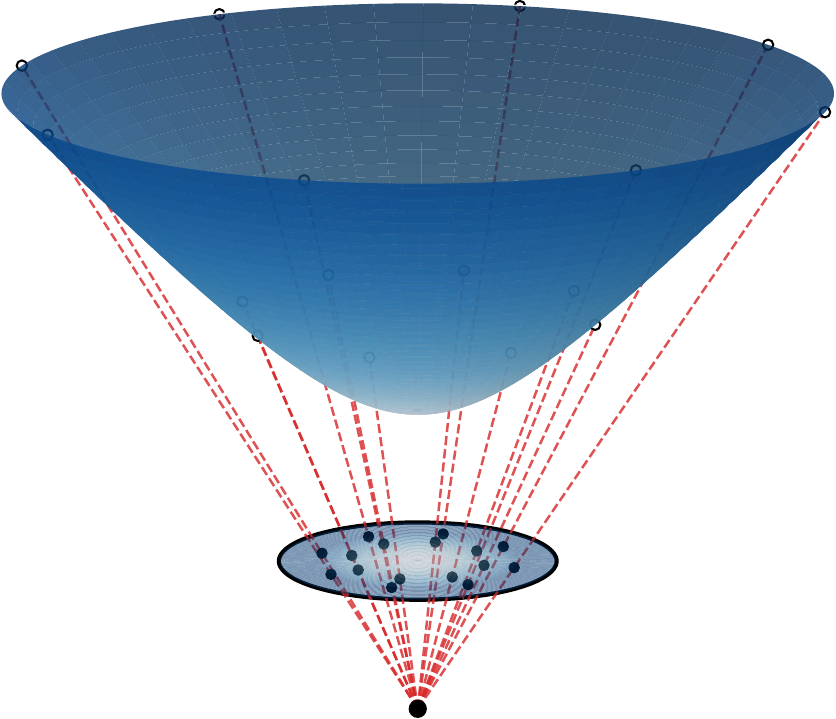}
        \caption{Diffeomorphism}
        \label{Fig:Diffeomorphism}
    \end{subfigure}
    \caption{Two hyperbolic models ($Q=2$) and the diffeomorphism between them.}
    \vspace{-1em}
    \label{Fig:Hyperbolic-Models}
\end{figure}

We adopt the \textit{Lorentz model} $\mathcal{L}^{Q}=(\mathbb{H}^{Q},\mathbf{G}_{l})$~\cite{cannon1997hyperbolic} (\Cref{Fig:Lorentz-Model}), the upper sheet of a $Q$-dimensional hyperboloid equipped with the metric induced by $\mathbf{G}_{l}=\mathrm{diag}(-1,1,\ldots,1)\in\mathbb{R}^{(Q+1)\times(Q+1)}$.
A point is a vector $\mathbf{x}=[x_{t},\mathbf{x}_{s}^{\top}]^{\top}$ with $x_{t}=\sqrt{1+\|\mathbf{x}_{s}\|_{2}^{2}}$.
The Lorentz inner product is $\langle\mathbf{x},\mathbf{x}'\rangle_{\mathcal{L}} = -x_{t}x'_{t}+\mathbf{x}_{s}^{\top}\mathbf{x}'_{s}$, and the geodesic distance is $d_{\mathcal{L}}(\mathbf{x},\mathbf{x}')=\cosh^{-1}(-\langle\mathbf{x},\mathbf{x}'\rangle_{\mathcal{L}})$.
The Lorentz model confines the manifold constraint to the single coordinate $x_{t}$, simplifying numerical implementation.
For visualization, we project onto the Poincar\'e ball (\Cref{Fig:Poincare-ball-model}) via the diffeomorphism $\mathbf{x}\mapsto\mathbf{x}_{s}/(1+x_{t})$ (\Cref{Fig:Diffeomorphism}).

\subsection{Random Fourier Features}
\label{Subsec:RFF}
Bochner's theorem~\cite{rasmussen2003gaussian} represents any continuous positive-definite \textit{stationary} kernel, normalized so that $k(\mathbf{x},\mathbf{x})=1$, by its spectral density $p(\bm{\omega})$:
\begin{align}
    k(\mathbf{x},\mathbf{x}') =
    \int_{\mathbb{R}^{Q}}p(\bm{\omega})
    e^{i\bm{\omega}^{\top}(\mathbf{x}-\mathbf{x}')}d\bm{\omega}.
    \label{eq:Fourier-Transform}
\end{align}
RFF~\cite{rahimi2007random} draws $\bm{\omega}_{1},\dots,\bm{\omega}_{M}$ \textit{i.i.d.}\ from $p(\bm{\omega})$ and approximates this integral by $k(\mathbf{x},\mathbf{x}')\approx \psi(\mathbf{x})^{\top}\psi(\mathbf{x}')$, where $\psi(\mathbf{x})=\frac{1}{\sqrt{M}}[\psi_{\bm{\omega}_{1}}(\mathbf{x})^{\top},\ldots,\psi_{\bm{\omega}_{M}}(\mathbf{x})^{\top}]^{\top}$ stacks $\psi_{\bm{\omega}_{j}}(\mathbf{x})=[\cos(\bm{\omega}_{j}^{\top}\mathbf{x}),\sin(\bm{\omega}_{j}^{\top}\mathbf{x})]^{\top}$, turning the kernel method into a linear model with feature matrix $\bm{\Psi}=[\psi(\mathbf{x}_{1}),\ldots,\psi(\mathbf{x}_{N})]^{\top}$ and reducing the cost from $\mathcal{O}(N^{3})$ to $\mathcal{O}(NM^{2}+M^{3})$.
Extending this construction to hyperbolic space requires (i)~a spectral representation of the hyperbolic kernels that replaces~\Cref{eq:Fourier-Transform} and admits sampling from its density (\S~\ref{Subsec:HF-Transform}), and (ii)~a decomposition of the integrand into an inner product of factors depending on $\mathbf{x}$ and $\mathbf{x}'$ separately (\S~\ref{Subsec:RHFF}).

\subsection{Helgason--Fourier Transform}
\label{Subsec:HF-Transform}
Helgason's harmonic analysis~\cite{helgason2022groups} generalizes the Fourier transform to non-compact symmetric spaces.
A hyperbolic isotropic kernel $k(\mathbf{x},\mathbf{x}')=k(d_{\mathcal{L}}(\mathbf{x},\mathbf{x}'))\triangleq k(r)$~\cite{azangulov2024stationary} is analyzed via the spherical transform \cite[p.~39]{helgason2022groups}:
\begin{align}
    \tilde{k}(\lambda) =
    \int_{0}^{\infty}k(r)\phi_{\lambda}(r)\sinh^{Q-1}(r)\,dr,
    \label{Eq:Spherical-Transform}
\end{align}
where $\lambda\in\mathbb{R}_{+}$, $\rho=(Q-1)/2$, and $\phi_{\lambda}(r)={}_2F_1(\rho+i\lambda,\rho-i\lambda;Q/2;-\sinh^2(r/2))$ is the elementary spherical function. Here ${}_2F_1$ is the Gauss hypergeometric function, defined by ${}_2F_1(a,b;c;z)=\sum_{n\ge0}(a)_n(b)_n z^n/((c)_n n!)$ for $|z|<1$ and by analytic continuation elsewhere, with $(a)_n$ the rising factorial.
By the Plancherel theorem \cite[Ch.~IV, Thm.~7.5]{helgason2022groups}, the kernel is recovered as
\begin{align}
    k(r) =
    \int_{0}^{\infty}\phi_{\lambda}(r)\tilde{k}(\lambda)
    |c(\lambda)|^{-2}d\lambda,
    \label{Eq:Inv-Spherical-Transform}
\end{align}
where $c(\lambda)$ is the Harish--Chandra $c$-function, whose explicit
form is given in \cite[Ch.~IV, Thm.~6.14]{helgason2022groups};
where $|c(\lambda)|^{-2} \propto \kappa_Q(\lambda) \prod_{j}(\lambda^2 + j^2)$ with $j \in \{0, 1, \dots, \rho-1\}$ and $\kappa_Q \equiv 1$ for odd $Q$, and $j \in \{\tfrac12, \tfrac32, \dots, \rho-1\}$ and $\kappa_Q(\lambda) = \lambda\tanh(\pi\lambda)$ for even $Q$.

\paragraph{Relation to the Euclidean Fourier transform.}
The weight $\tilde{k}(\lambda)|c(\lambda)|^{-2}$ plays the role of the spectral density $p(\bm{\omega})$, and $\phi_{\lambda}(r)$ that of the complex exponential $e^{i\bm{\omega}^{\top}(\mathbf{x}-\mathbf{x}')}$; the factor $|c(\lambda)|^{-2}$ compensates for the exponential volume growth of hyperbolic space.
Since $\phi_{\lambda}(0)=1$, evaluating~\Cref{Eq:Inv-Spherical-Transform} at $r=0$ shows that the weight $\tilde{k}(\lambda)|c(\lambda)|^{-2}$ integrates to $k(0)$ over $\lambda$. 
For a kernel normalized as $k(0)=1$, it thus defines a valid density $p(\lambda)\propto\tilde{k}(\lambda)|c(\lambda)|^{-2}$ on $\mathbb{R}_{+}$, so sampling $\lambda\sim p(\lambda)$ yields a Monte Carlo approximation of~\Cref{Eq:Inv-Spherical-Transform}.
The decomposition of $\phi_{\lambda}(r)$ into factors depending on $\mathbf{x}$ and $\mathbf{x}'$ is derived in \S~\ref{Subsec:RHFF}.

\section{Random Helgason-Fourier Features}
\label{Section:RHFF}
This section derives the closed-form RHFF map in
\Cref{eq:Busemann-Func,Eq:RHFF-Mapping}; readers interested mainly in
the model may skip to \S~\ref{Section:RHFF-LVM}.

\subsection{Busemann Function}
\label{Subsec:Busemann_Function}
\begin{figure}[t]
  \vspace{-1ex}
  \centering
  \begin{subfigure}{0.20\textwidth}
    \centering
    \includegraphics[width=0.8\textwidth]{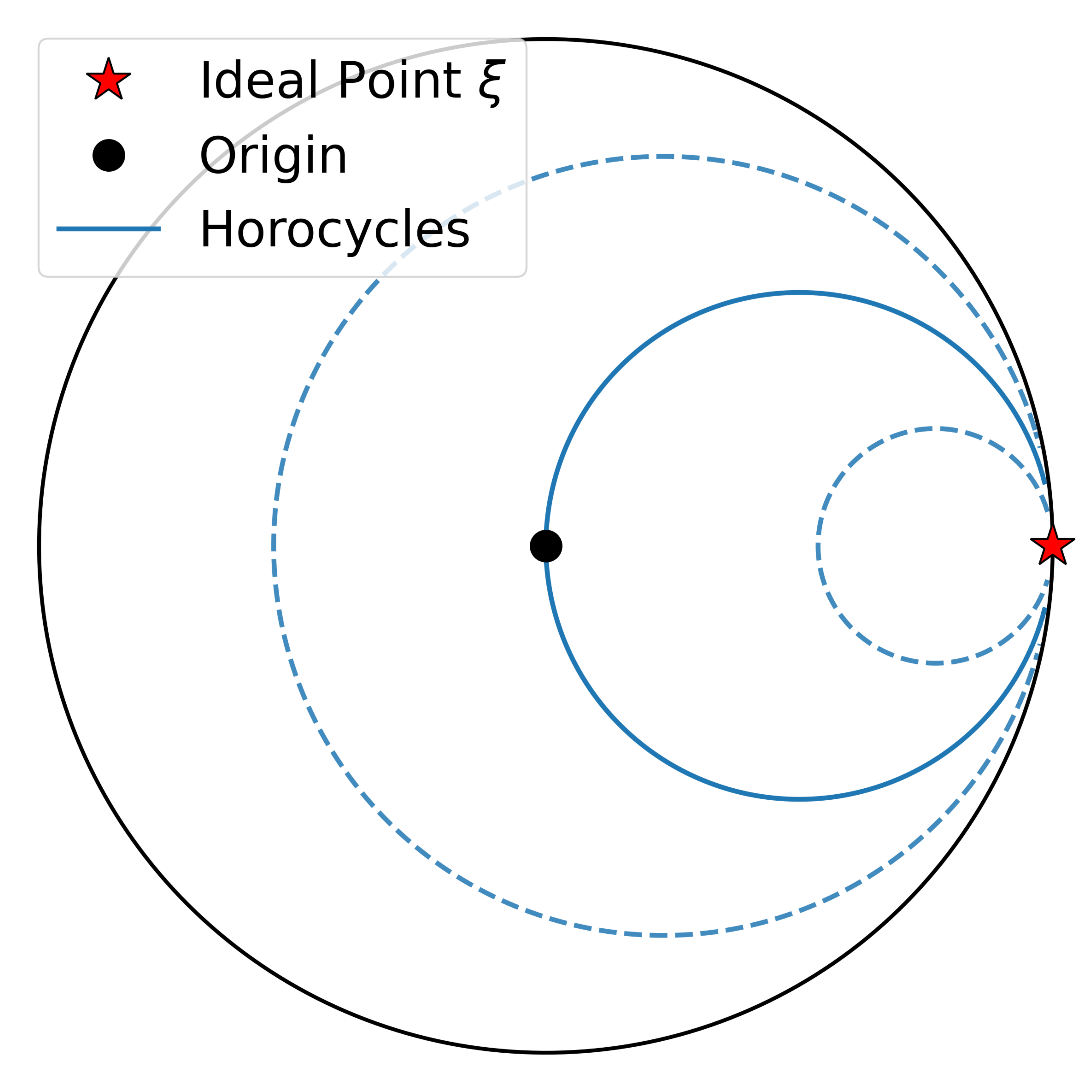}
    \caption{Horosphere.}
    \label{Fig:Horocycle}
  \end{subfigure}
  \begin{subfigure}{0.24\textwidth}
    \centering
    \includegraphics[width=0.8\textwidth]{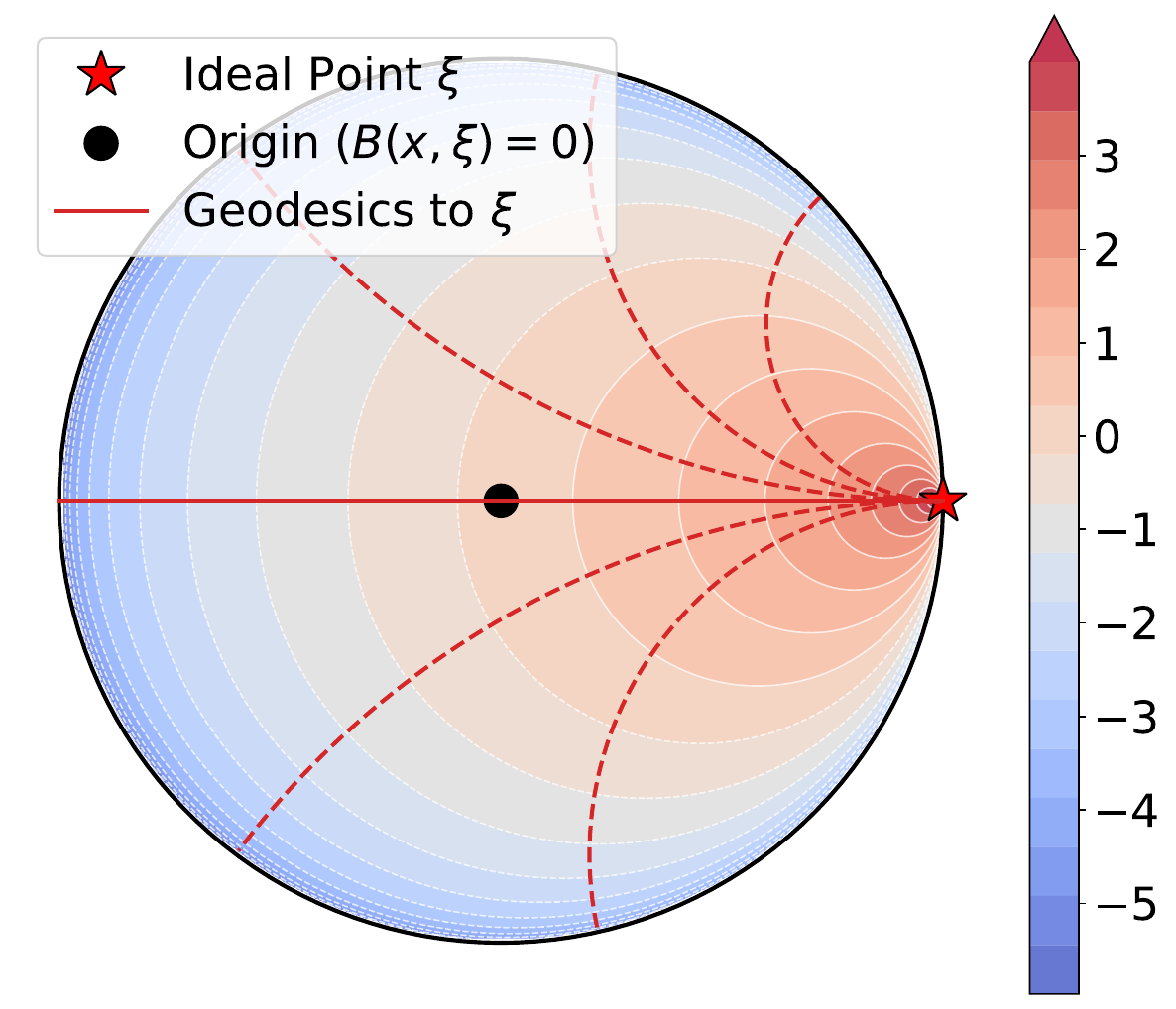}
    \caption{Busemann function.}
    \label{Fig:Busemann-func}
  \end{subfigure}
  \caption{The horosphere and Busemann function.}
  \label{Fig:Horocycle-Busemann}
  \vspace{-1ex}
\end{figure}

The building block of our features is the \textit{Busemann function}.
In hyperbolic geometry, directions are represented by \textit{ideal points} $\bm{\xi} \in \partial\mathcal{L}^{Q}$ on the boundary at infinity, identified with light-cone vectors satisfying $\langle\bm{\xi},\bm{\xi}\rangle_{\mathcal{L}}=0$ and $\xi_{t}=1$.
The Busemann function associated with $\bm{\xi}$ is defined as
\begin{align}
    B(\mathbf{x}, \bm{\xi}) &= - \log \left( -\langle\mathbf{x}, \bm{\xi}\rangle_{\mathcal{L}} \right). \label{eq:Busemann-Func}
\end{align}
Its level sets are horospheres (\Cref{Fig:Horocycle}).
The value $B(\mathbf{x},\bm{\xi})$ measures the signed distance from $\mathbf{x}$ toward the ideal point $\bm{\xi}$ (\Cref{Fig:Busemann-func}), providing an analogue of the Euclidean inner product $\bm{\omega}^{\top}\mathbf{x}$.

\subsection{Randomized Spherical Function}
\label{Subsec:RHFF}
To apply the RFF construction to~\Cref{Eq:Inv-Spherical-Transform}, we must decompose the elementary spherical function into factors depending on $\mathbf{x}$ and $\mathbf{x}'$ separately.
For this purpose, we employ the following integral representation of the spherical function \cite[Ch.~IV, Lemma 4.4]{helgason2022groups}, which holds for any $\mathbf{x},\mathbf{x}'\in\mathcal{L}^{Q}$ with $r=d_{\mathcal{L}}(\mathbf{x},\mathbf{x}')$:
\begin{align}
    \phi_{\lambda}(r) &= \int_{\partial\mathcal{L}^{Q}}e^{(\rho+i\lambda)B(\mathbf{x},\bm{\xi})}e^{(\rho-i\lambda)B(\mathbf{x}',\bm{\xi})}d\bm{\xi}, \label{Eq:Harish-Chandra-Formula-of-Spherical-Function}
\end{align}
where $d\bm{\xi}$ denotes the uniform probability measure on the boundary $\partial\mathcal{L}^{Q} \cong \mathbb{S}^{Q-1}$.
Leveraging this plane-wave representation, we establish the following proposition:
\begin{prop}
    Let $\psi_{\lambda,\bm{\xi}}(\mathbf{x})=\left[e^{\rho B(\mathbf{x},\bm{\xi})}\cos\left(\lambda B(\mathbf{x},\bm{\xi})\right), e^{\rho B(\mathbf{x},\bm{\xi})}\sin\left(\lambda B(\mathbf{x},\bm{\xi})\right)\right]^{\top}$ be the feature map associated with a frequency $\lambda\in\mathbb{R}_{+}$ and an ideal direction $\bm{\xi} \in \partial\mathcal{L}^{Q}$.
    Then, for any $\mathbf{x},\mathbf{x}'\in\mathcal{L}^{Q}$ with $r=d_{\mathcal{L}}(\mathbf{x},\mathbf{x}')$,
    \begin{align}
        \phi_{\lambda}(r)
        &= \int_{\partial\mathcal{L}^{Q}} \psi_{\lambda,\bm{\xi}}(\mathbf{x})^{\top}\psi_{\lambda,\bm{\xi}}(\mathbf{x}')d\bm{\xi}. \label{Eq:wave-decomposition}
    \end{align}
    \label{Prop:Discrete-Spherical-Func}
    \vspace{-1.5em}
\end{prop}
\begin{proof}
    The proof is provided in \S~\ref{Section:Proof-of-Discrete-Spherical-Func}.
\end{proof}
\begin{figure}[t]
    \centering
    \begin{subfigure}{0.11\textwidth}
        \centering
        \includegraphics[width=0.99\textwidth]{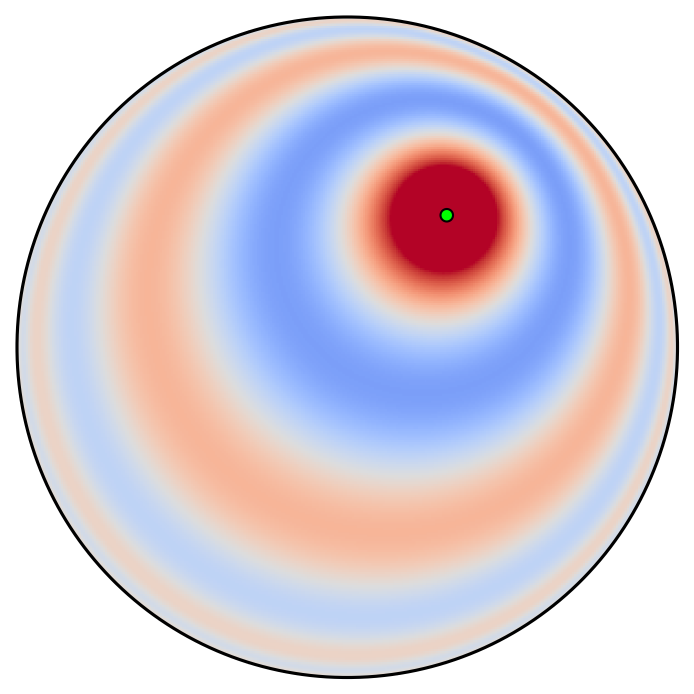}
        \caption{$\phi_{\lambda}(r)$}
    \end{subfigure}
    \begin{subfigure}{0.11\textwidth}
        \centering
        \includegraphics[width=0.99\textwidth]{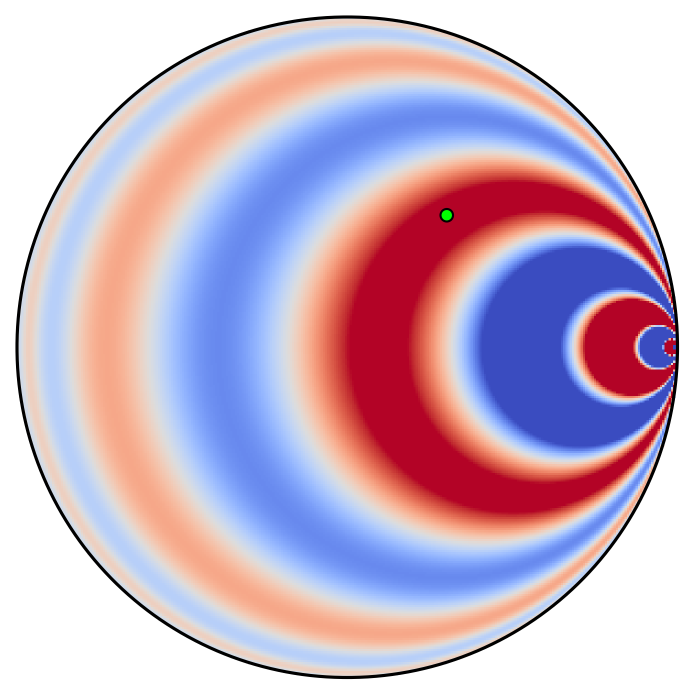}
        \caption{$M=1$}
    \end{subfigure}
    \begin{subfigure}{0.11\textwidth}
        \centering
        \includegraphics[width=0.99\textwidth]{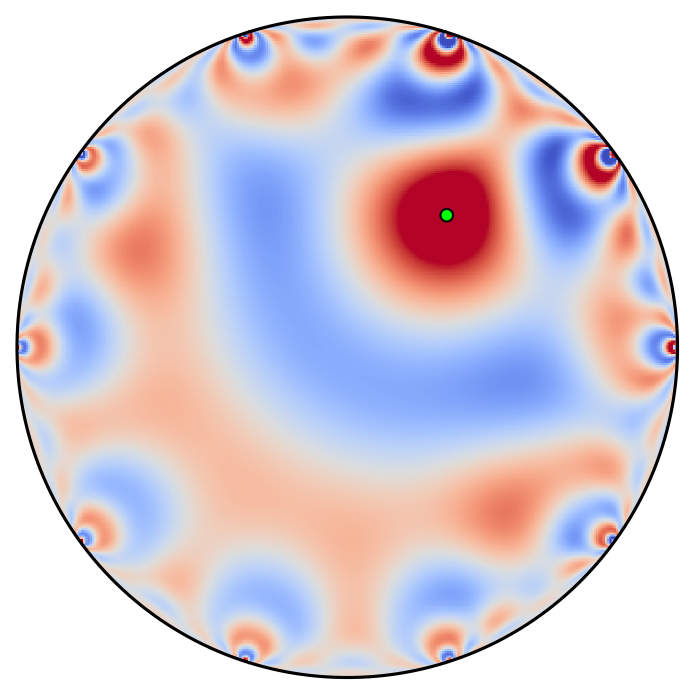}
        \caption{$M=10$}
    \end{subfigure}
    \begin{subfigure}{0.11\textwidth}
        \centering
        \includegraphics[width=0.99\textwidth]{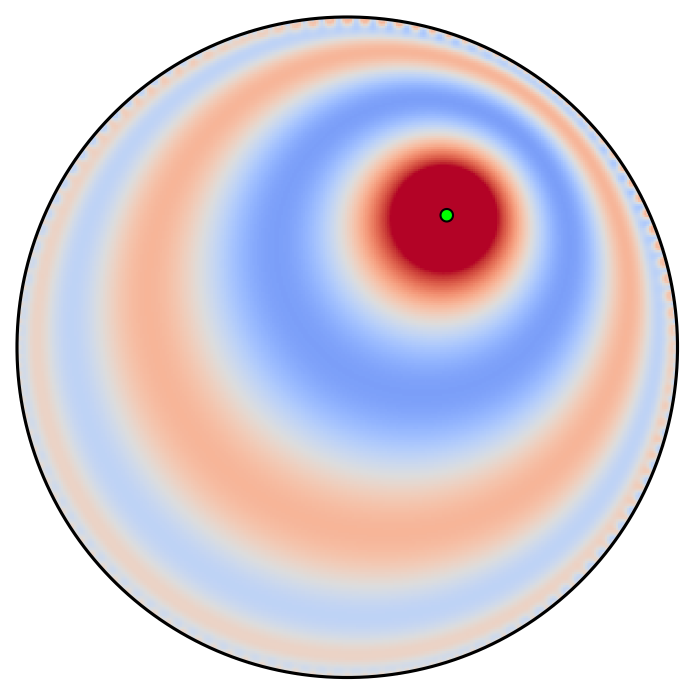}
        \caption{$M=120$}
    \end{subfigure}
    \caption{The elementary spherical function $\phi_{\lambda}(r)$, viewed as a function of $\mathbf{x}$ with the origin $\mathbf{o}$ as the reference point, and its approximation with several plane waves $\psi_{\lambda,\bm{\xi}}(\mathbf{x})$.}
    \label{Fig:Discrete-Spherical-Func}
    \vspace{-1ex}
\end{figure}
\Cref{Fig:Discrete-Spherical-Func} illustrates \Cref{Prop:Discrete-Spherical-Func}.
The elementary spherical function is a continuous superposition of plane waves emanating from the ideal points.
Geometrically, averaging over all ideal directions how the two points sit relative to the horospheres reconstructs their distance $r$.
Substituting the decomposition of~\Cref{Eq:wave-decomposition} into~\Cref{Eq:Inv-Spherical-Transform} and applying Monte Carlo sampling yields
\begin{align}
    k(r) &= \int_{0}^{\infty}\int_{\partial\mathcal{L}^{Q}} \psi_{\lambda,\bm{\xi}}(\mathbf{x})^{\top}\psi_{\lambda,\bm{\xi}}(\mathbf{x}')\tilde{k}(\lambda)|c(\lambda)|^{-2}\,d\bm{\xi}\,d\lambda \notag \\
    &\approx \frac{1}{M}\sum_{j=1}^{M} \psi_{\lambda_{j},\bm{\xi}_{j}}(\mathbf{x})^{\top}\psi_{\lambda_{j},\bm{\xi}_{j}}(\mathbf{x}'), \label{Eq:RHFF-Approximation}
\end{align}
where the $2$-dimensional random feature for the $j$-th sample is defined as
\begin{align}
    \psi_{\lambda_{j},\bm{\xi}_{j}}(\mathbf{x})
    = e^{\rho B(\mathbf{x},\bm{\xi}_{j})}
    \begin{bmatrix}
        \cos\bigl(\lambda_{j} B(\mathbf{x},\bm{\xi}_{j})\bigr) \\[2pt]
        \sin\bigl(\lambda_{j} B(\mathbf{x},\bm{\xi}_{j})\bigr)
    \end{bmatrix}. \label{Eq:RHFF-Mapping}
\end{align}
We refer to these features collectively as \textit{Random Helgason--Fourier Features} (RHFF).
Here, $\bm{\xi}_{j}$ are sampled \textit{i.i.d.}\ uniformly from the boundary $\mathbb{S}^{Q-1}$ and $\lambda_{j}$ \textit{i.i.d.}\ from the density $p(\lambda)$ defined in \S~\ref{Subsec:HF-Transform}.
Stacking these over $j$ into $\psi(\mathbf{x}) = \frac{1}{\sqrt{M}}\left[\psi_{\lambda_{1},\bm{\xi}_{1}}(\mathbf{x})^{\top},\ldots,\psi_{\lambda_{M},\bm{\xi}_{M}}(\mathbf{x})^{\top}\right]^{\top}\in\mathbb{R}^{2M}$ yields the feature map used in \S~\ref{Section:RHFF-LVM}.
$\{\lambda_{j},\bm{\xi}_{j}\}_{j=1}^{M}$ are sampled once and kept fixed thereafter, unlike inducing inputs which are optimized jointly with the latents.
\par
Because RHFF samples directly from $p(\lambda)$, any isotropic kernel with a known spherical transform can be specified directly in the frequency domain~\cite{azangulov2024stationary}.
For instance, the heat kernel has spherical transform
\begin{align}
    \tilde{k}_{t}(\lambda) &\propto \exp\left(-t(\lambda^{2}+\rho^{2})\right), \label{Eq:Heat-Kernel}
\end{align}
where $t$ denotes the length-scale hyperparameter.
To mitigate its rapid spectral decay, the Mat\'ern family~\cite{azangulov2024stationary} is also widely used, characterized by
\begin{align}
    \tilde{k}_{\nu\kappa}(\lambda) &\propto \left( \lambda^{2}+\rho^{2}+\frac{2\nu}{\kappa^{2}} \right)^{-\left(\nu+\frac{Q}{2}\right)}, \label{Eq:Matern-Kernel}
\end{align}
where $\nu$ and $\kappa$ are hyperparameters controlling the smoothness and length-scale, respectively.

\paragraph{Connection to HyLa~\cite{yu2023random}.}
Sampling $\lambda \sim \mathcal{N}(0,s^{2})$ and randomizing the phase of~\Cref{Eq:RHFF-Mapping} recovers the feature map of HyLa~\cite{yu2023random}, which is therefore the instance of RHFF that implicitly specifies
\begin{align}
    \tilde{k}_{\mathrm{HyLa}}(\lambda) \propto e^{-\lambda^{2}/(2s^{2})}|c(\lambda)|^{2}.
    \label{Eq:HyLa-Spectrum}
\end{align}
This differs from the heat spectrum of~\Cref{Eq:Heat-Kernel} by the inverse of the Plancherel weight and therefore over-weights low frequencies.
The proof is given in \S~\ref{Section:Connection-to-HyLa}, which additionally shows that RHFF Rao--Blackwellizes HyLa.

\section{RHFF-LVM}
\label{Section:RHFF-LVM}
Having constructed the features, we now build the latent variable model (RHFF-LVM) on them. 
\S~\ref{Subsec:Objective} places a Bayesian linear model on the RHFF representation, and \S~\ref{Subsec:Pullback} optimizes the latents by natural gradient descent on unconstrained Euclidean coordinates, without Riemannian updates.

\subsection{Objective Function}
\label{Subsec:Objective}
Let $\mathbf{Y}=[\mathbf{y}_{1},\dots,\mathbf{y}_{N}]^{\top}\in\mathbb{R}^{N\times D}$ denote the $N$ observed $D$-dimensional data points, and $\mathbf{X}=[\mathbf{x}_{1},\dots,\mathbf{x}_{N}]^{\top}$ the corresponding latent variables, where each $\mathbf{x}_{n} \in \mathcal{L}^{Q}$.
Dimensionality reduction then amounts to learning $\mathbf{X}$ from $\mathbf{Y}$.
We model the observations as generated by a Bayesian linear model on the RHFF representation, $\mathbf{Y}=\bm{\Psi}\mathbf{W}+\mathbf{E}$, where the weights $\mathbf{W}\in\mathbb{R}^{2M\times D}$ have i.i.d.\ $\mathcal{N}(0,1)$ entries and the noise $\mathbf{E}$ has i.i.d.\ $\mathcal{N}(0,\beta^{-1})$ entries~\cite{gundersen2021latent,lawrence2005probabilistic}.
Marginalizing $\mathbf{W}$, we optimize the latent variables and $\beta$ by minimizing the negative log-marginal likelihood (up to constants):
\begin{align}
    \ell(\mathbf{X},\beta) = \frac{D}{2} \log |\mathbf{C}| + \frac{1}{2}\mathrm{tr}\left( \mathbf{C}^{-1}\mathbf{Y}\mathbf{Y}^{\top} \right), \label{Eq:GP-Likelihood}
\end{align}
where $\mathbf{C}=\mathbf{\Psi}\mathbf{\Psi}^{\top}+\beta^{-1}\mathbf{I}$ is the covariance matrix and $\mathbf{\Psi}\in\mathbb{R}^{N \times 2M}$ is the feature matrix whose $n$-th row is $\psi(\mathbf{x}_{n})^{\top}$.
We further add a latent prior that penalizes the squared geodesic distance to the origin, $\frac{\gamma}{D}\sum_{n=1}^{N} d_{\mathcal{L}}(\mathbf{o},\mathbf{x}_{n})^{2}$ with $\gamma=0.5$, and minimize the resulting MAP objective.
Beyond regularization, this prior encourages the latents to stay within the bounded domain in which the approximation guarantee of \S~\ref{Section:Error-Bound} applies.
The time complexity is reduced from $\mathcal{O}(N^{3})$ to $\mathcal{O}(NM^{2}+M^{3})$ via the Woodbury identity and the matrix determinant lemma (\Cref{Eq:Det-Lemma,Eq:Inv-Lemma}).

\subsection{Optimization via Pullback Metric}
\label{Subsec:Pullback}
In standard Riemannian gradient descent on the Lorentz manifold~\cite{nickel2018learning}, numerical errors from tangent-space projections and retractions may violate the hyperboloid constraint $x_{t}=\sqrt{1+\|\mathbf{x}_{s}\|_{2}^{2}}$.
We circumvent this by exploiting the fact that the objective in~\Cref{Eq:GP-Likelihood} depends on each latent variable only through $\mathbf{x}_{n}=\mathcal{P}(\mathbf{x}_{n,s})$, where $\mathcal{P}(\mathbf{x}_{s}) = [\sqrt{1+\|\mathbf{x}_{s}\|_{2}^{2}}, \mathbf{x}_{s}^{\top}]^{\top}$ lifts the spatial component $\mathbf{x}_{s}\in\mathbb{R}^{Q}$ onto the manifold.
Since $\mathcal{P}$ is a global diffeomorphism from $\mathbb{R}^{Q}$ onto $\mathcal{L}^{Q}$, minimizing $\ell$ over the unconstrained variables $\{\mathbf{x}_{n,s}\}_{n=1}^{N}$ is exactly equivalent to minimizing it over the manifold.
The natural gradient can therefore be computed in the unconstrained space under the pullback metric of $\mathcal{P}$, with no projections or retractions.
\par
The pullback metric is available in closed form: the Jacobian of $\mathcal{P}$ is
\begin{align}
    \frac{\partial \mathcal{P}(\mathbf{x}_{s})}{\partial\mathbf{x}_{s}} &= \begin{pmatrix} \frac{\mathbf{x}_{s}^\top}{\sqrt{1 + \|\mathbf{x}_{s}\|^2_2}} \\ \mathbf{I} \end{pmatrix} \in \mathbb{R}^{(Q+1) \times Q}, \label{Eq:Jacobian-of-Px}
\end{align}
and the pullback metric $\mathbf{G} = \left(\frac{\partial \mathcal{P}(\mathbf{x}_{s})}{\partial\mathbf{x}_{s}}\right)^{\top} \mathbf{G}_{l} \left(\frac{\partial \mathcal{P}(\mathbf{x}_{s})}{\partial\mathbf{x}_{s}}\right)$ simplifies to $\mathbf{G} = \mathbf{I} - (1+\|\mathbf{x}_s\|_2^2)^{-1}\mathbf{x}_s \mathbf{x}_s^\top$.
Inverting $\mathbf{G}$ via the Sherman--Morrison formula yields the natural gradient with respect to this metric in terms of the Euclidean gradient $\nabla^{E}_{\mathbf{x}_{s}}\ell$:
\begin{align}
    \nabla_{\mathbf{x}_{s}}^{\mathcal{L}}\ell &= \left(\mathbf{I}+\mathbf{x}_{s}\mathbf{x}_{s}^{\top}\right)\nabla^{E}_{\mathbf{x}_{s}}\ell. \label{Eq:Natural-Gradient}
\end{align}
The latent variables are estimated by iterating $\mathbf{x}_{s} \leftarrow \mathbf{x}_{s} - \eta \nabla_{\mathbf{x}_{s}}^{\mathcal{L}}\ell$, recovering the Lorentz point as $\mathcal{P}(\mathbf{x}_{s})$ at each step; the step size $\eta$ is set by backtracking line search, with gradient-norm clipping (\S~\ref{App:Setup}). 
The Euclidean gradient computation is provided in \S~\ref{Section:Gradient-Computation}, and details of the RHFF-LVM algorithm are in \S~\ref{Section:Algorithm}.

\section{Uniform Bound on the RHFF Error}
\label{Section:Error-Bound}
Finally, we analyze the approximation quality of RHFF, adapting the standard framework of prior work~\cite{rahimi2007random,sutherland2015error} to the hyperbolic setting.
\begin{prop}
    Let $\mathcal{X}$ be a compact subset of $\mathcal{L}^Q$ ($Q \ge 2$) containing the origin, with radius $R$.
    Suppose $k(\mathbf{x}, \mathbf{x}') = k(d_{\mathcal{L}}(\mathbf{x}, \mathbf{x}'))$ is an isotropic kernel on $\mathcal{L}^{Q}$ with $k(0) = 1$, differentiable at $r=0$, and let $\{\lambda_{j},\bm{\xi}_{j}\}_{j=1}^{M}$ be i.i.d.\ with $\lambda\sim p(\lambda)$, $\bm{\xi}$ uniform on $\partial\mathcal{L}^{Q}$, and $\sigma^{2}_{\lambda}=\mathbb{E}[\lambda^2]<\infty$.
    Let $\hat{k}(\mathbf{x}, \mathbf{x}') = \frac{1}{M} \sum_{j=1}^{M}\psi_{\lambda_{j},\bm{\xi}_{j}}(\mathbf{x})^{\top}\psi_{\lambda_{j},\bm{\xi}_{j}}(\mathbf{x}')$ denote the RHFF estimator, which satisfies $k(\mathbf{x},\mathbf{x}')=\mathbb{E}_{\lambda,\bm{\xi}}[\psi_{\lambda,\bm{\xi}}(\mathbf{x})^{\top}\psi_{\lambda,\bm{\xi}}(\mathbf{x}')]$ on $\mathcal{X}$.
    Then, for any $\epsilon > 0$ and $M \ge 1$, the error $\hat{f}=\hat{k} - k$ satisfies
    \begin{equation}
        \begin{split}
            & \Pr\left( \sup_{\mathbf{x}, \mathbf{x}' \in \mathcal{X}} |\hat{f}(\mathbf{x},\mathbf{x}')| \geq \epsilon \right) \\
            &\leq C e^{2(Q-1)R - \frac{M\epsilon^2}{8(Q+1)e^{2(Q-1)R}}},
        \end{split}
        \label{Eq:Uniform-bound}
    \end{equation}
    where $C$, which depends on $\epsilon$ but not on $M$, is given by
    \begin{equation*}
        C = \frac{16(\rho^{2}+\sigma_{\lambda}^{2})}{\epsilon^{2}} + 2Q^{2}\left(R+\frac{1}{4}\right)^2 \coth^{2Q}\left(\frac{1}{4}\right).
    \end{equation*}
    \label{Prop:Error-Bound}
\vspace{-2em}
\end{prop}
\begin{proof}
    The proof combines a covering argument on $\mathcal{X}\times\mathcal{X}$ (necessary since isotropic kernels on $\mathcal{L}^{Q}$ are not translation invariant),
    Lipschitz bounds for the Busemann features, hyperbolic volume comparison,
    and Hoeffding's inequality; details are in
    \S~\ref{Section:Proof-of-Error-Bound}.
\end{proof}
\paragraph{Discussion.}
The exponent in~\Cref{Eq:Uniform-bound} scales linearly with $M$, so the probability of a large error decays exponentially as $M \to \infty$.
In contrast, the bound deteriorates exponentially with the radius $R$ and the dimension $Q$, reflecting the exponential volume growth of hyperbolic balls.
The $M$ required for a target accuracy scales exponentially with both $Q$ and $R$---a \textit{curse of dimensionality and space} that delineates the practical limits of hyperbolic random features.
\par
\begin{figure}[t]
  \vspace{-1ex}
  \centering
  \begin{subfigure}{0.15\textwidth}
    \centering
    \includegraphics[width=0.99\textwidth]{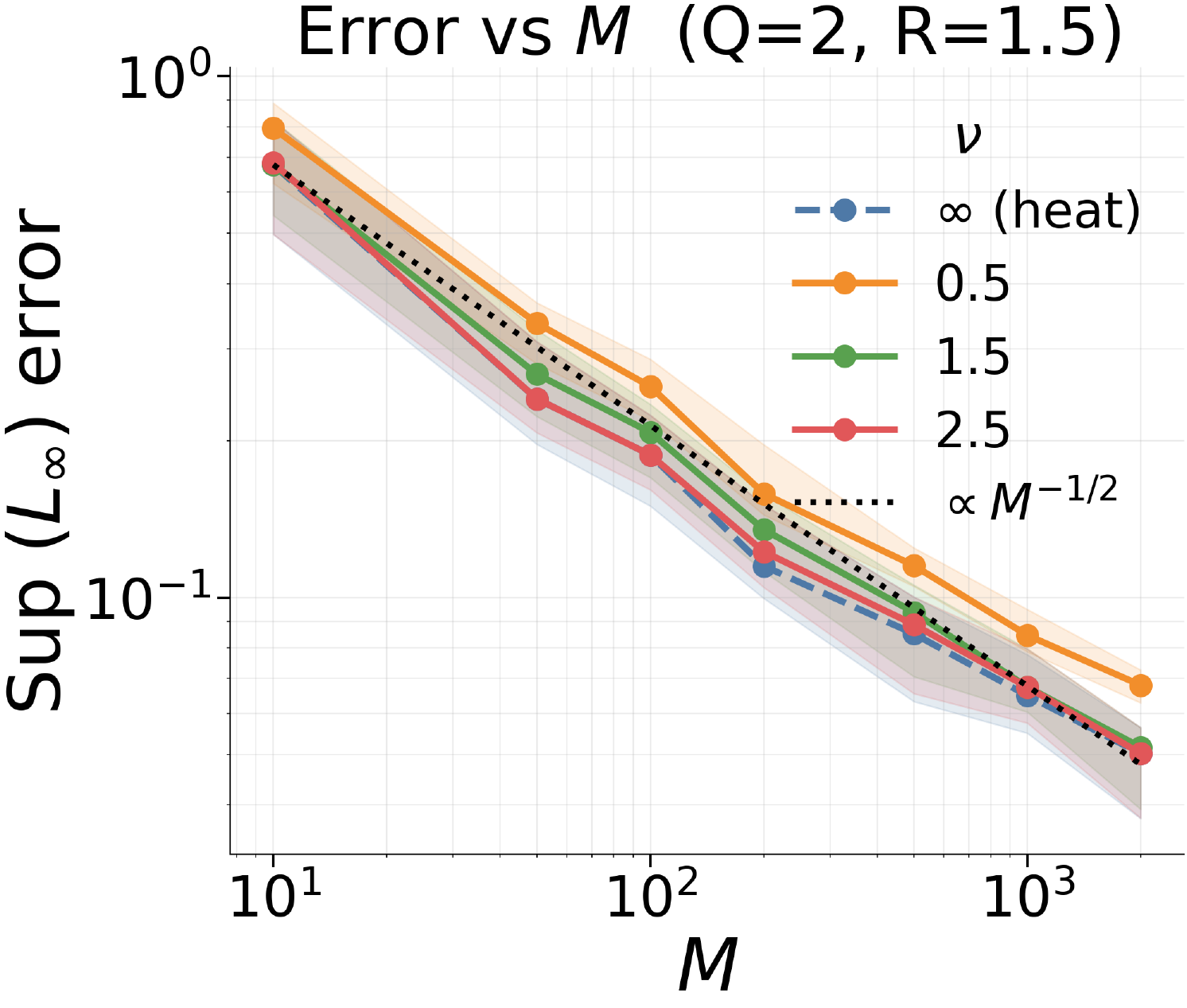}
    \caption{Basis size $M$.}
    \label{Fig:Kernel-Approx-Sup-M}
  \end{subfigure}
  \begin{subfigure}{0.15\textwidth}
    \centering
    \includegraphics[width=0.99\textwidth]{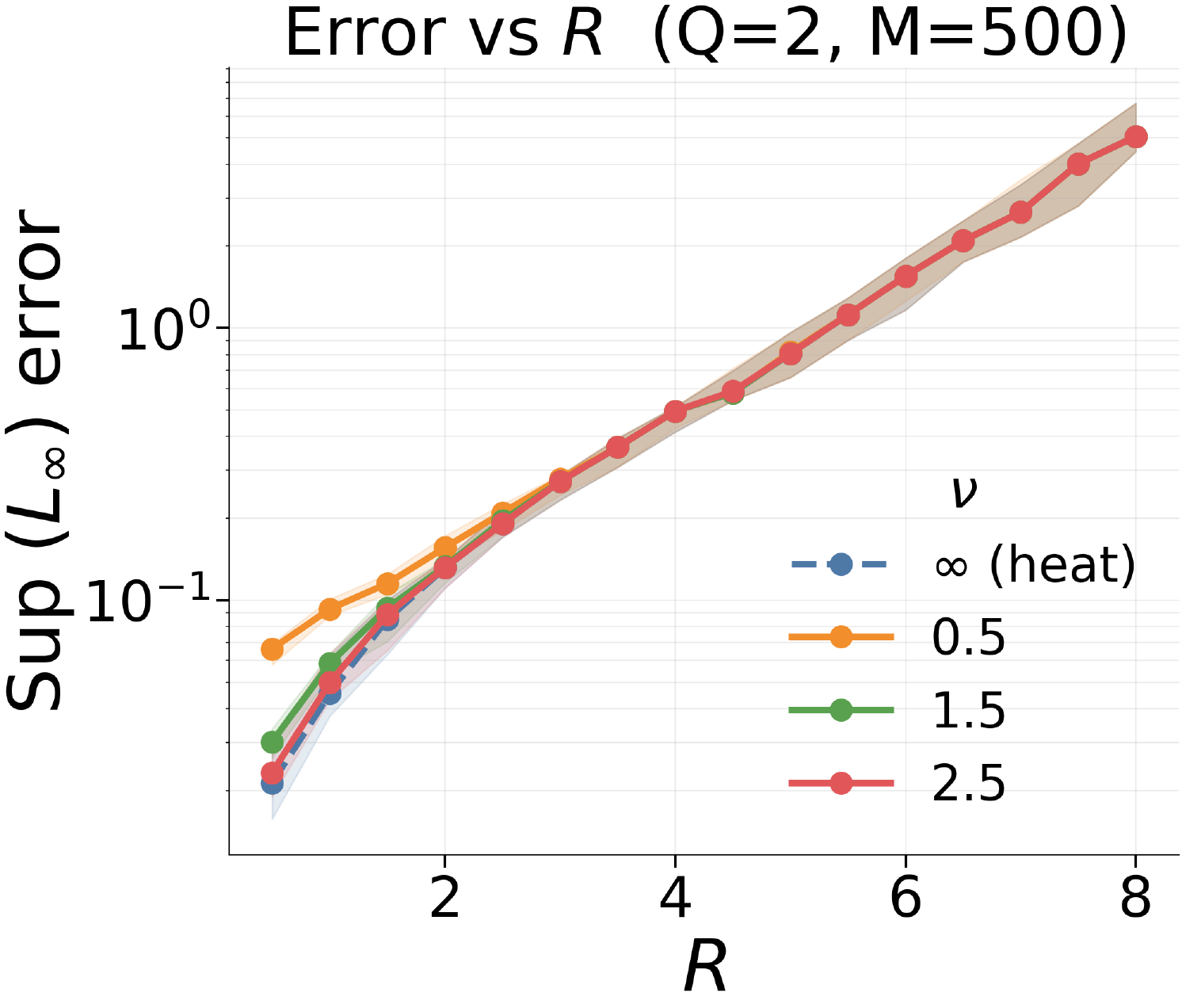}
    \caption{Radius $R$.}
    \label{Fig:Kernel-Approx-Sup-R}
  \end{subfigure}
  \begin{subfigure}{0.15\textwidth}
    \centering
    \includegraphics[width=0.99\textwidth]{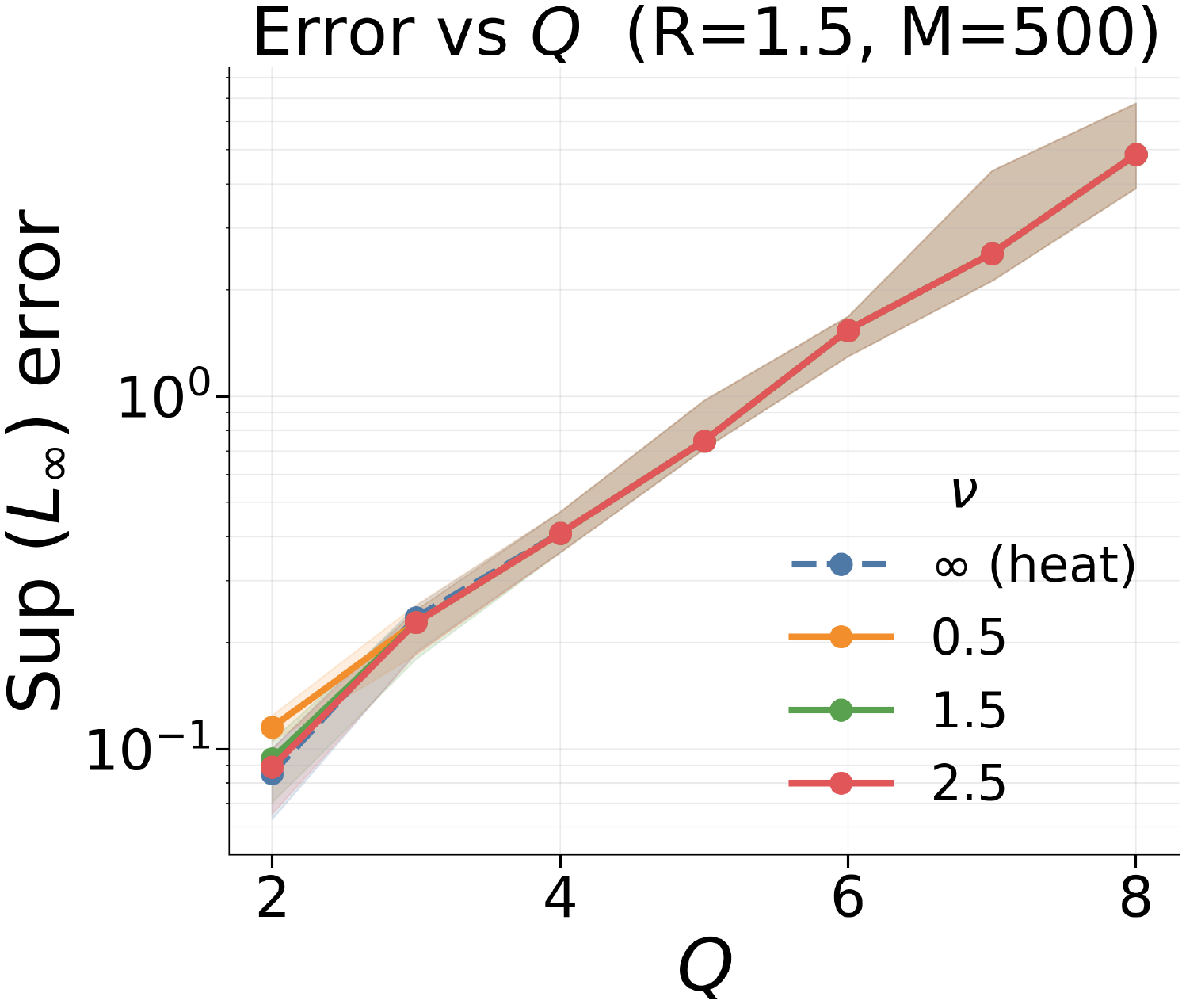}
    \caption{Dimension $Q$.}
    \label{Fig:Kernel-Approx-Sup-Q}
  \end{subfigure}
  \caption{Approximation error of RHFF relative to the exact kernel \cite{azangulov2024stationary} over a geodesic ball of radius $R$ (median and interquartile range over 20 random feature draws).}
  \label{Fig:Approximation-Error}
  \vspace{-1ex}
\end{figure}

\Cref{Fig:Approximation-Error} empirically corroborates both trends in the same uniform norm as the bound.
The error decays at the Monte Carlo rate $M^{-1/2}$ (\Cref{Fig:Kernel-Approx-Sup-M}), and at a fixed $M$ grows sharply with $R$ and exponentially with $Q$ (\Cref{Fig:Kernel-Approx-Sup-R,Fig:Kernel-Approx-Sup-Q}), milder than but qualitatively matching the worst-case rate.
Heavier-tailed spectra are consistently harder to approximate.
In practice, for the heat and Mat\'ern spectra we sample $\lambda$ by rejection sampling from $p(\lambda)$ restricted to a bounded range, which also restores the finiteness of $\sigma^{2}_{\lambda}$ for the Mat\'ern family at small $\nu$.
The discarded tail probability is negligible except for Mat\'ern $\nu{=}0.5$, which we quantify in \S~\ref{App:Setup}.

\section{Experiments}
\label{Section:Experiments}

\subsection{Experimental Setup}
\label{Subsec:Setup}
As Euclidean baselines, we use PCA, UMAP~\cite{mcinnes2018umap}, and the sparse variational Bayesian GP-LVM (BGP-LVM)~\cite{titsias2010bayesian}.
As hyperbolic baselines, we use Poincar\'e Map~\cite{klimovskaia2020poincare}, the Bayesian hyperbolic GP-LVM (BHGP-LVM)~\cite{pmlr-v258-watanabe25a}, and accelerated hyperbolic t-SNE (Ht-SNE)~\cite{skrodzki2024accelerating}, of which BHGP-LVM and Ht-SNE run at $Q=2$.
As an ablation, RFF-LVM~\cite{gundersen2021latent} replaces our RHFF with Gaussian random features~\cite{rahimi2007random}, keeping the model and optimizer identical.
Since the Gaussian kernel is the Euclidean heat kernel, comparing RFF-LVM with RHFF-LVM (heat) isolates the effect of the geometry.
We evaluate RHFF ($M=500$) with five spectra: the HyLa spectrum~\cite{yu2023random}, the heat kernel ($t=1.0$), and the Mat\'ern family with $\nu=0.5,\,1.5,\,2.5,\,(\kappa=2.0)$.
We report the correlation between observed and embedded pairwise distances~\cite{espadoto2019toward} (\textbf{glo}; Spearman in \S~\ref{Subsec:Toy}, Pearson in \S~\ref{Subsec:RealData}), and the $k$NN recall (\textbf{loc}), the fraction of each point's $k=10$ ambient neighbors recovered. 
PCA is equivalent to classical MDS and is thus the linear optimum for global distance preservation, serving as the reference point for \textbf{glo}.
Both metrics are reported $\times 100$, averaged over five seeds that vary the latent initialization and the feature draw.

\subsection{Synthetic Binary Trees (SBT)}
We first isolate the effect of each design choice on a synthetic binary tree dataset, a standard benchmark for hyperbolic generative models~\cite{cho2022rotated,nagano2019wrapped,pmlr-v258-watanabe25a}.
Every node of a complete binary tree of depth $6$ is represented by its indicator vector ($D=63$).
For each node, we generate $16$ samples by scaling each path entry by $\cos\theta$ and its sibling entry by $\sin\theta$, with $\theta \sim U[0,\pi/4)$ ($N=(2^{6}-1)\times16= 1{,}008$).
In addition to \textbf{glo} and \textbf{loc}, we report \textbf{tree corr}, the Spearman correlation between embedded geodesic distances and true path lengths, which measures hierarchy preservation directly.
\par
\begin{table}[t]
  \centering
  \caption{Local and global embedding quality at $Q=2$. \textbf{loc} denotes $k$NN recall, and \textbf{glo} denotes Shepard goodness, the Spearman correlation of the pairwise distances. Values are $\times100$. \textbf{tree corr} is the Spearman correlation with the true tree path-length (geodesic) distance. \textbf{bold} denotes the best per column, and \colorbox{lightblue}{shaded} denotes the best RHFF-LVM variant.}
  \label{Tab:Toy-Results}
  \resizebox{0.47\textwidth}{!}{%
  \begin{tabular}{l|cc|c}
    \toprule
    & \multicolumn{2}{c|}{\shortstack{Toy-tree\\{\scriptsize $N{=}1.0k$}}} & \multicolumn{1}{c}{} \\
    \cmidrule(lr){2-3}
    Method & loc & glo & tree corr \\
    \midrule
    PCA & \ms{14.9}{0.0} & \ms{79.7}{0.0} & \ms{77.2}{0.0} \\
    UMAP~\cite{mcinnes2018umap} & \ms{69.0}{0.4} & \ms{58.7}{3.9} & \ms{58.4}{3.4} \\
    BGP-LVM~\cite{titsias2010bayesian} & \ms{71.5}{0.3} & \ms{75.8}{1.5} & \ms{75.5}{1.3} \\
    RFF-LVM~\cite{gundersen2021latent} & \ms{72.2}{0.7} & \ms{74.9}{1.6} & \ms{74.9}{1.4} \\
    \midrule
    Poincar\'e Map~\cite{klimovskaia2020poincare} & \ms{69.4}{0.4} & \ms{83.6}{0.8} & \ms{82.4}{0.7} \\
    BHGP-LVM~\cite{pmlr-v258-watanabe25a} & \ms{1.9}{0.1} & \ms{51.4}{3.0} & \ms{51.8}{2.8} \\
    Ht-SNE~\cite{skrodzki2024accelerating} & \ms{54.3}{3.0} & \ms{50.4}{9.7} & \ms{51.5}{8.8} \\
    \midrule
    RHFF-LVM (HyLa)~\cite{yu2023random} & \ms{74.3}{0.6} & \ms{82.8}{0.9} & \ms{83.4}{0.8} \\
    RHFF-LVM ($\nu=0.5$) & \ms{71.8}{1.3} & \ms{74.6}{1.7} & \ms{74.5}{1.6} \\
    RHFF-LVM ($\nu=1.5$) & \ms{67.7}{2.0} & \ms{74.9}{3.2} & \ms{75.1}{2.7} \\
    RHFF-LVM ($\nu=2.5$) & \ms{69.8}{5.9} & \ms{79.8}{3.6} & \ms{80.3}{3.3} \\
    RHFF-LVM (heat) & \cellcolor{lightblue}\textbf{\ms{75.0}{0.9}} & \cellcolor{lightblue}\textbf{\ms{83.7}{1.2}} & \cellcolor{lightblue}\textbf{\ms{84.3}{1.1}} \\
    \bottomrule
  \end{tabular}%
  }
  \vspace{-1ex}
\end{table}

\paragraph{Comparison with Euclidean methods.}
\Cref{Tab:Toy-Results} reports the results.
RHFF-LVM (heat) improves over its Euclidean counterpart RFF-LVM on every metric, most strongly on the ground-truth one (\textbf{tree corr}: $84.3$ vs.\ $74.9$; \textbf{glo}: $83.7$ vs.\ $74.9$; \textbf{loc}: $75.0$ vs.\ $72.2$).
More broadly, no Euclidean method exceeds the tree correlation of $77.2$ attained by PCA.

\paragraph{Comparison with hyperbolic methods.}
Poincar\'e Map is the strongest external baseline here.
RHFF-LVM (heat) leads on two of the three metrics--- on \textbf{loc} ($75.0$ vs.\ $69.4$) and on \textbf{tree corr} ($84.3$ vs.\ $82.4$)---while the two are indistinguishable on \textbf{glo} ($83.7$ vs.\ $83.6$).
Poincar\'e Map embeds proximities precomputed from a neighborhood graph, which remain faithful on this mildly perturbed tree but degrade on noisy, high-dimensional real data (\S~\ref{Subsec:RealData}).
BHGP-LVM consistently collapses to degenerate embeddings.
Ht-SNE exhibits by far the largest variance across seeds (\textbf{glo} $\pm 9.7$).
On this dataset, the heat spectrum dominates among the RHFF-LVM variants across all three metrics, though its margin over the HyLa spectrum is within one standard deviation.
The two orderings of \textbf{glo} and \textbf{tree corr} almost coincide (\Cref{Tab:Toy-Results}), supporting its use as a proxy for hierarchy preservation on the real datasets, where the true tree is unavailable.
\label{Subsec:Toy}

\subsection{Scalability Analysis}
\label{Subsec:Scalability}
\begin{figure}[t]
  \centering
  \includegraphics[width=0.91\linewidth]{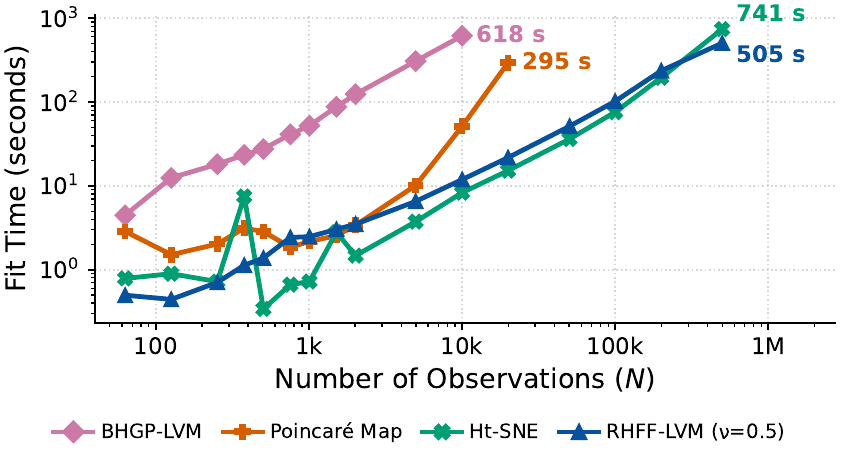}
  \caption{Fit time versus $N$ on the SBT scaling benchmark (log--log, $100$ iterations for all methods). Endpoints mark where each method exceeds the time or memory budget.}
  \label{Fig:Scaling-Time}
  \vspace{-1ex}
\end{figure}

We grow the SBT dataset by increasing only the number of samples per node, so that $N$ alone varies, from $63$ to $500{,}031$.
\Cref{Fig:Scaling-Time} shows the results.
BHGP-LVM does not scale beyond $N \approx 10^{4}$ ($618$\,s), limited by its variational inference, and Poincar\'e Map beyond $N \approx 2 \times 10^{4}$ ($295$\,s), where its dense $\mathcal{O}(N^{2})$ affinity matrix already occupies $25.6$\,GB.
RHFF-LVM ($\nu=0.5$) scales near-linearly, consistent with its $\mathcal{O}(NM^{2}+M^{3})$ complexity, and fits half a million points in $505$ seconds.
Ht-SNE also reaches $N=5\times10^{5}$ ($741$\,s) with smaller memory usage ($3.8$\,GB vs.\ $16.5$\,GB for RHFF-LVM).
It is, however, a neighbor embedding restricted to $Q=2$, and its global scores collapse on every dataset (\S~\ref{Subsec:RealData}).
 
\subsection{Real-World Hierarchical Data}
\label{Subsec:RealData}
\begin{table*}[t]
  \centering
  \caption{Local and global embedding quality at $Q=2$ (seed-wise ranks averaged over datasets and seeds; per-seed deviations in \Cref{Tab:Real-Data-All}). \textbf{loc} denotes $k$NN recall ($k=10$), and \textbf{glo} denotes the Pearson correlation between the pairwise distances in the observed and embedding spaces. Values are $\times100$. The last three columns give the dataset-averaged mean rank on the \textbf{loc} / \textbf{glo} / both metrics (lower is better), computed over the methods available on all datasets; - marks runs exceeding the time or memory budget. \textbf{bold} = best per column, \colorbox{lightblue}{shaded} = best RHFF-LVM variant.}
  \label{Tab:Real-Data-Q2}
  \resizebox{\textwidth}{!}{%
  \begin{tabular}{l|cc|cc|cc|cc|cc|cc|cc|cc||ccc}
    \toprule
    & \multicolumn{2}{c|}{\shortstack{Paul\\{\scriptsize $N{=}2.7k$}}} & \multicolumn{2}{c|}{\shortstack{Plass\\{\scriptsize $N{=}21.6k$}}} & \multicolumn{2}{c|}{\shortstack{Tasic\\{\scriptsize $N{=}23.8k$}}} & \multicolumn{2}{c|}{\shortstack{\emph{C.\,elegans}\\{\scriptsize $N{=}54.6k$}}} & \multicolumn{2}{c|}{\shortstack{MNIST\\{\scriptsize $N{=}60.0k$}}} & \multicolumn{2}{c|}{\shortstack{Zebrafish\\{\scriptsize $N{=}71.2k$}}} & \multicolumn{2}{c|}{\shortstack{Zeisel\\{\scriptsize $N{=}160.8k$}}} & \multicolumn{2}{c|}{\shortstack{MouseDev\\{\scriptsize $N{=}292.5k$}}} & \multicolumn{3}{c}{Mean rank $\downarrow$} \\
    \cmidrule(lr){2-3} \cmidrule(lr){4-5} \cmidrule(lr){6-7} \cmidrule(lr){8-9} \cmidrule(lr){10-11} \cmidrule(lr){12-13} \cmidrule(lr){14-15} \cmidrule(lr){16-17} \cmidrule(lr){18-20}
    Method & loc & glo & loc & glo & loc & glo & loc & glo & loc & glo & loc & glo & loc & glo & loc & glo & loc & glo & both \\
    \midrule
    PCA & 1.5 & 27.9 & 1.1 & 65.9 & 1.1 & \textbf{88.1} & 1.9 & 71.0 & 0.8 & 52.7 & 0.9 & \textbf{92.3} & 0.4 & \textbf{77.2} & 0.5 & \textbf{94.5} & 9.00 & 3.35 & 6.18 \\
    UMAP~\cite{mcinnes2018umap} & 5.3 & 12.4 & \textbf{13.3} & 67.2 & \textbf{14.3} & 49.1 & \textbf{22.4} & 60.6 & \textbf{11.7} & 35.7 & \textbf{13.2} & 79.3 & \textbf{11.6} & 53.1 & \textbf{8.3} & 63.4 & \textbf{1.12} & 7.83 & 4.47 \\
    BGP-LVM~\cite{titsias2010bayesian} & 2.8 & 71.8 & 3.3 & 72.4 & 4.8 & 73.3 & 2.5 & 47.8 & 0.9 & \textbf{67.0} & 0.2 & 10.5 & - & - & - & - & - & - & - \\
    RFF-LVM~\cite{gundersen2021latent} & 2.8 & 69.3 & 9.1 & 85.1 & 6.7 & 73.8 & 6.2 & 74.4 & 3.3 & 41.6 & 3.0 & 84.6 & 1.7 & 72.5 & 0.9 & 86.2 & 4.30 & 4.93 & 4.61 \\
    \midrule
    Poincar\'e Map~\cite{klimovskaia2020poincare} & 4.1 & 13.5 & 10.9 & 78.8 & 13.9 & 47.6 & - & - & - & - & - & - & - & - & - & - & - & - & - \\
    BHGP-LVM~\cite{pmlr-v258-watanabe25a} & 2.5 & 23.2 & 3.0 & 73.9 & 3.2 & 56.9 & - & - & - & - & - & - & - & - & - & - & - & - & - \\
    Ht-SNE~\cite{skrodzki2024accelerating} & \textbf{5.9} & 27.4 & 4.7 & 17.0 & 2.7 & 14.7 & 6.8 & 44.6 & 8.1 & 21.7 & 4.5 & 36.5 & 2.3 & 36.0 & 3.6 & 21.7 & 3.77 & 8.85 & 6.31 \\
    \midrule
    RHFF-LVM (HyLa)~\cite{yu2023random} & 2.1 & 64.1 & 6.2 & \textbf{90.6} & 4.6 & 85.3 & 4.9 & 77.9 & 2.9 & 44.2 & 2.5 & 89.8 & 1.4 & 75.1 & 0.8 & 89.8 & 6.62 & 3.33 & 4.97 \\
    RHFF-LVM ($\nu=0.5$) & \cellcolor{lightblue}2.9 & \cellcolor{lightblue}\textbf{72.0} & \cellcolor{lightblue}10.1 & 85.3 & \cellcolor{lightblue}8.3 & 69.9 & \cellcolor{lightblue}9.3 & 68.0 & \cellcolor{lightblue}3.8 & 45.1 & \cellcolor{lightblue}4.6 & 76.8 & \cellcolor{lightblue}3.5 & 68.0 & \cellcolor{lightblue}1.7 & 69.1 & 2.45 & 5.70 & \textbf{4.08} \\
    RHFF-LVM ($\nu=1.5$) & 2.5 & 65.7 & 7.5 & 89.0 & 6.0 & 78.7 & 6.3 & 73.9 & 3.4 & 43.9 & 3.2 & 81.1 & 1.9 & 72.6 & 1.0 & 82.7 & 4.15 & 4.73 & 4.44 \\
    RHFF-LVM ($\nu=2.5$) & 2.2 & 64.9 & 6.4 & 89.6 & 4.8 & 82.0 & 5.1 & 77.3 & 2.9 & 44.8 & 2.6 & 85.6 & 1.5 & 74.6 & 0.9 & 88.3 & 5.92 & 3.75 & 4.84 \\
    RHFF-LVM (heat) & 2.1 & 64.2 & 5.6 & \cellcolor{lightblue}90.1 & 4.2 & \cellcolor{lightblue}87.9 & 4.7 & \cellcolor{lightblue}\textbf{78.8} & 2.8 & \cellcolor{lightblue}45.3 & 2.4 & \cellcolor{lightblue}90.9 & 1.3 & \cellcolor{lightblue}76.4 & 0.8 & \cellcolor{lightblue}91.1 & 7.65 & \textbf{2.55} & 5.10 \\
    \bottomrule
  \end{tabular}%
  }
\end{table*}

We next test whether these findings generalize to eight real-world datasets.
Seven are single-cell atlases spanning two orders of magnitude in size ($2.7\mathrm{k}$--$292\mathrm{k}$ cells), annotated with either differentiation trees (Paul~\cite{paul2015transcriptional}, Plass~\cite{plass2018cell}, \emph{C.~elegans}~\cite{packer2019lineage}, Zebrafish~\cite{wagner2018single}, MouseDev~\cite{la2021molecular}) or deep cell-type taxonomies (Tasic~\cite{tasic2018shared}, Zeisel~\cite{zeisel2018molecular}).
MNIST~\cite{lecun1998gradient} serves as a non-hierarchical benchmark.
Preprocessing follows a common pipeline, except for Paul and Plass, which we use as distributed by \cite{klimovskaia2020poincare}, and Tasic, which follows \cite{kobak2019art} (details in \S~\ref{App:Datasets}).
For \textbf{glo} we use the Pearson correlation, as the Spearman correlation~\cite{espadoto2019toward} is prohibitively expensive at this scale.

\paragraph{Global structure at scale.}
\Cref{Tab:Real-Data-Q2} reports the results at $Q=2$ (results for $Q \in \{3,5\}$ with std. and all embedding visualizations are in \S~\ref{App:Additional-Results}).
On the global metric, RHFF-LVM (heat) attains the best mean rank among all methods ($2.55$, ahead of PCA at $3.35$), and the best score outright on \emph{C.~elegans} ($78.8$ vs.\ $71.0$ for PCA).
PCA is strong globally but ranks last locally ($9.00$), while UMAP shows the mirror image (\textbf{loc}: $1.12$, \textbf{glo}: $7.83$).
On MNIST, the negative control, the hyperbolic advantage disappears as expected, and the Euclidean BGP-LVM leads ($67.0$).
Moreover, RHFF-LVM (heat) surpasses RHFF-LVM (HyLa) for \textbf{glo}, indicating the benefit of sampling from the spectrum properly weighted by the Plancherel factor $|c(\lambda)|^{-2}$.

\paragraph{Effect of the geometry.}
Since RFF-LVM differs from RHFF-LVM (heat) only in the geometry of the latent space, any difference is attributable to the geometry alone.
Hyperbolic latents improve the global metric on seven of the eight datasets, the exception being Paul ($69.3$ vs.\ $64.2$).
Unlike on the synthetic tree, the gain now comes at the expense of \textbf{loc}, a loss that the choice of spectrum recovers: RHFF-LVM ($\nu=0.5$) exceeds the \textbf{loc} of RFF-LVM on all eight datasets (e.g., $3.5$ vs.\ $1.7$ on Zeisel).

\paragraph{The spectrum traces the local--global trade-off.}
Within RHFF-LVM, the two metrics order the spectra in exactly opposite ways.
The mean \textbf{glo} rank improves monotonically with smoothness ($5.70 \to 4.73 \to 3.75 \to 2.55$ from $\nu=0.5$ to heat), while the mean \textbf{loc} rank improves monotonically with tail weight ($7.65 \to 5.92 \to 4.15 \to 2.45$ from heat to $\nu=0.5$), placing $\nu=0.5$ second only to UMAP locally.
The spectrum controls how strongly nearby pairs are weighted, while the objective remains the same marginal likelihood throughout.
Under equal weighting of the two metrics, RHFF-LVM ($\nu=0.5$) attains the best combined rank ($4.08$).
Moreover, RHFF-LVM ($\nu=0.5$) holds the best combined rank at every dimension ($3.78$ averaged over $Q$).
The results for varying dimensions are given in~\Cref{Tab:Real-Data-All}.

\begin{figure}[t]
  \centering
  \includegraphics[width=0.95\linewidth]{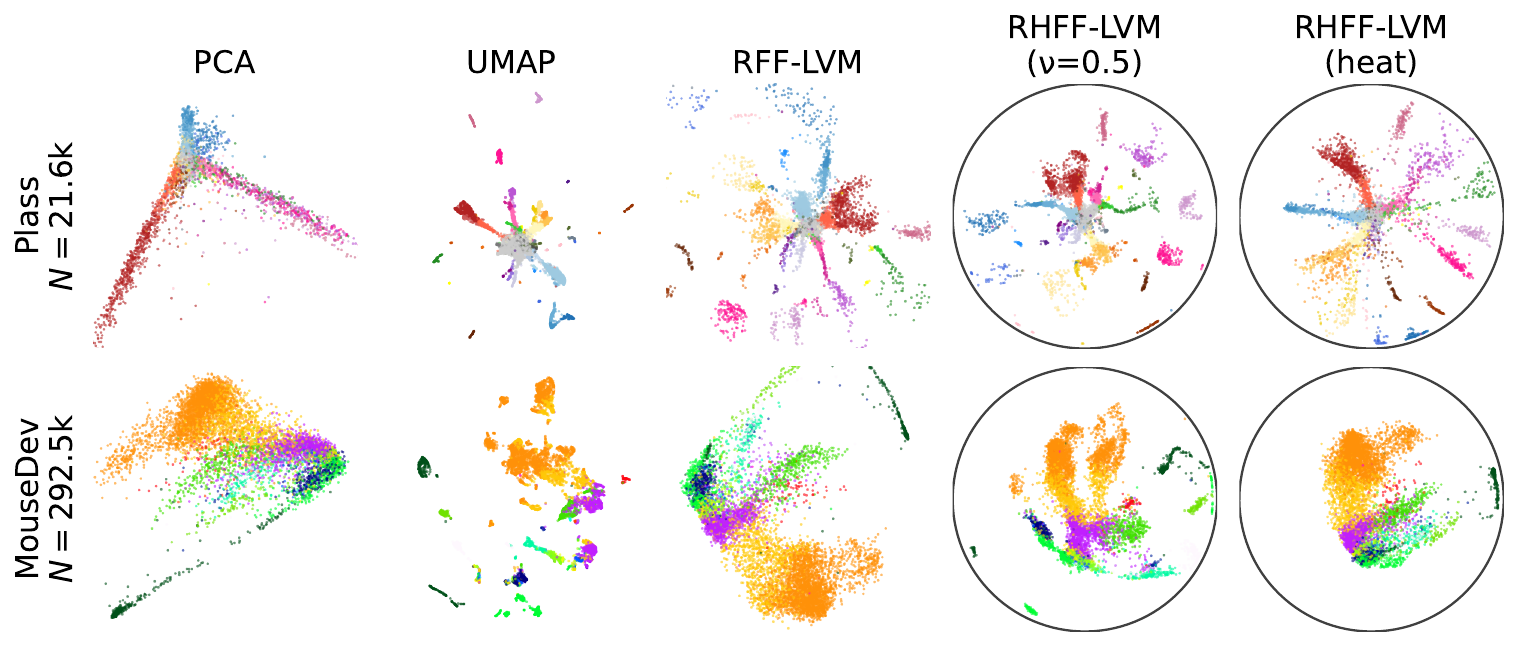}
  \caption{Embeddings at $Q{=}2$ on Plass ($21.6$k) and MouseDev ($292$k), colored by cell type. Hyperbolic methods (RHFF-LVM) are shown on the Poincar\'e ball, Euclidean methods (PCA, UMAP, RFF-LVM) directly.}
  \vspace{-1ex}
  \label{Fig:Visualization-Main}
\end{figure}

\paragraph{Qualitative embeddings.}
\Cref{Fig:Visualization-Main} makes these trends visible on a mid-size lineage atlas (Plass, $21.6$k) and the largest one (MouseDev, $292$k).
PCA flattens the lineage and overlaps its branches, whereas UMAP tears it into disconnected islands, mirroring the global and local extremes seen in \Cref{Tab:Real-Data-Q2}.
RFF-LVM collapses the branches toward the center, whereas switching only the latent space to hyperbolic (RHFF-LVM) unfolds the hierarchy radially.
Within RHFF-LVM, the spectrum sets the operating point: $\nu{=}0.5$ preserves compact, well-separated local clusters, whereas heat emphasizes the continuous global branching.

\paragraph{Statistical analysis.}
Since several margins in \Cref{Tab:Real-Data-Q2} are small, we assess their statistical significance.
We run significance tests on the \textbf{glo} for three contrasts against RHFF-LVM (heat): PCA, which trails by only a small margin; RFF-LVM, as a geometry ablation; and RHFF-LVM (HyLa), which isolates the effect of the Plancherel weighting $|c(\lambda)|^{-2}$ (\Cref{Tab:Key-Results}).
Against PCA, the eight datasets split $3$--$5$ in PCA's favor in head-to-head wins, yet our three wins are large (up to $+36$, all on the tree-structured atlases) while PCA's five are small (all $\le 7.4$).
Against RFF-LVM and RHFF-LVM (HyLa), RHFF-LVM (heat) improves the global metric on seven of the eight datasets (Wilcoxon signed-rank test across datasets: $p=0.11$ and $p=0.02$, respectively).
For the geometry ablation, every within-dataset gap is significant, the sole exception being Paul.
Finally, sweeping the spectrum from $\nu{=}0.5$ to heat moves the local and global metrics monotonically in opposite directions in trend (Page trend test, $p\approx9\times10^{-12}$ for \textbf{loc} and $p\approx3\times10^{-4}$ for \textbf{glo}), quantifying the local--global trade-off.

\section{Conclusion}
\label{Section:Conclusion}
We introduced Random Helgason--Fourier Features, hyperbolic random features whose spectral density is explicitly characterized, and built on them the first hyperbolic non-parametric latent variable model that scales to hundreds of thousands of points.
The RHFF approximation admits a uniform error bound depending on the latent dimension and domain radius.
Empirically, the model attains the best mean rank on global-structure metrics at Q = 2, and the best combined rank at every latent dimension.
\paragraph{Limitations.}
The approximation error grows exponentially with $Q$ and the domain diameter, restricting RHFF to the low-dimensional, bounded regime.
This is where hyperbolic geometry is most useful, but it is a genuine limit.
Peak memory is dominated by the $N\times2M$ feature matrix, and mini-batch training is left to future work.

\section*{AI use statement}
In this work, we used generative AI tools to assist in proving the mathematical claims, including suggesting key steps of the proofs; to translate parts of the manuscript into English; and, partially, to implement the methods: the core implementation of the proposed method was first written by the authors, and generative AI tools were then used to
refactor it and to implement experiment and data-preprocessing scripts under author-specified designs and pipelines.
We have not used generative AI tools to generate data sets, develop the theoretical models or conceptual frameworks, formulate the mathematical claims, propose or refine hypotheses, design the methodology or experiments, or interpret the results, and the remaining tasks requiring disclosure (dataset cleaning and reformatting, and qualitative data
analysis) are not applicable to this work.
Additionally, we used generative AI tools to draft parts of the appendices describing the experimental setup and to edit the manuscript for readability.
We have reviewed all AI-assisted work: all mathematical claims were formulated by the authors and the final proofs were constructed and verified line by line by the authors based on the AI suggestions; AI-assisted code was reviewed and tested by the authors; and all translations and edits were checked against the intended content.
We take responsibility for the final content of this work, including text, claims, or artifacts produced with the aid of generative AI.

\section*{Acknowledgement}
This study was supported in part by JSPS KAKENHI Grant Numbers JP24K02942, JP24KJ0324, JP25H01494, JP26K23861, and JP26K02989.

\bibliographystyle{plain}
\bibliography{ref}

@article{nickel2017poincare,
  title={{Poincar{\'e}} embeddings for learning hierarchical representations},
  author={Nickel, Maximillian and Kiela, Douwe},
  journal={Advances in Neural Information Processing Systems},
  volume={30},
  year={2017}
}

@inproceedings{nickel2018learning,
  title={Learning continuous hierarchies in the {Lorentz} model of hyperbolic geometry},
  author={Nickel, Maximillian and Kiela, Douwe},
  booktitle={Proceedings of the International Conference on Machine Learning},
  pages={3779--3788},
  year={2018}
}

@inproceedings{sarkar2011low,
  title={Low distortion {Delaunay} embedding of trees in hyperbolic plane},
  author={Sarkar, Rik},
  booktitle={Proceedings of the International Symposium on Graph Drawing},
  pages={355--366},
  year={2011}
}

@inproceedings{sala2018representation,
  title={Representation tradeoffs for hyperbolic embeddings},
  author={Sala, Frederic and De Sa, Chris and Gu, Albert and R{\'e}, Christopher},
  booktitle={Proceedings of the International Conference on Machine Learning},
  pages={4460--4469},
  year={2018}
}

@article{azangulov2024stationary,
  title={Stationary Kernels and {Gaussian} Processes on {Lie} Groups and their Homogeneous Spaces {II}: non-compact symmetric spaces},
  author={Azangulov, Iskander and Smolensky, Andrei and Terenin, Alexander and Borovitskiy, Viacheslav},
  journal={Journal of Machine Learning Research},
  volume={25},
  number={281},
  pages={1--51},
  year={2024}
}

@inproceedings{jaquier2024bringing,
  title={Bringing Motion Taxonomies to Continuous Domains via {GPLVM} on Hyperbolic manifolds},
  author={Jaquier, No{\'e}mie and Rozo, Leonel and Gonz{\'a}lez-Duque, Miguel and Borovitskiy, Viacheslav and Asfour, Tamim},
  booktitle={Proceedings of the International Conference on Machine Learning},
  pages={21373--21409},
  year={2024}
}

@InProceedings{pmlr-v258-watanabe25a,
  title = 	 {Hyperboloid {GPLVM} for Discovering Continuous Hierarchies via Nonparametric Estimation},
  author =       {Watanabe, Koshi and Maeda, Keisuke and Ogawa, Takahiro and Haseyama, Miki},
  booktitle = 	 {Proceedings of the International Conference on Artificial Intelligence and Statistics},
  pages = 	 {1783--1791},
  year = 	 {2025}
}

@article{rahimi2007random,
  title={Random features for large-scale kernel machines},
  author={Rahimi, Ali and Recht, Benjamin},
  journal={Advances in Neural Information Processing Systems},
  pages = {1--8},
  volume={20},
  year={2007}
}

@inproceedings{yu2023random,
  title={Random {Laplacian} Features for Learning with Hyperbolic Space},
  author={Yu, Tao and De Sa, Christopher},
  booktitle={Proceedings of the International Conference on Learning Representations},
  pages={1--23},
  year={2023}
}

@book{helgason2022groups,
  title={Groups and geometric analysis: integral geometry, invariant differential operators, and spherical functions},
  author={Helgason, Sigurdur},
  volume={83},
  year={2022},
  publisher={American Mathematical Society}
}

@article{cannon1997hyperbolic,
  title={Hyperbolic geometry},
  author={Cannon, James W and Floyd, William J and Kenyon, Richard and Parry, Walter R},
  journal={Flavors of Geometry},
  volume={31},
  pages={59-115},
  number={2},
  year={1997}
}

@article{paul2015transcriptional,
  title={Transcriptional heterogeneity and lineage commitment in myeloid progenitors},
  author={Paul, Franziska and Arkin, Ya’ara and Giladi, Amir and Jaitin, Diego Adhemar and Kenigsberg, Ephraim and Keren-Shaul, Hadas and Winter, Deborah and Lara-Astiaso, David and Gury, Meital and Weiner, Assaf and others},
  journal={Cell},
  volume={163},
  number={7},
  pages={1663--1677},
  year={2015}
}

@article{mcinnes2018umap,
  title={{UMAP}: Uniform manifold approximation and projection for dimension reduction},
  author={McInnes, Leland and Healy, John and Melville, James},
  journal={arXiv preprint arXiv:1802.03426},
  year={2018}
}

@incollection{rasmussen2003gaussian,
  title={Gaussian processes in machine learning},
  author={Rasmussen, Carl Edward},
  booktitle={Summer School on Machine Learning},
  pages={63--71},
  year={2003},
  publisher={Springer}
}

@inproceedings{sutherland2015error,
  title={On the error of random {Fourier} features},
  author={Sutherland, Danica J and Schneider, Jeff},
  booktitle={Proceedings of the Thirty-First Conference on Uncertainty in Artificial Intelligence},
  pages={862--871},
  year={2015}
}

@inproceedings{nagano2019wrapped,
  title={A wrapped normal distribution on hyperbolic space for gradient-based learning},
  author={Nagano, Yoshihiro and Yamaguchi, Shoichiro and Fujita, Yasuhiro and Koyama, Masanori},
  booktitle={Proceedings of the International Conference on Machine Learning},
  pages={4693--4702},
  year={2019}
}

@article{cho2022rotated,
  title={A rotated hyperbolic wrapped normal distribution for hierarchical representation learning},
  author={Cho, Seunghyuk and Lee, Juyong and Park, Jaesik and Kim, Dongwoo},
  journal={Advances in Neural Information Processing Systems},
  volume={35},
  pages={17831--17843},
  year={2022}
}

@article{klimovskaia2020poincare,
  title={Poincar{\'e} maps for analyzing complex hierarchies in single-cell data},
  author={Klimovskaia, Anna and Lopez-Paz, David and Bottou, L{\'e}on and Nickel, Maximilian},
  journal={Nature Communications},
  volume={11},
  number={1},
  pages={2966},
  year={2020},
  publisher={Nature Publishing Group UK London}
}

@article{plass2018cell,
  title={Cell type atlas and lineage tree of a whole complex animal by single-cell transcriptomics},
  author={Plass, Mireya and Solana, Jordi and Wolf, F Alexander and Ayoub, Salah and Misios, Aristotelis and Gla{\v{z}}ar, Petar and Obermayer, Benedikt and Theis, Fabian J and Kocks, Christine and Rajewsky, Nikolaus},
  journal={Science},
  volume={360},
  number={6391},
  year={2018},
  publisher={American Association for the Advancement of Science}
}

@inproceedings{lalchand2022generalised,
  title={Generalised {GPLVM} with stochastic variational inference},
  author={Lalchand, Vidhi and Ravuri, Aditya and Lawrence, Neil D},
  booktitle={International Conference on Artificial Intelligence and Statistics},
  pages={7841--7864},
  year={2022},
  organization={PMLR}
}

@article{kobak2019art,
  title={The art of using t-{SNE} for single-cell transcriptomics},
  author={Kobak, Dmitry and Berens, Philipp},
  journal={Nature Communications},
  volume={10},
  number={1},
  pages={5416},
  year={2019},
  publisher={Nature Publishing Group UK London}
}

@article{espadoto2019toward,
  title={Toward a quantitative survey of dimension reduction techniques},
  author={Espadoto, Mateus and Martins, Rafael M and Kerren, Andreas and Hirata, Nina ST and Telea, Alexandru C},
  journal={IEEE Transactions on Visualization and Computer Graphics},
  volume={27},
  number={3},
  pages={2153--2173},
  year={2019},
  publisher={IEEE}
}

@article{lawrence2005probabilistic,
  title={Probabilistic non-linear principal component analysis with {Gaussian} process latent variable models.},
  author={Lawrence, Neil},
  journal={Journal of Machine Learning Research},
  volume={6},
  number={11},
  year={2005}
}

@inproceedings{titsias2010bayesian,
  title={Bayesian {Gaussian} process latent variable model},
  author={Titsias, Michalis and Lawrence, Neil D},
  booktitle={Proceedings of the International Conference on Artificial Intelligence and Statistics},
  pages={844--851},
  year={2010}
}

@article{van2008visualizing,
  title={Visualizing data using t-{SNE}.},
  author={Van der Maaten, Laurens and Hinton, Geoffrey},
  journal={Journal of Machine Learning Research},
  volume={9},
  number={11},
  pages={2579--2605},
  year={2008}
}

@article{kruskal1964multidimensional,
  title={Multidimensional scaling by optimizing goodness of fit to a nonmetric hypothesis},
  author={Kruskal, Joseph B},
  journal={Psychometrika},
  volume={29},
  number={1},
  pages={1--27},
  year={1964},
  publisher={Cambridge University Press}
}

@article{saitou1987neighbor,
  title={The neighbor-joining method: a new method for reconstructing phylogenetic trees.},
  author={Saitou, Naruya and Nei, Masatoshi},
  journal={Molecular biology and evolution},
  volume={4},
  number={4},
  pages={406--425},
  year={1987}
}

@article{skrodzki2024accelerating,
  title={Accelerating hyperbolic t-{SNE}},
  author={Skrodzki, Martin and van Geffen, Hunter and Chaves-de-Plaza, Nicolas F and H{\"o}llt, Thomas and Eisemann, Elmar and Hildebrandt, Klaus},
  journal={IEEE Transactions on Visualization and Computer Graphics},
  volume={30},
  number={7},
  pages={4403--4415},
  year={2024},
  publisher={IEEE}
}

@inproceedings{li2024preventing,
title={Preventing Model Collapse in {Gaussian} Process Latent Variable Models},
author={Ying Li and Zhidi Lin and Feng Yin and Michael Minyi Zhang},
booktitle={Proceedings of the International Conference on Machine Learning},
year={2024},
url={https://openreview.net/forum?id=4byOXWrJay}
}

@article{kury2025dreams,
title={{DREAMS}: Preserving both Local and Global Structure in Dimensionality Reduction},
author={No{\"e}l Kury and Dmitry Kobak and Sebastian Damrich},
journal={Transactions on Machine Learning Research},
issn={2835-8856},
year={2026}
}

@article{wang2021understanding,
  title={Understanding how dimension reduction tools work: an empirical approach to deciphering {t-SNE, UMAP, TriMAP, and PaCMAP} for data visualization},
  author={Wang, Yingfan and Huang, Haiyang and Rudin, Cynthia and Shaposhnik, Yaron},
  journal={Journal of Machine Learning Research},
  volume={22},
  number={201},
  pages={1--73},
  year={2021}
}

@article{moon2019visualizing,
  title={Visualizing structure and transitions in high-dimensional biological data},
  author={Moon, Kevin R and Van Dijk, David and Wang, Zheng and Gigante, Scott and Burkhardt, Daniel B and Chen, William S and Yim, Kristina and Elzen, Antonia van den and Hirn, Matthew J and Coifman, Ronald R and others},
  journal={Nature Biotechnology},
  volume={37},
  number={12},
  pages={1482--1492},
  year={2019},
  publisher={Nature Publishing Group US New York}
}

@inproceedings{lawrence2006local,
  title={Local distance preservation in the {GP-LVM} through back constraints},
  author={Lawrence, Neil D and Quinonero-Candela, Joaquin},
  booktitle={Proceedings of the International Conference on Machine Learning},
  pages={513--520},
  year={2006}
}

@article{moreno2022revisiting,
  title={Revisiting active sets for {Gaussian} process decoders},
  author={Moreno-Mu{\~n}oz, Pablo and Feldager, Cilie and Hauberg, S{\o}ren},
  journal={Advances in Neural Information Processing Systems},
  volume={35},
  pages={6603--6614},
  year={2022}
}

@inproceedings{gundersen2021latent,
  title={Latent variable modeling with random features},
  author={Gundersen, Gregory and Zhang, Michael and Engelhardt, Barbara},
  booktitle={International Conference on Artificial Intelligence and Statistics},
  pages={1333--1341},
  year={2021},
  organization={PMLR}
}

@inproceedings{mishne2023numerical,
  title={The numerical stability of hyperbolic representation learning},
  author={Mishne, Gal and Wan, Zhengchao and Wang, Yusu and Yang, Sheng},
  booktitle={Proceedings of the International Conference on Machine Learning},
  pages={24925--24949},
  year={2023},
  organization={PMLR}
}

@article{packer2019lineage,
  title={A lineage-resolved molecular atlas of {C}. elegans embryogenesis at single-cell resolution},
  author={Packer, Jonathan S and Zhu, Qin and Huynh, Chau and Sivaramakrishnan, Priya and Preston, Elicia and Dueck, Hannah and Stefanik, Derek and Tan, Kai and Trapnell, Cole and Kim, Junhyong and others},
  journal={Science},
  volume={365},
  number={6459},
  pages={eaax1971},
  year={2019},
  publisher={American Association for the Advancement of Science}
}

@article{wagner2018single,
  title={Single-cell mapping of gene expression landscapes and lineage in the zebrafish embryo},
  author={Wagner, Daniel E and Weinreb, Caleb and Collins, Zach M and Briggs, James A and Megason, Sean G and Klein, Allon M},
  journal={Science},
  volume={360},
  number={6392},
  pages={981--987},
  year={2018},
  publisher={American Association for the Advancement of Science}
}

@article{la2021molecular,
  title={Molecular architecture of the developing mouse brain},
  author={La Manno, Gioele and Siletti, Kimberly and Furlan, Alessandro and Gyllborg, Daniel and Vinsland, Elin and Mossi Albiach, Alejandro and Mattsson Langseth, Christoffer and Khven, Irina and Lederer, Alex R and Dratva, Lisa M and others},
  journal={Nature},
  volume={596},
  number={7870},
  pages={92--96},
  year={2021},
  publisher={Nature Publishing Group UK London}
}

@article{tasic2018shared,
  title={Shared and distinct transcriptomic cell types across neocortical areas},
  author={Tasic, Bosiljka and Yao, Zizhen and Graybuck, Lucas T and Smith, Kimberly A and Nguyen, Thuc Nghi and Bertagnolli, Darren and Goldy, Jeff and Garren, Emma and Economo, Michael N and Viswanathan, Sarada and others},
  journal={Nature},
  volume={563},
  number={7729},
  pages={72--78},
  year={2018},
  publisher={Nature Publishing Group UK London}
}

@article{zeisel2018molecular,
  title={Molecular architecture of the mouse nervous system},
  author={Zeisel, Amit and Hochgerner, Hannah and L{\"o}nnerberg, Peter and Johnsson, Anna and Memic, Fatima and Van Der Zwan, Job and H{\"a}ring, Martin and Braun, Emelie and Borm, Lars E and La Manno, Gioele and others},
  journal={Cell},
  volume={174},
  number={4},
  pages={999--1014},
  year={2018},
  publisher={Elsevier}
}

@article{lecun1998gradient,
  title={Gradient-based learning applied to document recognition},
  author={LeCun, Yann and Bottou, L{\'e}on and Bengio, Yoshua and Haffner, Patrick},
  journal={Proceedings of the IEEE},
  volume={86},
  number={11},
  pages={2278--2324},
  year={1998},
  publisher={Ieee}
}

@inproceedings{chami2021horopca,
  title={{HoroPCA}: Hyperbolic dimensionality reduction via horospherical projections},
  author={Chami, Ines and Gu, Albert and Nguyen, Dat P and R{\'e}, Christopher},
  booktitle={Proceedings of the International Conference on Machine Learning},
  pages={1419--1429},
  year={2021},
  organization={PMLR}
}

@inproceedings{guo2022co,
  title={{Co-SNE}: Dimensionality reduction and visualization for hyperbolic data},
  author={Guo, Yunhui and Guo, Haoran and Yu, Stella X},
  booktitle={Proceedings of the IEEE/CVF Conference on Computer Vision and Pattern Recognition},
  pages={21--30},
  year={2022}
}

@article{yu2019numerically,
  title={Numerically accurate hyperbolic embeddings using tiling-based models},
  author={Yu, Tao and De Sa, Christopher},
  journal={Advances in Neural Information Processing Systems},
  volume={32},
  paages={1-11},
  year={2019}
}

@article{yu2021representing,
  title={Representing hyperbolic space accurately using multi-component floats},
  author={Yu, Tao and De Sa, Christopher M},
  journal={Advances in Neural Information Processing Systems},
  volume={34},
  pages={15570--15581},
  year={2021}
}


\clearpage
\onecolumn
\appendix

\thispagestyle{empty}

\section{Proof of Proposition \ref{Prop:Discrete-Spherical-Func}}
\label{Section:Proof-of-Discrete-Spherical-Func}
Starting from the Harish-Chandra integral representation of the elementary spherical function, we have:
\begin{align*}
    \phi_{\lambda}(r)&=\int_{\partial\mathcal{L}^{Q}}\exp\Big\{(i\lambda+\rho)B(\mathbf{x},\bm{\xi})\Big\}\exp\Big\{(-i\lambda+\rho)B(\mathbf{x}',\bm{\xi})\Big\}d\bm{\xi} \\
    &= \int_{\partial\mathcal{L}^{Q}}\exp\Big\{\rho B(\mathbf{x},\bm{\xi}) + \rho B(\mathbf{x}',\bm{\xi})\Big\}\exp{\Big\{i\lambda B(\mathbf{x},\bm{\xi})-i\lambda B(\mathbf{x}',\bm{\xi})\Big\}}d\bm{\xi}.
\end{align*}
Since the elementary spherical function $\phi_{\lambda}(r)$ is real-valued, we can express the complex exponential through its real part:
\begin{align*}
    \phi_{\lambda}(r) &= \int_{\partial\mathcal{L}^{Q}}\exp{\Big\{\rho B(\mathbf{x},\bm{\xi}) + \rho B(\mathbf{x}',\bm{\xi})\Big\}}\cos\Big\{\lambda B(\mathbf{x},\bm{\xi})-\lambda B(\mathbf{x}',\bm{\xi})\Big\}d\bm{\xi}.
\end{align*}
Applying the trigonometric identity $\cos(A - B) = \cos A \cos B + \sin A \sin B$, we obtain:
\begin{align*}
    \phi_{\lambda}(r) &= \int_{\partial\mathcal{L}^{Q}}\exp{\Big\{\rho B(\mathbf{x},\bm{\xi}) + \rho B(\mathbf{x}',\bm{\xi})\Big\}} \Big\{\cos\lambda B(\mathbf{x},\bm{\xi})\cos\lambda B(\mathbf{x}',\bm{\xi}) \\
    &\quad +\sin\lambda B(\mathbf{x},\bm{\xi})\sin\lambda B(\mathbf{x}',\bm{\xi})\Big\}d\bm{\xi}.
\end{align*}
Finally, by defining the feature map as $\psi_{\lambda,\bm{\xi}}(\mathbf{x})=[\exp{\rho B(\mathbf{x},\bm{\xi})}\cos\lambda B(\mathbf{x},\bm{\xi}),\exp{\rho B(\mathbf{x},\bm{\xi})}\sin\lambda B(\mathbf{x},\bm{\xi})]^{\top}$, the integral can be rewritten as the inner product of these features:
\begin{align*}
    \phi_{\lambda}(r) &= \int_{\partial\mathcal{L}^{Q}} \psi_{\lambda,\bm{\xi}}(\mathbf{x})^{\top}\psi_{\lambda,\bm{\xi}}(\mathbf{x}')d\bm{\xi}.
\end{align*}

\section{Connection to HyLa}
\label{Section:Connection-to-HyLa}
 
The key novelty over HyLa~\cite{yu2023random} is the explicit characterization of $p(\lambda)$~\cite{azangulov2024stationary}. 
Exposing the density allows any isotropic kernel
with a known spherical transform to be specified directly in the frequency domain, as in~\Cref{Eq:Heat-Kernel,Eq:Matern-Kernel}. 
This relationship can be made exact. 
 
\begin{corollary}[RHFF recovers HyLa]
\label{cor:hyla}
Let $\eta_{\lambda, \boldsymbol{\xi}, b}(\mathbf{x}) = \sqrt{2}\,
e^{\rho B(\mathbf{x}, \boldsymbol{\xi})} \cos( \lambda B(\mathbf{x},
\boldsymbol{\xi}) + b )$ with $b \sim \mathrm{Unif}[0, 2\pi)$, which
under the identification of Lemma~\ref{lem:busemann-horocycle} (stated below) is the HyLa feature of~\cite{yu2023random} up to normalization. Then, for
all $\mathbf{x}, \mathbf{x}' \in \mathcal{L}^Q$,
\begin{equation*}
  \mathbb{E}_{b}\!\left[ \eta_{\lambda, \boldsymbol{\xi}, b}(\mathbf{x})\,
  \eta_{\lambda, \boldsymbol{\xi}, b}(\mathbf{x}') \right]
  = \psi_{\lambda, \boldsymbol{\xi}}(\mathbf{x})^\top
    \psi_{\lambda, \boldsymbol{\xi}}(\mathbf{x}')
  = g_{\lambda, \boldsymbol{\xi}}(\mathbf{x}, \mathbf{x}'),
\end{equation*}
and with $\lambda_j \sim \mathcal{N}(0, s^2)$, $\boldsymbol{\xi}_j$
uniform on $\partial\mathcal{L}^Q$, and $b_j$ uniform,
$\frac{1}{M} \sum_{j} \eta_j(\mathbf{x})\, \eta_j(\mathbf{x}')$ is an
unbiased estimator of $k_s(r) = \mathbb{E}_{\lambda \sim \mathcal{N}(0,
s^2)}[\phi_{|\lambda|}(r)]$, the kernel approximated by HyLa. Hence
HyLa is the instance of RHFF obtained by (a) the Gaussian choice
$p(\lambda) = \mathcal{N}(0, s^2)$ and (b) replacing the paired feature
of~\Cref{Eq:RHFF-Mapping} by its phase-randomized counterpart.
\end{corollary}

The remainder of this section proves Corollary~\ref{cor:hyla}. 
We first fix the dictionary between our notation and that of~\cite{yu2023random}: their dimension $n$ is our $Q$, their $\frac{n-1}{2}$ is our $\rho$, their feature count $D$ is our $M$, and their boundary direction $\boldsymbol{\omega} \in \partial\mathcal{B}^n$ corresponds to our ideal point $\boldsymbol{\xi} = [1, \boldsymbol{\omega}^\top]^\top \in
\partial\mathcal{L}^Q$. 
The following lemma identifies the arguments of the two feature maps.
 
\begin{lemma}[Identification of the feature arguments]
\label{lem:busemann-horocycle}
Let $\mathbf{x} \in \mathcal{L}^Q$ and let $\mathbf{z} = \mathbf{x}_s /
(1 + x_t) \in \mathcal{B}^Q$ be its image under the diffeomorphism of Section~\ref{Section:Background}. 
Identify $\boldsymbol{\xi} = [1,\boldsymbol{\omega}^\top]^\top \in \partial\mathcal{L}^Q$ with $\boldsymbol{\omega} \in \mathbb{S}^{Q-1} = \partial\mathcal{B}^Q$.
Then the Busemann function of~\Cref{eq:Busemann-Func} equals the signed horocycle
distance $\langle \boldsymbol{\omega}, \mathbf{z} \rangle_H$ of~\cite{yu2023random}:
\begin{equation*}
  B(\mathbf{x}, \boldsymbol{\xi})
  = \log\!\left( \frac{1 - \|\mathbf{z}\|_2^2}{\|\mathbf{z} -
    \boldsymbol{\omega}\|_2^2} \right)
  = \langle \boldsymbol{\omega}, \mathbf{z} \rangle_H .
\end{equation*}
\end{lemma}
 
\begin{proof}
The inverse of the diffeomorphism is $x_t = (1 +
\|\mathbf{z}\|_2^2)/(1 - \|\mathbf{z}\|_2^2)$ and $\mathbf{x}_s =
2\mathbf{z}/(1 - \|\mathbf{z}\|_2^2)$. Using $\|\boldsymbol{\omega}\|_2
= 1$,
\begin{equation*}
  -\langle \mathbf{x}, \boldsymbol{\xi} \rangle_{\mathcal{L}}
  = x_t - \mathbf{x}_s^\top \boldsymbol{\omega}
  = \frac{1 + \|\mathbf{z}\|_2^2 - 2\,\mathbf{z}^\top
    \boldsymbol{\omega}}{1 - \|\mathbf{z}\|_2^2}
  = \frac{\|\mathbf{z} - \boldsymbol{\omega}\|_2^2}{1 -
    \|\mathbf{z}\|_2^2},
\end{equation*}
and taking $-\log(\cdot)$ of both sides yields the claim.
\end{proof}
 
Under Lemma~\ref{lem:busemann-horocycle}, the HyLa eigenfunction
(Equation~(1) of~\cite{yu2023random}) reads $\mathrm{HyLa}_{\lambda, b,
\boldsymbol{\xi}}(\mathbf{x}) = e^{\rho B(\mathbf{x}, \boldsymbol{\xi})} \cos( \lambda B(\mathbf{x}, \boldsymbol{\xi}) + b )$, so that $\eta_{\lambda, \boldsymbol{\xi}, b} = \sqrt{2}\,
\mathrm{HyLa}_{\lambda, b, \boldsymbol{\xi}}$.
\footnote{The factor $\sqrt{2}$ normalizes the phase average, as in~\cite{rahimi2007random}; \cite{yu2023random} omits it and instead carries the resulting factor $\tfrac{1}{2}$ in the kernel (Theorem~4.1 of~\cite{yu2023random}).}
 
\begin{proof}[Proof of Corollary~\ref{cor:hyla}]
By the product-to-sum identity, with $B = B(\mathbf{x},
\boldsymbol{\xi})$ and $B' = B(\mathbf{x}', \boldsymbol{\xi})$,
\begin{equation*}
  2 \cos(\lambda B + b) \cos(\lambda B' + b)
  = \cos\!\big( \lambda (B - B') \big)
  + \cos\!\big( \lambda (B + B') + 2b \big).
\end{equation*}
The second term integrates to zero over $b \sim \mathrm{Unif}[0,
2\pi)$, hence
\begin{equation*}
  \mathbb{E}_{b}\!\left[ \eta_{\lambda, \boldsymbol{\xi},
  b}(\mathbf{x})\, \eta_{\lambda, \boldsymbol{\xi}, b}(\mathbf{x}')
  \right]
  = e^{\rho (B + B')} \cos\!\big( \lambda (B - B') \big)
  = g_{\lambda, \boldsymbol{\xi}}(\mathbf{x}, \mathbf{x}'),
\end{equation*}
and the identity $\psi_{\lambda, \boldsymbol{\xi}}(\mathbf{x})^\top
\psi_{\lambda, \boldsymbol{\xi}}(\mathbf{x}') = e^{\rho(B + B')}
\cos(\lambda(B - B')) = g_{\lambda, \boldsymbol{\xi}}(\mathbf{x},
\mathbf{x}')$ is the trigonometric expansion of Section~\ref{Section:Proof-of-Discrete-Spherical-Func}. 
For the second statement, the tower property gives
\begin{equation*}
  \mathbb{E}_{\lambda, \boldsymbol{\xi}, b}\!\left[
  \eta(\mathbf{x})\, \eta(\mathbf{x}') \right]
  = \mathbb{E}_{\lambda}\!\left[ \mathbb{E}_{\boldsymbol{\xi}}\!\left[
    g_{\lambda, \boldsymbol{\xi}}(\mathbf{x}, \mathbf{x}') \right]
  \right]
  = \mathbb{E}_{\lambda \sim \mathcal{N}(0, s^2)}\!\left[
    \phi_{|\lambda|}(r) \right]
  = k_s(r),
\end{equation*}
where the middle equality is Proposition~\ref{Prop:Discrete-Spherical-Func} and $r =
d_{\mathcal{L}}(\mathbf{x}, \mathbf{x}')$; the restriction to
$|\lambda|$ is immaterial since $\phi_\lambda = \phi_{-\lambda}$, so
sampling $\lambda$ from $\mathcal{N}(0, s^2)$ on $\mathbb{R}$ is
equivalent to sampling $|\lambda|$ from the corresponding half-normal
density on $\mathbb{R}_+$, matching the convention of Section~\ref{Section:Background}.
That $k_s$ is the kernel approximated by HyLa is Equation~(2) of~\cite{yu2023random}
combined with Lemma~\ref{lem:busemann-horocycle}.
\end{proof}
 
\begin{remark}[RHFF Rao--Blackwellizes HyLa]
\label{rem:rao-blackwell}
The first display of Corollary~\ref{cor:hyla} states that the RHFF
per-sample estimator $g_{\lambda_j, \boldsymbol{\xi}_j}$ is the
conditional expectation of the HyLa per-sample estimator given
$(\lambda_j, \boldsymbol{\xi}_j)$. 
By the Rao--Blackwell theorem, every entry of the RHFF Gram matrix has variance no larger than that of the HyLa Gram matrix built from the same $M$ samples $\{\lambda_j, \boldsymbol{\xi}_j\}_{j=1}^M$. 
The two Gram matrices coincide after marginalizing the phases $\{b_j\}$, not pathwise. 
Note the feature-dimension accounting: $M$ samples yield $2M$ RHFF coordinates but $M$ HyLa coordinates, so the comparison at equal sample count favors RHFF in dimension, while the comparison at equal dimension ($M$ RHFF samples versus $2M$ HyLa samples) mirrors the Euclidean analysis of the two random-feature variants in~\cite{sutherland2015error}.
\end{remark}
 
\begin{remark}[The spectrum implicit in HyLa's Gaussian sampling]
\label{rem:implicit-spectrum}
Inverting $p(\lambda) \propto \tilde{k}(\lambda) |c(\lambda)|^{-2}$, the Gaussian sampling of Corollary~\ref{cor:hyla} implicitly specifies the spherical transform
\begin{equation*}
  \tilde{k}_{\mathrm{HyLa}}(\lambda)
  \;\propto\;
  e^{-\lambda^2 / (2 s^2)}\, |c(\lambda)|^{2},
\end{equation*}
which differs from the heat spectrum of~\Cref{Eq:Heat-Kernel} by the inverse of the Plancherel weight. 
Relative to the heat kernel with $t = 1/(2s^2)$, HyLa thus over-weights low frequencies---for $Q = 2$, $|c(\lambda)|^{2} \propto (\lambda \tanh(\pi\lambda))^{-1}$ diverges as $\lambda \to 0$---and the discrepancy grows with $Q$, since $|c(\lambda)|^{-2}$ behaves as a polynomial of degree $Q - 1$ for large $\lambda$.
\footnote{The induced kernel $k_s$ remains well-defined: the
inverse transform integrates $\phi_\lambda(r)\, p(\lambda)$ directly, and $p$ is Gaussian.} 
Exact agreement with RHFF (heat), in the sense of equal expected Gram matrices, therefore requires sampling $\lambda$ from $p(\lambda) \propto e^{-t\lambda^2} |c(\lambda)|^{-2}$ rather than
from a plain Gaussian. 
This makes precise the sense in which the
Gaussian choice of~\cite{yu2023random} ``corresponds closely to a heat kernel'': it matches the Gaussian factor of the heat spectrum while omitting the Plancherel measure, and exposing $p(\lambda)$ is exactly what restores the correspondence.
\end{remark}

\section{Gradient Computation}
\label{Section:Gradient-Computation}

The Euclidean gradient with respect to $\mathbf{x}_{s}$ is derived using the chain rule. 
First, letting $\mathbf{K} = \mathbf{\Psi}\mathbf{\Psi}^{\top}$, the derivative with respect to the random feature matrix $\mathbf{\Psi}$ is obtained as follows:
\begin{align}
    \frac{\partial \ell}{\partial \mathbf{\Psi}} &= D (\mathbf{K}+\beta^{-1}\mathbf{I})^{-1}\mathbf{\Psi} - (\mathbf{K}+\beta^{-1}\mathbf{I})^{-1}\mathbf{Y}\mathbf{Y}^{\top}(\mathbf{K}+\beta^{-1}\mathbf{I})^{-1}\mathbf{\Psi}. \label{Eq:Nabla_Psi_GP-Likelihood}
\end{align}
To obtain the gradient with respect to the spatial components, we simply need to compute the derivative of the elements of $\mathbf{\Psi}$, which reduces to evaluating $\frac{\partial \psi_{\lambda_{j},\bm{\xi}_{j}}(\mathbf{x})}{\partial \mathbf{x}_{s}}$. 
For notational simplicity, we differentiate the complex representation of the random feature, denoted by $\acute{\psi}_{\lambda_{j},\bm{\xi}_{j}}(\mathbf{x})$:
\begin{align}
    \acute{\psi}_{\lambda_{j},\bm{\xi}_{j}}(\mathbf{x}) &= \frac{1}{\sqrt{M}}\exp\left( (\rho+i \lambda_{j})B(\mathbf{x},\bm{\xi}_{j}) \right), \quad B(\mathbf{x},\bm{\xi}_{j}) = -\log(x_{t}-\mathbf{x}_{s}^{\top}\bm{\xi}_{j}'). \label{Eq:Complex-Phi}
\end{align}
Applying the chain rule yields the following derivatives:
\begin{align*}
    \frac{\partial \acute{\psi}_{\lambda_{j},\bm{\xi}_{j}}(\mathbf{x})}{\partial \mathbf{x}_{s}} &= (\rho + i\lambda_{j}) \acute{\psi}_{\lambda_{j},\bm{\xi}_{j}}(\mathbf{x}) \frac{\partial B(\mathbf{x},\bm{\xi}_{j})}{\partial \mathbf{x}_{s}}, \\
    \frac{\partial B(\mathbf{x},\bm{\xi}_{j})}{\partial \mathbf{x}_{s}} &= \frac{1}{x_{t}-\mathbf{x}_{s}^{\top}\bm{\xi}_{j}'} \cdot \frac{\partial}{\partial \mathbf{x}_{s}}\left(\mathbf{x}_{s}^{\top}\bm{\xi}_{j}'-x_{t}\right) = \frac{1}{x_{t}-\mathbf{x}_{s}^{\top}\bm{\xi}'_{j}} \left(\bm{\xi}'_{j} - \frac{\mathbf{x}_{s}}{x_{t}}\right).
\end{align*}
Substituting the latter into the former, we obtain:
\begin{align}
    \frac{\partial \acute{\psi}_{\lambda_{j},\bm{\xi}_{j}}(\mathbf{x})}{\partial \mathbf{x}_{s}} &= (\rho+i \lambda_{j})\frac{\acute{\psi}_{\lambda_{j},\bm{\xi}_{j}}(\mathbf{x})}{x_{t}-\mathbf{x}_{s}^{\top}\bm{\xi}_{j}'}\left(\bm{\xi}_{j}' - \frac{\mathbf{x}_{s}}{x_{t}}\right). \label{Eq:Derivative-of-Psi}
\end{align}
Finally, by extracting the real and imaginary parts of this result, the Euclidean gradient with respect to the spatial components can be efficiently evaluated.
For efficient computation of the gradients, we apply the following matrix lemma:
\begin{align}
    |\mathbf{\Psi}\mathbf{\Psi}^{\top}+\beta^{-1}\mathbf{I}| &= \beta^{-N}|\mathbf{I}+\beta\mathbf{\Psi}^{\top}\mathbf{\Psi}|, \label{Eq:Det-Lemma} \\ 
    (\mathbf{\Psi}\mathbf{\Psi}^{\top}+\beta^{-1}\mathbf{I})^{-1} &= \beta\mathbf{I}-\beta^{2}\mathbf{\Psi}(\mathbf{I}+\beta\mathbf{\Psi}^{\top}\mathbf{\Psi})^{-1}\mathbf{\Psi}^{\top}. \label{Eq:Inv-Lemma}
\end{align}

\section{Algorithm}
\label{Section:Algorithm}

\begin{algorithm}[t]
\caption{RHFF-LVM: hyperbolic latent-variable dimensionality reduction}
\label{Alg:RHFF-LVM}
\begin{algorithmic}[1]
\Require data $\mathbf{Y}\in\mathbb{R}^{N\times D}$, features $M$, latent dim $Q$,
         kernel hyperparameters, epochs $T$, prior weight $\gamma$, patience $P$
\Statex \textbf{Output:} latents $\mathbf{X}\in(\mathcal{L}^{Q})^{N}$ and their
        Poincar\'e-ball coordinates $\{\mathbf{p}_n\}$ for visualization
\State $\psi \gets \Call{ConstructRHFF}{M,Q,\text{kernel}}$ \Comment{Alg.~\ref{Alg:ConstructRHFF}}
\State initialize $\mathbf{x}_{n,s}\sim\mathrm{Unif}(0,10^{-3})^{Q}$ and
       center $\mathbf{x}_{n,s}\!\mathrel{-}=\!\frac1N\sum_m \mathbf{x}_{m,s}$
       \Comment{latents near the origin $\mathbf{o}$}
\State $\beta\gets\beta_0$;\; $\eta\gets\eta_0$;\;
       $\ell^{\star}_{\mathrm{LOO}}\gets-\infty$;\; $c\gets0$
\For{$\tau=1$ \textbf{to} $T$}
  \State $(\mathbf{X}_s,\beta,\eta)\gets
         \Call{NaturalGradientStep}{\mathbf{X}_s,\beta,\eta,\mathbf{Y},\psi,\gamma}$
         \Comment{Alg.~\ref{Alg:Step}}
  \State $\ell_{\mathrm{LOO}}\gets\sum_{n}\log p(\mathbf{y}_n\mid\mathbf{Y}_{-n})$
         \Comment{closed-form GP leave-one-out log predictive likelihood}
  \If{$\ell_{\mathrm{LOO}}>\ell^{\star}_{\mathrm{LOO}}$}
     \State $\ell^{\star}_{\mathrm{LOO}}\gets\ell_{\mathrm{LOO}}$;\;
            $\mathbf{X}_s^{\star}\gets\mathbf{X}_s$;\; $c\gets0$
  \Else
     \State $c\gets c+1$;\; \textbf{if } $c\ge P$ \textbf{ then break}
            \Comment{early stopping on a plateau}
  \EndIf
\EndFor
\State $\mathbf{x}_n\gets\mathcal{P}(\mathbf{x}_{n,s}^{\star})$ for all $n$
       \Comment{best-LOO latents, lifted to the Lorentz model}
\State $\mathbf{p}_n\gets \mathbf{x}_{n,s}/(1+x_{n,t})$ for all $n$
       \Comment{diffeomorphism $\mathcal{L}^Q\!\to$ Poincar\'e ball for visualization}
\State \textbf{return} $\mathbf{X}=\{\mathbf{x}_n\}$,\; $\{\mathbf{p}_n\}$
\end{algorithmic}
\end{algorithm}

\begin{algorithm}[t]
\caption{\textsc{ConstructRHFF}: sample and fix the feature map}
\label{Alg:ConstructRHFF}
\begin{algorithmic}[1]
\Require number of features $M$, latent dimension $Q$, kernel type and
         hyperparameters ($\nu,\kappa$ for Mat\'ern; $t$ for heat)
\Statex \textbf{Output:} feature map $\psi(\cdot)\in\mathbb{R}^{2M}$ with frozen
        $\{\lambda_j,\bm{\xi}_j\}_{j=1}^{M}$
\State $\rho \gets (Q-1)/2$
\State define the (unnormalized) spectral density from the hyperparameters,
       \[
         p(\lambda)\;\propto\;\tilde k(\lambda)\,|c(\lambda)|^{-2},\quad
         \tilde k(\lambda)=
         \begin{cases}
           \exp\!\big(-t(\lambda^2+\rho^2)\big) & \text{(heat)}\\[2pt]
           \big(\lambda^2+\rho^2+\tfrac{2\nu}{\kappa^2}\big)^{-(\nu+Q/2)} & \text{(Mat\'ern)}
         \end{cases}
       \]
       \Comment{$|c(\lambda)|^{-2}$ is the Plancherel measure}
\State $\lambda_{\max}\gets 50\sqrt{2\nu}/\kappa$ (Mat\'ern) or $5/\sqrt{t}+10$ (heat);\quad
       $\bar p \gets 1.05\,\max_{\lambda\in(0,\lambda_{\max}]} p(\lambda)$
       \Comment{rejection envelope}
\For{$j=1$ \textbf{to} $M$}
  \State draw $\mathbf{g}\sim\mathcal{N}(\mathbf{0},\mathbf{I}_Q)$;\quad
         $\bm{\xi}_j \gets \mathbf{g}/\|\mathbf{g}\|_2$
         \Comment{ideal direction, uniform on $\mathbb{S}^{Q-1}$}
  \Repeat \Comment{rejection sampling of $\lambda_j\sim p(\lambda)$}
    \State draw $\lambda \sim \mathrm{Unif}(0,\lambda_{\max})$,\;
           $u \sim \mathrm{Unif}(0,\bar p)$
  \Until{$u \le p(\lambda)$}
  \State $\lambda_j \gets \lambda$
\EndFor
\State \textbf{return} the map, with Busemann function
       $B(\mathbf{x},\bm{\xi})=-\log\!\big(-\langle\mathbf{x},\bm{\xi}\rangle_{\mathcal{L}}\big)$,
       \[
         \psi(\mathbf{x})=\tfrac{1}{\sqrt{M}}
         \big[\psi_{\lambda_1,\bm{\xi}_1}(\mathbf{x})^{\top},\dots,
              \psi_{\lambda_M,\bm{\xi}_M}(\mathbf{x})^{\top}\big]^{\top},\;
         \psi_{\lambda_j,\bm{\xi}_j}(\mathbf{x})=e^{\rho B}\!
         \begin{bmatrix}\cos(\lambda_j B)\\ \sin(\lambda_j B)\end{bmatrix}
       \]
       \Comment{$\{\lambda_j,\bm{\xi}_j\}$ sampled once and kept fixed}
\end{algorithmic}
\end{algorithm}

In this section, we describe the proposed RHFF-LVM algorithm in detail.
RHFF-LVM consists of three main steps: construction of the RHFF feature map, initialization of the latent variables, and update of the latent variables by natural gradient descent.
\par
We first give an overview of RHFF-LVM in \Cref{Alg:RHFF-LVM}.
The algorithm takes as input the data $\mathbf{Y}$ to be reduced in dimension, the number of features $M$, the latent dimension $Q$, and the kernel hyperparameters, together with the optimization parameters: the number of epochs $T$, the weight of the prior, and the number of consecutive non-improving iterations $P$ after which the algorithm terminates.
The output is the corresponding latent variables in hyperbolic space.
The algorithm terminates when either (i) the maximum number of epochs is reached, or (ii) no improvement is observed for $P$ consecutive iterations, where improvement at each iteration is measured by the leave-one-out predictive likelihood.
\par
Next, \Cref{Alg:ConstructRHFF} details the construction of the RHFF feature map.
Constructing the RHFF feature map requires sampling its parameters.
The direction $\boldsymbol{\xi}$ only needs to be drawn from the uniform distribution on the sphere.
RHFF obtains $\boldsymbol{\xi}$ by normalizing a sample drawn from a standard normal distribution.
Sampling $\lambda$ is more involved than sampling $\boldsymbol{\xi}$, as it must be drawn from an unnormalized density.
Nevertheless, since $\lambda$ is a one-dimensional random variable, there is no need for the sophisticated sampling methods required in high-dimensional settings.
In this work, we restrict the domain of $\lambda$ and sample it by simple rejection sampling.
\par
\begin{algorithm}[t]
\caption{\textsc{NaturalGradientStep}: one batched line-search update of all latents}
\label{Alg:Step}
\begin{algorithmic}[1]
\Require spatial latents $\mathbf{X}_s\in\mathbb{R}^{N\times Q}$, noise precision $\beta$,
         trial step $\eta$, data $\mathbf{Y}$, feature map $\psi$, prior weight $\gamma$,
         clip $\tau$
\Statex \textbf{Output:} updated $\mathbf{X}_s,\beta,\eta$
\Function{Obj}{$\mathbf{X}_s,\beta$}
  \State $\mathbf{x}_n \gets \mathcal{P}(\mathbf{x}_{n,s})
          =[\sqrt{1+\|\mathbf{x}_{n,s}\|_2^2},\,\mathbf{x}_{n,s}^{\top}]^{\top}$
          for all $n$ \Comment{lift to $\mathcal{L}^{Q}$}
  \State $\bm{\Psi}\gets[\psi(\mathbf{x}_1),\dots,\psi(\mathbf{x}_N)]^{\top}$;\quad
         $\mathbf{C}=\bm{\Psi}\bm{\Psi}^{\top}+\beta^{-1}\mathbf{I}$
  \State \textbf{return} $\ell=\tfrac{D}{2}\log|\mathbf{C}|
         +\tfrac12\mathrm{tr}(\mathbf{C}^{-1}\mathbf{Y}\mathbf{Y}^{\top})
         +\tfrac{\gamma}{D}\sum_{n}\operatorname{arccosh}(x_{n,t})^2$
         \Comment{$|\mathbf{C}|,\mathbf{C}^{-1}$ via det.-lemma / Woodbury: $\mathcal{O}(NM^2{+}M^3)$}
\EndFunction
\State compute Euclidean gradients $\nabla^{E}_{\mathbf{X}_s}\ell$ and $\partial_\beta\ell$
       at $(\mathbf{X}_s,\beta)$ \Comment{batched over all $N$ latents}
\For{$n=1$ \textbf{to} $N$} \Comment{natural gradient under the pullback metric}
  \State $\mathbf{d}_n \gets
         \big(\mathbf{I}+\mathbf{x}_{n,s}\mathbf{x}_{n,s}^{\top}\big)
         \nabla^{E}_{\mathbf{x}_{n,s}}\ell$
\EndFor
\State $\mathbf{D}\gets[\mathbf{d}_1,\dots,\mathbf{d}_N]^{\top}$;\quad
       $\mathbf{D}\gets \mathbf{D}\cdot\min\!\big(1,\tau/\|\mathbf{D}\|\big)$
       \Comment{gradient-norm clipping}
\State $\ell_0\gets\Call{Obj}{\mathbf{X}_s,\beta}$
\Repeat \Comment{backtracking line search on $(\mathbf{X}_s,\beta)$}
  \State $\mathbf{X}_s'\gets\mathbf{X}_s-\eta\,\mathbf{D}$;\quad
         $\beta'\gets\beta\cdot\exp(-\eta\,\partial_{\log\beta}\ell)$
  \If{$\Call{Obj}{\mathbf{X}_s',\beta'}<\ell_0$}
     \State $\mathbf{X}_s,\beta\gets\mathbf{X}_s',\beta'$;\quad
            $\eta\gets\min(1.1\,\eta,\,5\eta_0)$ \Comment{accept, grow}
     \State \textbf{break}
  \Else
     \State $\eta\gets\eta/2$ \Comment{reject, shrink}
  \EndIf
\Until{$\eta<\eta_{\min}$}
\State \textbf{return} $\mathbf{X}_s,\beta,\eta$
\end{algorithmic}
\end{algorithm}
Finally, \Cref{Alg:Step} shows a single update step of the gradient-based optimization.
The objective of the proposed method is the log posterior density, namely the log-likelihood plus a term given by the sum of squared distances of the latent variables from the origin.
This objective is evaluated after applying the Woodbury identity, which reduces the computational cost to the order of the number of features $M$.
From the resulting objective value and its gradient with respect to $\mathbf{x}_{s}$, we compute the update direction in hyperbolic space via the pullback metric, and move along this direction by a step size $\eta$ determined by backtracking line search.
Likewise, the precision parameter $\beta$ is optimized jointly in log space.
Combining these algorithms, RHFF-LVM learns latent variables in hyperbolic space.

\section{Proof of Proposition \ref{Prop:Error-Bound}}
\label{Section:Proof-of-Error-Bound}

Let $\mathcal{X} \subset \mathcal{L}^{Q}$ be a compact subset of the $Q$-dimensional Lorentz model ($Q \ge 2$) that contains the origin and has a radius $R$. 
This assumption restricts the inputs to a bounded region within the manifold. 
We define the kernel and its estimator as follows:
\begin{align*}
    k(\mathbf{x},\mathbf{x}') &= \mathbb{E}_{\lambda \sim p(\lambda),\bm{\xi}\sim p(\bm{\xi})}\left[g_{\lambda,\bm{\xi}}(\mathbf{x},\mathbf{x}')\right], \\
    \hat{k}(\mathbf{x},\mathbf{x}') &= \frac{1}{M}\sum_{j=1}^{M} g_{\lambda_{j},\bm{\xi}_{j}}(\mathbf{x},\mathbf{x}'),\\
    g_{\lambda,\bm{\xi}}(\mathbf{x},\mathbf{x}') &= \exp\left(\rho B(\mathbf{x},\bm{\xi})+\rho B(\mathbf{x}',\bm{\xi})\right)\cos\left(\lambda B(\mathbf{x},\bm{\xi})-\lambda B(\mathbf{x}',\bm{\xi})\right).
\end{align*}
The explicit form of $g_{\lambda,\bm{\xi}}(\mathbf{x},\mathbf{x}')$ follows from the derivation of the randomized features in Section \ref{Section:Proof-of-Discrete-Spherical-Func}. 
Let $\hat{f}(\mathbf{x},\mathbf{x}')=\hat{k}(\mathbf{x},\mathbf{x}')-k(\mathbf{x},\mathbf{x}')$ denote the approximation error. In the remainder of this proof, we establish a uniform error bound on this function over $\mathcal{X}$.

\subsection{Construction of the spatial $r$-net}
\label{Subsec:Metric-Definition}
First, we construct an $r$-net over the input domain of the error function. 
Since the kernels targeted by conventional Random Fourier Features are stationary, the error function depends only on the difference $\hat{f}(\mathbf{x}-\mathbf{x}')$, requiring an $r$-net only to cover $\mathcal{X}$. 
However, since translation invariance does not hold in hyperbolic spaces, we must construct an $r$-net to cover the product space $\mathcal{X}\times\mathcal{X}$ with hyperbolic balls of radius $r$.

\subsubsection{Definition of the metric}
To this end, we define a metric on the product space $\mathcal{X} \times \mathcal{X}$. 
Specifically, we endow the product space with the following $\ell_{1}$-type metric.

\begin{definition}\label{Definition:ell1-Metric}
    Let $\mathbf{x}, \mathbf{x}', \mathbf{z}, \mathbf{z}' \in \mathcal{X}$.
    The $\ell_{1}$-type metric between two points $(\mathbf{x}, \mathbf{x}')$ and $(\mathbf{z}, \mathbf{z}')$ in the product space $\mathcal{X} \times \mathcal{X}$ is defined as:
    \begin{align*}
        d_\Pi((\mathbf{x},\mathbf{x}'),(\mathbf{z},\mathbf{z}')) = d_\mathcal{L}(\mathbf{x},\mathbf{z}) + d_\mathcal{L}(\mathbf{x}',\mathbf{z}')
    \end{align*}
    Furthermore, the open ball of radius $r > 0$ centered at $(\mathbf{x}, \mathbf{x}')$ with respect to $d_\Pi$ is denoted by $\mathcal{B}_\Pi((\mathbf{x}, \mathbf{x}'), r)$ and defined by:
    \begin{align*}
        \mathcal{B}_\Pi((\mathbf{x},\mathbf{x}'),r) = \left\{ (\mathbf{z},\mathbf{z}') \in \mathcal{X} \times \mathcal{X} \mathrel{}\middle|\mathrel{} d_\Pi((\mathbf{x},\mathbf{x}'),(\mathbf{z},\mathbf{z}')) < r \right\}.
    \end{align*}
\end{definition}
Next, we determine the Lipschitz constant of the error function $\hat{f}(\mathbf{x}, \mathbf{x}')$. 
First, let $L_{\hat{f}}$ denote the Lipschitz constant of $\hat{f}$ with respect to a single variable, which satisfies:
\begin{align*}
    |\hat{f}(\mathbf{x},\mathbf{x}')-\hat{f}(\mathbf{z},\mathbf{x}')| \leq L_{\hat{f}} d_{\mathcal{L}}(\mathbf{x},\mathbf{z}).
\end{align*}
By exploiting the symmetry of $\hat{f}$ and the triangle inequality, we bound the difference for both variables as follows:
\begin{align*}
    |\hat{f}(\mathbf{x},\mathbf{x}')-\hat{f}(\mathbf{z},\mathbf{z}')| &= |\hat{f}(\mathbf{x},\mathbf{x}')-\hat{f}(\mathbf{z},\mathbf{x}') + \hat{f}(\mathbf{z},\mathbf{x}')-\hat{f}(\mathbf{z},\mathbf{z}')| \\
    &\leq |\hat{f}(\mathbf{x},\mathbf{x}')-\hat{f}(\mathbf{z},\mathbf{x}')| + |\hat{f}(\mathbf{z},\mathbf{x}')-\hat{f}(\mathbf{z},\mathbf{z}')| \\
    &\leq L_{\hat{f}}d_{\mathcal{L}}(\mathbf{x},\mathbf{z}) + L_{\hat{f}}d_{\mathcal{L}}(\mathbf{x}',\mathbf{z}') \\
    &= L_{\hat{f}}d_\Pi((\mathbf{x},\mathbf{x}'),(\mathbf{z},\mathbf{z}')) .
\end{align*}
This demonstrates that the Lipschitz constant of $\hat{f}$ on the $\ell_{1}$-type product space coincides with $L_{\hat{f}}$. 
Consequently, in the subsequent evaluations, we bound the Lipschitz constant using the upper bound of the Riemannian gradient norm $\left\|\nabla_{\mathbf{x}} \hat{f}(\mathbf{x},\mathbf{x}')\right\|_{\mathcal{L}}$ with respect to a single variable.

\subsubsection{Upper bound of the covering number in the product space $\mathcal{X}\times\mathcal{X}$}
Using the metric defined above, we derive an upper bound on the covering number $N$, which denotes the minimum number of balls $\mathcal{B}_\Pi((\mathbf{x}, \mathbf{x}'), r)$ required to cover the product space $\mathcal{X} \times \mathcal{X}$. 
For the $\ell_{1}$-type metric $d_\Pi$, the product space can be covered by the Cartesian product of $r/2$-nets covering $\mathcal{X}$. 
Following the volume comparison theorem in hyperbolic spaces, the covering number $T$ required to cover the product space $\mathcal{X} \times \mathcal{X}$ can be bounded as stated in the following lemma.
\begin{lemma}
    Let $d_\Pi$ be the $\ell_1$-type metric on the product space $\mathcal{X} \times \mathcal{X}$ as defined in Definition \ref{Definition:ell1-Metric}.
    The covering number $T$ of the product space at scale $r$ is bounded by:
    \begin{align*}
        T \leq Q^{2} \left( R + \frac{r}{4} \right)^2 \coth^{2Q}\left( \frac{r}{4} \right) \exp\left( 2(Q-1)R \right).
    \end{align*}
\end{lemma}
\begin{proof}
    First, the volume of a ball with radius $\rho$ in the $Q$-dimensional Lorentz model $\mathcal{L}^{Q}$ is expressed as:
    \begin{align*}
        V_{\mathcal{L}}(\rho) = \omega_{Q}\int_{0}^{\rho} \sinh^{Q-1}(t) dt,
    \end{align*}
    where $\omega_{Q}=2\pi^{\frac{Q}{2}}/ \Gamma(\frac{Q}{2})$ denotes the surface area of the unit $(Q-1)$-sphere. 
    According to the standard volume argument, an upper bound on the covering number of radius $r$ required to cover a set $\mathcal{X}$ is given by $V_{\mathcal{L}}(R+r/2) / V_{\mathcal{L}}(r/2)$. 
    In the product space $\mathcal{X}\times\mathcal{X}$ equipped with the $\ell_1$-type metric, covering the domain with a ball of radius $r$ is equivalent to covering each individual space with radius $r/2$. 
    Consequently, the upper bound on the covering number $T$ is obtained by squaring the covering bound for a single space:
    \begin{align*}
        T \leq \left\{\frac{V_{\mathcal{L}}(R+\frac{r}{4})}{V_{\mathcal{L}}(\frac{r}{4})}\right\}^{2}.
    \end{align*}
    Note that the volume argument constructs an $r$-net by initially packing balls with half the target radius and subsequently doubling their radius. Therefore, the bound is evaluated using half the required radius for the packing argument.
    \par
    To estimate the covering number, we establish upper and lower bounds for the volume $V_{\mathcal{L}}(\rho)$. 
    First, since $\sinh(t)$ is monotonically increasing, the inequality $\sinh(t) \leq \sinh(\rho)$ holds for all $t \in [0, \rho]$. 
    Applying this property, we obtain:
    \begin{align*}
        V_{\mathcal{L}}(\rho) \leq \omega_{Q}\int_{0}^{\rho} \sinh^{Q-1}(\rho) dt = \omega_{Q}\rho \sinh^{Q-1}(\rho).
    \end{align*}
    Next, to derive the lower bound, we proceed as follows. 
    Recalling that $\frac{d}{dt} \sinh^{Q}(t) = Q \sinh^{Q-1}(t) \cosh(t)$ and that $\cosh(t)$ is monotonically increasing for $t \ge 0$, the integral can be bounded as follows:
    \begin{align*}
        V_{\mathcal{L}}(\rho) = \omega_{Q}\int_{0}^{\rho}\sinh^{Q-1}(t)dt &\geq \omega_{Q}\int_{0}^{\rho}\sinh^{Q-1}(t)\frac{\cosh(t)}{\cosh(\rho)}dt \\
        &= \frac{\omega_{Q}}{\cosh(\rho)}\int_{0}^{\rho} \sinh^{Q-1}(t)\cosh(t) dt \\
        &= \frac{\omega_{Q}}{Q\cosh(\rho)}\sinh^{Q}(\rho).
    \end{align*}
    Consequently, the volume $V_{\mathcal{L}}(\rho)$ is bounded as follows:
    \begin{align}
        \frac{\omega_{Q}}{Q\cosh(\rho)}\sinh^{Q}(\rho) \leq V_{\mathcal{L}}(\rho) \leq \omega_{Q}\rho \sinh^{Q-1}(\rho).
    \end{align}
    Using these results, the upper bound for the volume ratio is computed as:
    \begin{align*}
        \frac{V_{\mathcal{L}}(R+\frac{r}{4})}{V_{\mathcal{L}}(\frac{r}{4})} &\leq \frac{\omega_{Q}(R+\frac{r}{4}) \sinh^{Q-1}(R+\frac{r}{4})}{\frac{\omega_{Q}}{Q\cosh(\frac{r}{4})}\sinh^{Q}(\frac{r}{4})} \\
        &= Q\left(R+\frac{r}{4}\right) \coth\left(\frac{r}{4}\right) \left\{ \frac{\sinh(R+\frac{r}{4})}{\sinh(\frac{r}{4})} \right\}^{Q-1}.
    \end{align*}
    By applying the identity $\coth(t) = \frac{\cosh(t)}{\sinh(t)} \geq 1$ for $t \geq 0$ and the addition formula for $\sinh$, we obtain:
    \begin{align*}
        \frac{\sinh(R+\frac{r}{4})}{\sinh(\frac{r}{4})} &= \frac{\sinh(R)\cosh(\frac{r}{4}) + \cosh(R)\sinh(\frac{r}{4})}{\sinh(\frac{r}{4})} \\
        &= \sinh(R)\coth\left(\frac{r}{4}\right) + \cosh(R) \\
        &\leq \left( \sinh(R) + \cosh(R) \right) \coth\left(\frac{r}{4}\right) \\
        &= \coth\left(\frac{r}{4}\right) \exp(R).
    \end{align*}
    Substituting this inequality back into the volume ratio yields:
    \begin{align*}
        \frac{V_{\mathcal{L}}(R+\frac{r}{4})}{V_{\mathcal{L}}(\frac{r}{4})} &\leq Q\left(R+\frac{r}{4}\right) \coth^{Q}\left(\frac{r}{4}\right) \exp\left( (Q-1)R \right).
    \end{align*}
    Finally, squaring this result provides the following upper bound on the covering number $T$:
    \begin{align*}
        T \leq \left( \frac{V_{\mathcal{L}}(R+\frac{r}{4})}{V_{\mathcal{L}}(\frac{r}{4})} \right)^{2} \leq Q^{2}\left(R+\frac{r}{4}\right)^2 \coth^{2Q}\left(\frac{r}{4}\right) \exp\left( 2(Q-1)R \right).
    \end{align*}
\end{proof}

\subsubsection{Decomposition of the Error Function via an $r$-net}
In what follows, let $(\mathbf{u}_{t},\mathbf{v}_{t})$ for $t=1,\dots,T$ denote the centers of the balls in the $r$-net constructed above. We proceed to bound the supremum of the error function using these centers. Specifically, an application of the triangle inequality yields the following proposition:
\begin{prop}
    For any $\epsilon>0$, if $L_{\hat{f}}\leq \frac{\epsilon}{2r}$ and $|\hat{f}(\mathbf{u}_{t},\mathbf{v}_{t})|\leq \frac{\epsilon}{2}$ for all $t=1,\dots,T$, then $\sup_{(\mathbf{x},\mathbf{x}')\in\mathcal{X}\times\mathcal{X}}|\hat{f}(\mathbf{x},\mathbf{x}')|\leq \epsilon$.
    \label{Prop:Error-Decomp}
\end{prop}
\begin{proof}
    For any arbitrary point $(\mathbf{x},\mathbf{x}') \in \mathcal{X} \times \mathcal{X}$, there exists a center $(\mathbf{u}_{t},\mathbf{v}_{t})$ in the $r$-net such that $(\mathbf{x},\mathbf{x}') \in \mathcal{B}_{\Pi}((\mathbf{u}_{t},\mathbf{v}_{t}),r)$. Applying the triangle inequality, we obtain:
    \begin{align*}
        |\hat{f}(\mathbf{x},\mathbf{x}')| &= |\hat{f}(\mathbf{x},\mathbf{x}')-\hat{f}(\mathbf{u}_{t},\mathbf{v}_{t})+\hat{f}(\mathbf{u}_{t},\mathbf{v}_{t})| \\
        &\leq |\hat{f}(\mathbf{x},\mathbf{x}')-\hat{f}(\mathbf{u}_{t},\mathbf{v}_{t})| + |\hat{f}(\mathbf{u}_{t},\mathbf{v}_{t})|.
    \end{align*}
    By the definition of the Lipschitz constant $L_{\hat{f}}$ with respect to the $\ell_{1}$-type metric $d_{\Pi}$, we have:
    \begin{align*}
        |\hat{f}(\mathbf{x},\mathbf{x}')-\hat{f}(\mathbf{u}_{t},\mathbf{v}_{t})| &\leq L_{\hat{f}} d_{\Pi}((\mathbf{x},\mathbf{x}'),(\mathbf{u}_{t},\mathbf{v}_{t})) \\
        &< L_{\hat{f}}r.
    \end{align*}
    Combining this with the given assumptions yields:
    \begin{align*}
        |\hat{f}(\mathbf{x},\mathbf{x}')| &\leq L_{\hat{f}}r + \frac{\epsilon}{2} \leq \epsilon.
    \end{align*}
\end{proof}
Based on Proposition \ref{Prop:Error-Decomp}, bounding the uniform approximation error over $\mathcal{X} \times \mathcal{X}$ reduces to probabilistically bounding two distinct terms: (i) the Lipschitz constant $L_{\hat{f}}$, which governs the smoothness of the error function with respect to the $\ell_1$-type metric, and (ii) the error magnitude $|\hat{f}(\mathbf{u}_t, \mathbf{v}_t)|$ at the centers of the $r$-net. 
In the subsequent sections, we evaluate these bounds independently and combine them via a union bound to establish the final uniform error guarantee.

\subsection{Bounding the Lipschitz Constant}
\subsubsection{Boundedness of the Busemann Function}
First, we establish a bound on the Busemann function over the compact set $\mathcal{X}$, a result that is instrumental for the subsequent proofs. The following lemma provides this upper bound.

\begin{lemma}
    Let $\mathcal{X} \subset \mathcal{L}^{Q}$ be a compact subset with radius $R$ containing the origin. For any $\mathbf{x} \in \mathcal{X}$ and an ideal point $\bm{\xi} \in \partial\mathcal{L}^{Q}$, the magnitude of the Busemann function $B(\mathbf{x}, \bm{\xi})$ is bounded by:
    \begin{align*}
        |B(\mathbf{x}, \bm{\xi})| \le R.
    \end{align*}
    \label{Lemma:Bound-of-Bx}
\end{lemma}

\begin{proof}
    Since $\mathcal{X}$ contains the origin $\mathbf{o}$ and has a radius $R$, we have $d_{\mathcal{L}}(\mathbf{o}, \mathbf{x}) \le R$ for any $\mathbf{x} \in \mathcal{X}$. Recall that the Busemann function is defined using a unit-speed geodesic $\bm{\gamma}(a)$ directed toward the ideal point $\bm{\xi}$ as $a \to \infty$:
    \begin{align*}
        B(\mathbf{x}, \bm{\xi}) &= \lim_{a \to \infty} \left( a - d_{\mathcal{L}}(\mathbf{x}, \bm{\gamma}(a)) \right) = -\log \left(-\langle\mathbf{x},\bm{\xi}\rangle_{\mathcal{L}}\right).
    \end{align*}
    In the Lorentz model, the origin is represented as $\mathbf{o}=[1,0,\ldots ,0]^{\top} \in \mathbb{R}^{Q+1}$, and an ideal point $\bm{\xi}$ on the boundary is represented as $\bm{\xi}=[1,(\bm{\xi}')^{\top}]^{\top}$ for some $\bm{\xi}' \in \mathbb{S}^{Q-1}$. Consequently, $\langle\mathbf{o}, \bm{\xi}\rangle_{\mathcal{L}} = -1$, which implies $B(\mathbf{o}, \bm{\xi})=0$. Applying the triangle inequality to the geodesic distance yields:
    \begin{align*}
        |B(\mathbf{x}, \bm{\xi})| &= |B(\mathbf{x}, \bm{\xi})-B(\mathbf{o}, \bm{\xi})| \\
        &= \left| \lim_{a \to \infty} \left( d_{\mathcal{L}}(\mathbf{x}, \bm{\gamma}(a))-d_{\mathcal{L}}(\mathbf{o}, \bm{\gamma}(a)) \right) \right| \\
        &\leq d_{\mathcal{L}}(\mathbf{x}, \mathbf{o}) \leq R.
    \end{align*}
\end{proof}
\par
Next, we bound the gradient of the Busemann function $\nabla_{\mathbf{x}}B(\mathbf{x},\bm{\xi})$.

\begin{lemma}
    Let $\mathcal{X} \subset \mathcal{L}^{Q}$ be a compact subset. For any $\mathbf{x} \in \mathcal{X}$ and ideal point $\bm{\xi} \in \partial\mathcal{L}^{Q}$, the Busemann function $B(\mathbf{x},\bm{\xi})$ is 1-Lipschitz with respect to $\mathbf{x}$, and its gradient with respect to a single variable satisfies:
    \begin{align*}
        \|\nabla_{\mathbf{x}}B(\mathbf{x},\bm{\xi})\|_{\mathcal{L}} \le 1.
    \end{align*}
    \label{Lemma:Bound-of-Nabla-Bx}
\end{lemma}

\begin{proof}
    By the definition of the Busemann function and the triangle inequality, for any $\mathbf{x}, \mathbf{x}' \in \mathcal{X}$, we have:
    \begin{align*}
        |B(\mathbf{x}, \bm{\xi}) - B(\mathbf{x}', \bm{\xi})| &= \left| \lim_{a \to \infty} \left( d_{\mathcal{L}}(\mathbf{x}, \bm{\gamma}(a)) - d_{\mathcal{L}}(\mathbf{x}', \bm{\gamma}(a)) \right) \right| \leq d_{\mathcal{L}}(\mathbf{x},\mathbf{x}').
    \end{align*}
    This establishes that $B(\cdot,\bm{\xi})$ is 1-Lipschitz on $\mathcal{X}$. Consequently, the Riemannian norm of its gradient is bounded by its Lipschitz constant, yielding $\|\nabla_{\mathbf{x}}B(\mathbf{x},\bm{\xi})\|_{\mathcal{L}} \le 1$.
\end{proof}

\subsubsection{Justification for the Interchange of Differentiation and Integration}
Next, to compute the gradient of the error function, we verify that the conditions for interchanging integration and differentiation are satisfied for the kernel function $k(\mathbf{x},\mathbf{x}')$. 
We first recall the Leibniz integral rule:
\begin{thrm}[Leibniz Integral Rule]
    Let $X$ be an open subset of $\mathbb{R}$ and $\Omega$ be a measure space. Suppose a function $f: X \times \Omega \rightarrow \mathbb{R}$ satisfies the following conditions:
    \begin{itemize}
        \item $f(x,\omega)$ is a Lebesgue integrable function of $\omega$ for each $x \in X$.
        \item For almost all $\omega \in \Omega$, the partial derivative $\frac{\partial f}{\partial x}$ exists for all $x \in X$.
        \item There exists an integrable function $\Theta: \Omega \rightarrow \mathbb{R}$ such that $\left| \frac{\partial f}{\partial x}(x,\omega) \right| \leq \Theta(\omega)$ for all $x \in X$.
    \end{itemize}
    Then, for all $x \in X$, the following equality holds:
    \begin{align*}
        \frac{d}{dx} \left( \int_{\Omega}f(x,\omega)d\omega \right) = \int_{\Omega}\frac{\partial}{\partial x}f(x,\omega)d\omega.
    \end{align*}
\end{thrm}
Applying this theorem to each component of the gradient vector yields the following proposition:

\begin{prop}
    Assume that the random variable $\lambda \geq 0$ has a finite second moment, i.e., $\mathbb{E}[\lambda^2]<\infty$. For any $(\mathbf{x},\mathbf{x}') \in \mathcal{X} \times \mathcal{X}$, the gradient of the kernel satisfies:
    \begin{align*}
        \nabla_{\mathbf{x}}k(\mathbf{x},\mathbf{x}') = \mathbb{E}_{\lambda,\bm{\xi}}[\nabla_{\mathbf{x}}g_{\lambda,\bm{\xi}}(\mathbf{x},\mathbf{x}')].
    \end{align*}
    \label{Prop:Interchangeability}
\end{prop}

\begin{proof}
    We first establish the existence of an integrable majorant for each component of the gradient of the integrand $g_{\lambda,\bm{\xi}}(\mathbf{x},\mathbf{x}')$, satisfying the third condition of the Leibniz integral rule. The function $g_{\lambda,\bm{\xi}}(\mathbf{x},\mathbf{x}')$ is given by:
    \begin{align*}
        g_{\lambda,\bm{\xi}}(\mathbf{x},\mathbf{x}') &= \exp\left(\rho B(\mathbf{x},\bm{\xi})+\rho B(\mathbf{x}',\bm{\xi})\right)\cos\left(\lambda B(\mathbf{x},\bm{\xi})-\lambda B(\mathbf{x}',\bm{\xi})\right).
    \end{align*}
    Differentiating this expression with respect to $\mathbf{x}$ yields:
    \begin{align*}
        \nabla_{\mathbf{x}}g_{\lambda,\bm{\xi}}(\mathbf{x},\mathbf{x}') &= \exp\left(\rho B(\mathbf{x},\bm{\xi})+\rho B(\mathbf{x}',\bm{\xi})\right) \\
        &\quad \times \left[ \rho \cos\left(\lambda B(\mathbf{x},\bm{\xi}) - \lambda B(\mathbf{x}',\bm{\xi})\right) - \lambda \sin\left(\lambda B(\mathbf{x},\bm{\xi}) - \lambda B(\mathbf{x}',\bm{\xi})\right) \right] \nabla_{\mathbf{x}} B(\mathbf{x},\bm{\xi}).
    \end{align*}
    For the exponential term, Lemma \ref{Lemma:Bound-of-Bx} ensures that $\exp\left(\rho B(\mathbf{x},\bm{\xi})+\rho B(\mathbf{x}',\bm{\xi})\right) \leq \exp(2\rho R) = \exp((Q-1)R)$, since $\rho=(Q-1)/2$. 
    Furthermore, the magnitude of the oscillatory factor is bounded by $\sqrt{\rho^2 + \lambda^2}$ via the trigonometric identity for linear combinations of sine and cosine:
    \begin{align*}
        \left| \rho \cos\left(\lambda B(\mathbf{x},\bm{\xi}) - \lambda B(\mathbf{x}',\bm{\xi})\right) - \lambda \sin\left(\lambda B(\mathbf{x},\bm{\xi}) - \lambda B(\mathbf{x}',\bm{\xi})\right) \right| \leq \sqrt{\rho^2 + \lambda^2}.
    \end{align*}
    By Lemma \ref{Lemma:Bound-of-Nabla-Bx}, we also have $\|\nabla_{\mathbf{x}} B(\mathbf{x},\bm{\xi})\|_{\mathcal{L}}\leq 1$. 
    Combining these bounds, the Riemannian norm of the gradient $\nabla_{\mathbf{x}}g_{\lambda,\bm{\xi}}(\mathbf{x},\mathbf{x}')$ satisfies:
    \begin{align}
        \|\nabla_{\mathbf{x}}g_{\lambda,\bm{\xi}}(\mathbf{x},\mathbf{x}')\|_{\mathcal{L}} \leq \sqrt{\rho^2 + \lambda^2}\exp\left((Q-1)R\right). \label{Eq:Bound-of-Nabla-g}
    \end{align}
    Since the absolute value of each gradient component is bounded by the vector's norm, defining $\Theta(\lambda) = \sqrt{\rho^2 + \lambda^2}\exp\left((Q-1)R\right)$ yields a valid majorant independent of $\mathbf{x} \in \mathcal{X}$. 
    Using the inequality $\sqrt{\rho^2 + \lambda^2} \leq \rho + \lambda$ and the finite first moment of $\lambda$ (which follows from the assumption $\mathbb{E}[\lambda^2]<\infty$), we deduce that $\mathbb{E}[\Theta(\lambda)] < \infty$. 
    This establishes the existence of an integrable majorant, satisfying the third condition of the Leibniz integral rule.
    \par
    Next, we verify the integrability and differentiability required to apply the Leibniz integral rule. As established earlier, $g_{\lambda,\bm{\xi}}(\mathbf{x},\mathbf{x}')$ is a bounded function on the compact set $\mathcal{X}$ for any fixed $\mathbf{x}'$. Since the integration is taken over a probability measure, the function is trivially Lebesgue integrable. Furthermore, because hyperbolic spaces are smooth Riemannian manifolds and the Busemann function $B(\mathbf{x}, \bm{\xi})$ is differentiable with respect to $\mathbf{x}$, the partial derivatives of $g_{\lambda,\bm{\xi}}(\mathbf{x},\mathbf{x}')$ with respect to $\mathbf{x}$ exist for all $\mathbf{x} \in \mathcal{X}$.
    \par
    Consequently, all conditions of the Leibniz integral rule are satisfied for each component of the gradient. Thus, for any $\mathbf{x} \in \mathcal{X}$, we have:
    \begin{align*}
        \nabla_{\mathbf{x}} k(\mathbf{x}, \mathbf{x}') &= \nabla_{\mathbf{x}} \mathbb{E}_{\lambda,\bm{\xi}} \left[ g_{\lambda,\bm{\xi}}(\mathbf{x},\mathbf{x}') \right] = \mathbb{E}_{\lambda,\bm{\xi}} \left[ \nabla_{\mathbf{x}} g_{\lambda,\bm{\xi}}(\mathbf{x},\mathbf{x}') \right].
    \end{align*}
\end{proof}
Proposition \ref{Prop:Interchangeability} establishes the exchangeability between the gradient operator and the expectation operator for the kernel function $k(\mathbf{x}, \mathbf{x}')$, provided that $\lambda$ has a finite second moment.

\subsubsection{Deterministic Upper Bound for the Gradient Norm}
Using the results established above, we now derive a deterministic upper bound for the squared Lipschitz constant. 
As demonstrated in Section \ref{Subsec:Metric-Definition}, the Lipschitz constant of the bivariate function $\hat{f}(\mathbf{x},\mathbf{x}')$ coincides with that of a single variable. 
Consequently, we evaluate $L_{\hat{f}}$ by bounding the squared Riemannian gradient norm $\left\|\nabla_{\mathbf{x}} \hat{f}(\mathbf{x},\mathbf{x}')\right\|_{\mathcal{L}}^{2}$.
\par
First, leveraging Proposition \ref{Prop:Interchangeability} and the triangle inequality along with the identity $(a+b)^{2} \leq 2a^{2}+2b^{2}$, we bound the squared Riemannian norm of the gradient of the error function as follows:
\begin{align*}
    \left\|\nabla_{\mathbf{x}} \hat{f}(\mathbf{x},\mathbf{x}')\right\|_{\mathcal{L}}^{2}
    &= \left\|\nabla_{\mathbf{x}} \hat{k}(\mathbf{x},\mathbf{x}')-\nabla_{\mathbf{x}} k(\mathbf{x},\mathbf{x}')\right\|_{\mathcal{L}}^{2} \\
    &\leq \left( \left\|\nabla_{\mathbf{x}} \hat{k}(\mathbf{x},\mathbf{x}')\right\|_{\mathcal{L}} + \left\|\nabla_{\mathbf{x}} k(\mathbf{x},\mathbf{x}')\right\|_{\mathcal{L}} \right)^2 \\
    &\leq 2\left\|\nabla_{\mathbf{x}} \hat{k}(\mathbf{x},\mathbf{x}')\right\|_{\mathcal{L}}^{2} + 2\left\|\nabla_{\mathbf{x}} k(\mathbf{x},\mathbf{x}')\right\|_{\mathcal{L}}^{2}.
\end{align*}
To evaluate these two terms, we utilize the upper bound $\|\nabla_{\mathbf{x}}g_{\lambda,\bm{\xi}}(\mathbf{x},\mathbf{x}')\|_{\mathcal{L}}^2 \leq (\rho^2 + \lambda^2)\exp\left(2(Q-1)R\right)$ derived in Eq. \eqref{Eq:Bound-of-Nabla-g}.
\par
For the first term (the estimator), applying Jensen's inequality for finite sums yields:
\begin{align*}
    \left\|\nabla_{\mathbf{x}} \hat{k}(\mathbf{x},\mathbf{x}')\right\|_{\mathcal{L}}^{2} 
    &= \left\|\frac{1}{M}\sum_{j=1}^{M}\nabla_{\mathbf{x}} g_{\lambda_{j},\bm{\xi}_{j}}(\mathbf{x},\mathbf{x}')\right\|_{\mathcal{L}}^{2} \\
    &\leq \frac{1}{M}\sum_{j=1}^{M} \left\|\nabla_{\mathbf{x}} g_{\lambda_{j},\bm{\xi}_{j}}(\mathbf{x},\mathbf{x}')\right\|_{\mathcal{L}}^{2} \\
    &\leq \frac{1}{M}\sum_{j=1}^{M}(\rho^{2}+\lambda_{j}^{2})\exp\left(2(Q-1)R\right) \\
    &= \left(\rho^{2}+\frac{1}{M}\sum_{j=1}^{M}\lambda_{j}^{2}\right)\exp\left(2(Q-1)R\right). 
\end{align*}
Similarly, for the second term (the true kernel), we apply Jensen's inequality for expectations to obtain a similar bound:
\begin{align*}
    \left\|\nabla_{\mathbf{x}} k(\mathbf{x},\mathbf{x}')\right\|_{\mathcal{L}}^{2} 
    &= \left\|\nabla_{\mathbf{x}} \mathbb{E}_{\lambda,\bm{\xi}} \left[ g_{\lambda,\bm{\xi}}(\mathbf{x},\mathbf{x}') \right] \right\|_{\mathcal{L}}^{2} \\
    &= \left\|\mathbb{E}_{\lambda,\bm{\xi}} \left[ \nabla_{\mathbf{x}}g_{\lambda,\bm{\xi}}(\mathbf{x},\mathbf{x}') \right] \right\|_{\mathcal{L}}^{2} \\
    &\leq \mathbb{E}_{\lambda,\bm{\xi}} \left[ \left\|\nabla_{\mathbf{x}}g_{\lambda,\bm{\xi}}(\mathbf{x},\mathbf{x}') \right\|_{\mathcal{L}}^{2} \right] \\
    &\leq \left(\rho^{2}+\mathbb{E}[\lambda^{2}]\right)\exp\left(2(Q-1)R\right).
\end{align*}
Combining these bounds, we arrive at the following upper bound on the squared gradient norm of the error function:
\begin{align}
    \left\|\nabla_{\mathbf{x}} \hat{f}(\mathbf{x},\mathbf{x}')\right\|_{\mathcal{L}}^{2} \leq 2\left(2\rho^{2} + \frac{1}{M}\sum_{j=1}^{M}\lambda_{j}^{2} + \mathbb{E}[\lambda^{2}]\right)\exp\left(2(Q-1)R\right). \label{Eq:Bound-of-Nabla-hatf}
\end{align}

\subsubsection{Probabilistic Bound via Markov's Inequality}
Finally, we apply Markov's inequality to derive a probabilistic bound on the Lipschitz constant. 
To this end, we take the expectation of the previous inequality with respect to the random features. 
Since the samples $\lambda_j$ are i.i.d., we have $\mathbb{E} \left[ \frac{1}{M}\sum_{j=1}^{M}\lambda_{j}^{2} \right] = \mathbb{E}[\lambda^2]$, which yields the following bound on the expected squared Lipschitz constant:
\begin{align*}
    \mathbb{E} \left[ L_{\hat{f}}^2 \right] 
    &\leq 2\left(2\rho^{2} + \mathbb{E}[\lambda^{2}] + \mathbb{E}[\lambda^{2}]\right)\exp\left(2(Q-1)R\right) \\
    &= 4(\rho^{2} + \mathbb{E}[\lambda^{2}])\exp\left(2(Q-1)R\right).
\end{align*}
Denoting the second moment of $\lambda$ by $\sigma_{\lambda}^{2} = \mathbb{E}[\lambda^{2}]$, we obtain the following proposition regarding the probabilistic upper bound on $L_{\hat{f}}$.

\begin{prop}
    For any $\epsilon>0$, assuming the second moment $\sigma^2_{\lambda}=\mathbb{E}[\lambda^{2}]$ is finite, the Lipschitz constant $L_{\hat{f}}$ of the approximation error satisfies the following probabilistic bound:
    \begin{align*}
        \Pr \left( L_{\hat{f}} \ge \frac{\epsilon}{2r} \right) = \Pr \left( L_{\hat{f}}^2 \ge \frac{\epsilon^2}{4r^2} \right) \le \frac{16r^2(\rho^{2}+\sigma_{\lambda}^{2})}{\epsilon^2}\exp\left(2(Q-1)R\right).
    \end{align*}
    \label{Prop:Bound-of-Lipschitz-Const}
\end{prop}

\subsection{Bounding the Error at the $r$-net Centers}
Next, we bound the approximation error at the centers $(\mathbf{u}_{t},\mathbf{v}_{t})$ of the $r$-net. To this end, we first establish a deterministic upper bound on the estimator $\hat{k}(\mathbf{x},\mathbf{x}')$.

\begin{lemma}
    For any $(\mathbf{x},\mathbf{x}') \in \mathcal{X} \times \mathcal{X}$, the magnitude of the kernel estimator $\hat{k}(\mathbf{x},\mathbf{x}')$ is bounded by:
    \begin{align*}
        |\hat{k}(\mathbf{x},\mathbf{x}')| \le \exp\left((Q-1)R\right).
    \end{align*}
    \label{Lemma:Bound-of-hatk}
\end{lemma}

\begin{proof}
    By Lemma \ref{Lemma:Bound-of-Bx}, we have $|B(\mathbf{x},\bm{\xi})| \leq R$ for any $\mathbf{x} \in \mathcal{X}$ and ideal point $\bm{\xi}$. Applying the triangle inequality to the estimator yields:
    \begin{align*}
        |\hat{k}(\mathbf{x},\mathbf{x}')| &= \left| \frac{1}{M}\sum_{j=1}^{M}g_{\lambda_{j},\bm{\xi}_{j}}(\mathbf{x},\mathbf{x}') \right| \\
        &\leq \frac{1}{M}\sum_{j=1}^{M} \left| \exp\left(\rho B(\mathbf{x},\bm{\xi}_{j}) + \rho B(\mathbf{x}',\bm{\xi}_{j})\right)\cos\left(\lambda_{j} B(\mathbf{x},\bm{\xi}_{j}) - \lambda_{j}B(\mathbf{x}',\bm{\xi}_{j})\right) \right| \\
        &\leq \frac{1}{M} \sum_{j=1}^{M} \exp\left(2\rho R\right) \cdot 1 \\
        &= \exp\left((Q-1)R\right).
    \end{align*}
\end{proof}

\subsubsection{Probabilistic Bound via the Union Bound}
Since $\hat{k}(\mathbf{x},\mathbf{x}')$ is an unbiased estimator of the true kernel, the approximation error $\hat{f}(\mathbf{x},\mathbf{x}') = \hat{k}(\mathbf{x},\mathbf{x}') - k(\mathbf{x},\mathbf{x}')$ satisfies $\mathbb{E}[\hat{f}(\mathbf{x},\mathbf{x}')]=0$. Given the boundedness of the estimator established in Lemma \ref{Lemma:Bound-of-hatk}, we apply Hoeffding's inequality to $\hat{f}(\mathbf{x},\mathbf{x}')$, yielding the following proposition:

\begin{prop}
    For any $\epsilon>0$ and any fixed point $(\mathbf{x},\mathbf{x}') \in \mathcal{X} \times \mathcal{X}$, the approximation error is probabilistically bounded by:
    \begin{align*}
        \Pr \left( |\hat{f}(\mathbf{x},\mathbf{x}')| \ge \frac{\epsilon}{2} \right) \leq 2\exp\left(-\frac{M\epsilon^{2}}{8\exp\left(2(Q-1)R\right)}\right).
    \end{align*}
    Furthermore, applying the union bound over all $T$ centers of the $r$-net, the probability that the error exceeds $\epsilon/2$ at one or more centers $(\mathbf{u}_{t},\mathbf{v}_{t})$ is bounded by:
    \begin{align*}
        \Pr \left(\bigcup_{t=1}^{T} \left\{ |\hat{f}(\mathbf{u}_{t},\mathbf{v}_{t})| \ge \frac{\epsilon}{2} \right\} \right) \leq 2T\exp\left( -\frac{M\epsilon^{2}}{8\exp\left(2(Q-1)R\right)} \right).
    \end{align*}
    \label{Prop:Bound-at-Centers}
\end{prop}

\subsection{Synthesis of the Uniform Error Bound}
By synthesizing the probabilistic bounds derived above, we now establish the final uniform bound for the approximation error. Recall from Proposition \ref{Prop:Error-Decomp} that the uniform error exceeding $\epsilon$ implies the union of two events: (i) the error at one or more centers of the $r$-net exceeds $\epsilon/2$, or (ii) the Lipschitz constant $L_{\hat{f}}$ exceeds $\epsilon/(2r)$. Consequently, applying the union bound and substituting the upper bound for the covering number $T$, we obtain the following intermediate inequality for any covering radius $r > 0$:
\begin{equation}
    \Pr\left( \sup_{(\mathbf{x},\mathbf{x}') \in \mathcal{X} \times \mathcal{X}} |\hat{f}(\mathbf{x},\mathbf{x}')| \geq \epsilon \right) \leq \left[ \kappa_{1}r^{2} + 2Q^{2}\left(R+\frac{r}{4}\right)^2\coth^{2Q}\left(\frac{r}{4}\right) \exp(-M\kappa_{2})\right] \exp\left(2(Q-1)R\right), \label{Eq:Intermediate-Bound}
\end{equation}
where $\kappa_{1} = \frac{16(\rho^{2}+\sigma_{\lambda}^{2})}{\epsilon^{2}}$ and $\kappa_{2} = \frac{\epsilon^{2}}{8 \exp\left(2(Q-1)R\right)}$.
\par
Next, we set the covering radius $r$ as a function of the number of features $M$, denoted by $r_M$:
\begin{equation*}
    r_{M} = \exp\left( - \frac{M\kappa_{2}}{2(Q+1)} \right).
\end{equation*}
By definition, $r_M > 0$, ensuring a valid covering radius. Furthermore, since $\kappa_{2} > 0$ and $M \ge 1$, we have $r_M \leq 1$, which trivially implies $R + \frac{r_M}{4} \leq R + \frac{1}{4}$. Additionally, since the function $x \mapsto x \coth(x)$ is monotonically increasing for $x > 0$, the condition $r_M \leq 1$ yields $\frac{r_M}{4} \coth\left(\frac{r_M}{4}\right) \leq \frac{1}{4} \coth\left(\frac{1}{4}\right)$. This allows us to upper-bound the hyperbolic cotangent term as:
\begin{equation*}
    \coth\left(\frac{r_M}{4}\right) \leq \frac{\coth(1/4)}{r_M}.
\end{equation*}
\par
Substituting these inequalities into the right-hand side of Eq. \eqref{Eq:Intermediate-Bound}, the upper bound becomes:
\begin{align*}
    &\left[ \kappa_{1}r_{M}^{2} + 2Q^{2}\left(R+\frac{r_{M}}{4}\right)^2 \coth^{2Q}\left(\frac{r_{M}}{4}\right) \exp(-M\kappa_{2})\right] \exp\left(2(Q-1)R\right) \\
    &\leq \left[ \kappa_{1}r_{M}^{2} + 2Q^{2}\left(R+\frac{1}{4}\right)^2 \coth^{2Q}\left(\frac{1}{4}\right) r_{M}^{-2Q} \exp(-M\kappa_{2}) \right] \exp\left(2(Q-1)R\right).
\end{align*}
For the first term inside the brackets, we have $\kappa_{1}r_M^2 = \kappa_{1}\exp\left( - \frac{M\kappa_{2}}{Q+1} \right)$. The second term can be rearranged as follows:
\begin{align*}
    & 2Q^{2}\left(R+\frac{1}{4}\right)^2\coth^{2Q}\left(\frac{1}{4}\right) r_{M}^{-2Q} \exp(-M\kappa_{2}) \\
    &= 2Q^{2}\left(R+\frac{1}{4}\right)^2\coth^{2Q}\left(\frac{1}{4}\right)\exp\left( \frac{MQ\kappa_{2}}{Q+1} -M\kappa_{2} \right) \\
    &= 2Q^2 \left(R+\frac{1}{4}\right)^2\coth^{2Q}\left(\frac{1}{4}\right) \exp\left( - \frac{M\kappa_{2}}{Q+1}\right).
\end{align*}
\par
Thus, a common decay factor $\exp\left( - \frac{M\kappa_{2}}{Q+1} \right)$ can be factored out from both terms, yielding:
\begin{equation*}
    \left[ \kappa_{1} + 2Q^{2}\left(R+\frac{1}{4}\right)^2 \coth^{2Q}\left(\frac{1}{4}\right) \right] \exp\left( 2(Q-1)R - \frac{M\kappa_{2}}{Q+1} \right).
\end{equation*}
Substituting the definitions of $\kappa_1$ and $\kappa_2$, and defining the constant term within the brackets as $C$, we arrive at the desired inequality, which completes the proof.

\section{Dataset Details}
\label{App:Datasets}

Table~\ref{tab:app_datasets} summarizes the statistics of all datasets. 
Below we describe their provenance and preprocessing, which follows two pipelines depending on the distribution form of the source data.
 
\begin{table}[h]
\centering
\caption{Dataset statistics. $D$ is the input dimensionality after preprocessing; class sizes are the minimum / median / maximum number of points per annotated type.}
\label{tab:app_datasets}
\small
\begin{tabular}{lrrrrl}
\toprule
Dataset & $N$ & Types & $D$ & Class size (min/med/max) & Category \\
\midrule
toytree & 1{,}008 & 6 & 63 & 16 / 96 / 512 & synthetic control \\
Paul~\cite{paul2015transcriptional} & 2{,}730 & 20 & 1{,}000 & 1 / 138 / 373 & scRNA-seq \\
Plass~\cite{plass2018cell} & 21{,}612 & 51 & 50 & 13 / 202 / 6{,}343 & scRNA-seq \\
Tasic~\cite{tasic2018shared} & 23{,}822 & 133 & 3{,}000 & 4 / 115 / 1{,}404 & scRNA-seq \\
\emph{C.~elegans}~\cite{packer2019lineage} & 54{,}649 & 36 & 50 & 25 / 349 / 17{,}520 & scRNA-seq \\
MNIST~\cite{lecun1998gradient} & 60{,}000 & 10 & 784 & 5{,}421 / 5{,}936 / 6{,}742 & image control \\
Zebrafish~\cite{wagner2018single} & 71{,}203 & 63 & 50 & 8 / 416 / 7{,}001 & scRNA-seq \\
Zeisel~\cite{zeisel2018molecular} & 160{,}796 & 265 & 50 & 7 / 167 / 10{,}011 & scRNA-seq \\
MouseDev~\cite{la2021molecular} & 292{,}495 & 24 & 50 & 107 / 3{,}935 / 111{,}063 & scRNA-seq \\
\bottomrule
\end{tabular}
\end{table}
 
\paragraph{Atlases processed by our pipeline.}
\emph{C.~elegans}~\cite{packer2019lineage}, Zebrafish~\cite{wagner2018single}, Zeisel~\cite{zeisel2018molecular}, and MouseDev~\cite{la2021molecular} are ingested from their public releases and processed by a common pipeline: removal of unlabeled cells; count-per-10{,}000 (CP10k) normalization followed by $\log(1+x)$, keeping sparsity; selection of the top $2{,}000$ highly variable genes; z-scoring with clipping at $\pm 10$; and projection to $50$ principal components (fixed random state). 
No subsampling was applied: the pipeline subsamples only above $5 \times 10^{5}$ cells, a threshold none of the datasets reaches. 
MouseDev denotes the developing \emph{mouse} nervous system atlas of \cite{la2021molecular}, labeled at the Class level (24 types).
 
\paragraph{Atlases used in externally preprocessed form.}
Paul~\cite{paul2015transcriptional} and Plass~\cite{plass2018cell} are used exactly as distributed in the Poincar\'e Maps repository~\cite{klimovskaia2020poincare} (CC BY-NC 4.0), which in turn derives them from accessions GSE72857 and GSE103633.
Using the preprocessing distributed by \cite{klimovskaia2020poincare} also ensures that Poincar\'e Map is evaluated on its own preprocessing. 
Tasic~\cite{tasic2018shared} is preprocessed following \cite{kobak2019art} with their published pipeline: exon counts of the VISp and ALM areas (Allen Institute), quality-controlled cells and the 133 cluster labels of the original publication, selection of $3{,}000$ variable genes, counts-per-million normalization, $\log_2(1+x)$, and mean-centering. 
Unlike \cite{kobak2019art}, we do not apply their final 50-PC reduction and use the $3{,}000$-dimensional matrix directly.
 
\paragraph{Controls.}
The toytree dataset consists of noisy copies of the ancestor-path indicator vectors of a complete binary tree with heap numbering (depth $6$ ($63$ nodes) $16$ samples per node, $N=1{,}008$; labels are the 6 depth levels).
MNIST~\cite{lecun1998gradient} uses the raw $28 \times 28$ pixel intensities ($D=784$) without any preprocessing and serves as a non-hierarchical negative control.

\section{Detailed Experimental Setup}
\label{App:Setup}

\paragraph{Frequency sampling and truncation.}
For the heat and Mat\'ern spectra, we sample $\lambda$ by rejection sampling with a uniform proposal on $[0,\lambda_{\max}]$, which truncates $p(\lambda)$ at $\lambda_{\max}$ ($5/\sqrt{t}+10$ for heat; $50\sqrt{2\nu}/\kappa$ for Mat\'ern); RHFF (HyLa) samples directly from a half-normal and involves no truncation.
The discarded tail probability, computed by numerical integration of $p(\lambda)$, is below $4\times10^{-4}$ for all spectra except Mat\'ern $\nu{=}0.5$, where it is $3.0\%$, $5.7\%$, and $12.9\%$ at $Q=2$, $3$, and $5$, respectively.
The $\nu{=}0.5$ variant therefore realizes a slightly lighter-tailed spectrum than the nominal Mat\'ern.
Our claims rest on the ordering of tail weights across the evaluated spectra rather than on the nominal $\nu$, and this ordering is preserved under the truncation, so the local--global trends at $Q=2$ are unaffected.
The weakening of the local trend at $Q=5$ (\S~\ref{App:Significance-Test}), however, may be partly confounded by the larger truncation at that dimension.

\paragraph{Optimization.}
RHFF-LVM and RFF-LVM minimize the MAP objective, i.e., \Cref{Eq:GP-Likelihood} plus a latent prior that regularizes the squared geodesic distance to the origin ($\gamma=0.5$; a Gaussian prior for the Euclidean RFF-LVM).
The latents are initialized near the origin ($\mathbf{Z}\sim\mathrm{Unif}\,[0,10^{-3}]^{N\times Q}$, mean-centered) and optimized jointly with the noise precision $\beta$ (initialized at $10^{3}$, updated in log space) by gradient descent with backtracking line search: the step size starts at $0.01$ and is multiplied by $1.1$ on acceptance (capped at $5\times$ the initial value) and by $0.5$ on rejection; gradients are clipped to norm $10$.
Training runs for at most $500$ epochs with early stopping on the leave-one-out log predictive likelihood (patience $20$); the best-LOO embedding is returned.
All model computations use double precision.

\paragraph{Kernel hyperparameters.}
Spectral shape parameters are fixed throughout: $\kappa=2$, $\nu\in\{0.5,1.5,2.5\}$ for Mat\'ern, $t=1$ for heat, $s=1$ for the HyLa spectrum, and lengthscale $\ell=1$ for the Euclidean RFF; only the latents and $\beta$ are learned.
We use $M=500$ features ($2M=1000$ dimensions), chosen from the plateau of the approximation error in \Cref{Fig:Approximation-Error}.

\paragraph{Metrics.}
\textbf{glo} is the Pearson correlation between all $N(N-1)/2$ pairwise distances in the observed and embedding spaces, computed exactly in double precision by blockwise streaming (block size $8192$); no pair subsampling is used.
\textbf{loc} is the $k$NN recall ($k=10$), computed exactly over all $N$ points using the ambient neighbors of the preprocessed data and the geodesic (hyperbolic) or Euclidean neighbors of the embedding. 
Mean ranks average per-dataset ranks (ties receive average rank) over the methods available on all datasets; the combined rank averages the \textbf{loc} and \textbf{glo} ranks per dataset before averaging over datasets.

\paragraph{Baselines.}
PCA (scikit-learn); UMAP (\texttt{n\_neighbors}=15, \texttt{min\_dist}=0.1, $500$ epochs); RFF-LVM (identical engine to RHFF-LVM, Euclidean latent, $\gamma=0.5$); Poincar\'e Map ($k=15$, RiemannianSGD, up to $1{,}000$ epochs with loss-plateau early stopping); BHGP-LVM ($50$ inducing points, $Q=2$ only, $500$ iterations); BGP-LVM (GPy, $50$ inducing points, $500$ iterations); Ht-SNE (Barnes--Hut, $Q=2$ only, $1{,}000$ iterations).
All stochastic methods share the same five seeds.

\paragraph{Scaling protocol.}
All methods run $100$ iterations; \texttt{fit\_time} is wall-clock time including basis generation, with CUDA synchronization; peak memory is the maximum of host-RSS growth and GPU pool bytes.
The time budget equals the largest fit time of RHFF-LVM; a method is stopped when it exceeds this budget or its projected host memory exceeds the limit.
Experiments run on a single NVIDIA RTX A6000 (48\,GB).

\section{Statistical Protocol and Detailed Results}
\label{App:Significance-Test}
In this section, we run statistical tests to validate the main claims of \S~\ref{Section:Experiments}.
We focus on three comparisons, all against RHFF-LVM (heat) on the global metric (\textbf{glo}): PCA (the strongest baseline), RFF-LVM (a geometry ablation), and RHFF-LVM (HyLa), an ablation of the Plancherel weighting $|c(\lambda)|^{-2}$.
We further test the monotonicity of the local--global (\textbf{loc}--\textbf{glo}) trade-off as the smoothness $\nu$ varies.

\paragraph{Protocol.}
We compare methods across the eight real datasets, treating each dataset as one paired observation (the mean over five seeds) and analyzing \textbf{loc} ($k$NN recall with $k=10$) and \textbf{glo} (Pearson correlation between observed and embedded pairwise distances) separately.
Within each dataset, we use a paired $t$-test over the five seeds.
Across datasets, we use the Wilcoxon signed-rank test and report Cohen's $d_z$ as the effect size.
For the ordered (monotone) spectrum hypothesis, we use Page's trend test.
All numbers are at $Q=2$ unless stated otherwise and are summarized per dataset in \Cref{Tab:Key-Results}.

\paragraph{Against PCA.}
On the global metric, the per-dataset difference is significant within every dataset, but its sign is dataset-dependent.
RHFF-LVM (heat) wins on the strongly tree-structured atlases by a large margin (Paul $+36$, Plass $+24$, \emph{C.\,elegans} $+8$) and loses by a small one on the less hierarchical datasets ($\le 7.4$, the non-hierarchical MNIST being the largest).
The wins and losses therefore cancel in the across-dataset test ($3$--$5$ in PCA's favor, $p=0.74$, $d_z=0.45$).

\paragraph{Geometry (vs.\ RFF-LVM).}
Fixing the model and replacing only the Euclidean latent space with the hyperbolic one improves global fidelity on $7/8$ datasets, the sole exception being Paul.
The effect is consistent in direction and large in magnitude ($d_z=0.90$) but does not reach significance at $n=8$ (Wilcoxon $p=0.11$).

\paragraph{Frequency weighting (vs.\ HyLa).}
RHFF-LVM (HyLa) replaces the Plancherel weight $|c(\lambda)|^{-2}$ with a plain half-Gaussian over the frequency, so the comparison isolates the effect of the Plancherel weighting.
At $Q=2$ the weighting improves global fidelity (RHFF-LVM (heat) above RHFF-LVM (HyLa) on $7/8$ datasets, $p=0.02$), while RHFF-LVM (HyLa) attains a consistently, if slightly, higher \textbf{loc} ($8/8$, $p<0.01$).
As discussed in the main text, this reflects the frequency-domain global--local trade-off: both the heat kernel and RHFF-LVM (HyLa), which targets a heat-like spectrum, are geared toward preserving global structure, and doing so correctly requires the Plancherel weight.
Tuning $\nu$ is therefore a more principled way to move along the trade-off than switching to the HyLa spectrum.

\paragraph{Spectrum monotonicity.}
Sweeping the spectrum from $\nu{=}0.5$ through $\nu{=}1.5$ and $\nu{=}2.5$ to the heat kernel moves the two metrics monotonically in opposite directions.
The \textbf{loc} metric decreases (Page trend $p\approx9\times10^{-12}$; strictly monotone on all eight datasets) and \textbf{glo} increases (Page trend $p\approx3\times10^{-4}$; strictly monotone on six datasets, with Paul reversed and the non-hierarchical MNIST non-monotone).
The local trend weakens at $Q=5$ (Page $p\approx0.34$), consistent with the trade-off softening in higher dimensions.
\begin{table}[t]
  \centering
  \caption{\textbf{Significance test} at $Q=2$, per dataset. \emph{Left}: global fidelity (distance correlation, $\times100$) of RHFF-LVM (heat) and its gap $\Delta$ to PCA, to the same-model Euclidean RFF-LVM, and to RHFF-LVM (HyLa) (which drops the $c$-function frequency weight); \textbf{bold} $\Delta$ = significant over the five seeds ($p<0.05$); +\textcolor{green!55!black}{green}/$-$\textcolor{red!70!black}{red} by sign. \emph{Right}: whether each metric is strictly monotone as the spectrum sweeps $\nu{=}0.5\!\to\!$heat ($\uparrow$/$\downarrow$ = monotone in the expected direction, $\times$ = not). Pooled rows: across-dataset wins--losses, Wilcoxon $p$, and \#strictly-monotone datasets (Page's-test $p$).}
  \label{Tab:Key-Results}
  \begin{tabular}{ll|cccc|cc}
    \toprule
    & & \multicolumn{4}{c|}{Global fidelity ($\times100$)} & \multicolumn{2}{c}{Spectrum $\nu0.5\!\to\!$heat} \\
    \cmidrule(lr){3-6} \cmidrule(lr){7-8}
    Dataset & Type & heat & $\Delta_{\text{PCA}}$ & $\Delta_{\text{RFF}}$ & $\Delta_{\text{HyLa}}$ & glo & loc \\
    \midrule
    Paul & tree & 64.2 & \textcolor{green!55!black}{\textbf{+36.3}} & \textcolor{red!70!black}{\textbf{-5.1}} & \textcolor{green!55!black}{+0.1} & \textcolor{red!70!black}{$\times$} & \textcolor{green!55!black}{$\downarrow$} \\
    Plass & tree & 90.1 & \textcolor{green!55!black}{\textbf{+24.2}} & \textcolor{green!55!black}{\textbf{+5.0}} & \textcolor{red!70!black}{-0.5} & \textcolor{green!55!black}{$\uparrow$} & \textcolor{green!55!black}{$\downarrow$} \\
    \emph{C.\,elegans} & tree & 78.8 & \textcolor{green!55!black}{\textbf{+7.8}} & \textcolor{green!55!black}{\textbf{+4.5}} & \textcolor{green!55!black}{\textbf{+1.0}} & \textcolor{green!55!black}{$\uparrow$} & \textcolor{green!55!black}{$\downarrow$} \\
    Zebrafish & tree & 90.9 & \textcolor{red!70!black}{\textbf{-1.4}} & \textcolor{green!55!black}{\textbf{+6.3}} & \textcolor{green!55!black}{\textbf{+1.1}} & \textcolor{green!55!black}{$\uparrow$} & \textcolor{green!55!black}{$\downarrow$} \\
    MouseDev & tree & 91.1 & \textcolor{red!70!black}{\textbf{-3.3}} & \textcolor{green!55!black}{\textbf{+5.0}} & \textcolor{green!55!black}{\textbf{+1.3}} & \textcolor{green!55!black}{$\uparrow$} & \textcolor{green!55!black}{$\downarrow$} \\
    Tasic & taxonomy & 87.9 & \textcolor{red!70!black}{-0.2} & \textcolor{green!55!black}{\textbf{+14.2}} & \textcolor{green!55!black}{\textbf{+2.7}} & \textcolor{green!55!black}{$\uparrow$} & \textcolor{green!55!black}{$\downarrow$} \\
    Zeisel & taxonomy & 76.4 & \textcolor{red!70!black}{\textbf{-0.8}} & \textcolor{green!55!black}{\textbf{+3.9}} & \textcolor{green!55!black}{\textbf{+1.3}} & \textcolor{green!55!black}{$\uparrow$} & \textcolor{green!55!black}{$\downarrow$} \\
    MNIST & non-hier. & 45.3 & \textcolor{red!70!black}{\textbf{-7.4}} & \textcolor{green!55!black}{\textbf{+3.7}} & \textcolor{green!55!black}{+1.1} & \textcolor{red!70!black}{$\times$} & \textcolor{green!55!black}{$\downarrow$} \\
    \addlinespace[1pt]
    \multicolumn{2}{l|}{Pooled: W--L / \#mono} & & 3--5 & 7--1 & 7--1 & 6/8 & 8/8 \\
    \multicolumn{2}{l|}{$p$ (Wilcoxon / Page)} & & 0.74 & 0.11 & 0.02 & $3\times10^{-4}$ & $9\times10^{-12}$ \\
    \bottomrule
  \end{tabular}
\end{table}

\section{Additional Results}
\label{App:Additional-Results}
\begin{sidewaystable*}[p]
  \centering
  \caption{Local and global embedding quality for every latent dimension and their average. \textbf{loc} denotes $k$NN recall, and \textbf{glo} denotes the Pearson correlation between all high-dimensional and embedding pairwise distances. Values are $\times100$. The last three columns of each block give the dataset-averaged mean rank on the \textbf{loc} / \textbf{glo} / both metrics (lower is better). \textbf{bold} = best per column within a block, \colorbox{lightblue}{shaded} = best RHFF-LVM variant.}
  \label{Tab:Real-Data-All}
  \resizebox{\textwidth}{!}{%
  \begin{tabular}{l|cc|cc|cc|cc|cc|cc|cc|cc||ccc}
    \toprule
    & \multicolumn{2}{c|}{\shortstack{Paul\\{\scriptsize $N{=}2.7k$}}} & \multicolumn{2}{c|}{\shortstack{Plass\\{\scriptsize $N{=}21.6k$}}} & \multicolumn{2}{c|}{\shortstack{Tasic\\{\scriptsize $N{=}23.8k$}}} & \multicolumn{2}{c|}{\shortstack{\emph{C.\,elegans}\\{\scriptsize $N{=}54.6k$}}} & \multicolumn{2}{c|}{\shortstack{MNIST\\{\scriptsize $N{=}60.0k$}}} & \multicolumn{2}{c|}{\shortstack{Zebrafish\\{\scriptsize $N{=}71.2k$}}} & \multicolumn{2}{c|}{\shortstack{Zeisel\\{\scriptsize $N{=}160.8k$}}} & \multicolumn{2}{c|}{\shortstack{MouseDev\\{\scriptsize $N{=}292.5k$}}} & \multicolumn{3}{c}{Mean rank $\downarrow$} \\
    \cmidrule(lr){2-3} \cmidrule(lr){4-5} \cmidrule(lr){6-7} \cmidrule(lr){8-9} \cmidrule(lr){10-11} \cmidrule(lr){12-13} \cmidrule(lr){14-15} \cmidrule(lr){16-17} \cmidrule(lr){18-20}
    Method & loc & glo & loc & glo & loc & glo & loc & glo & loc & glo & loc & glo & loc & glo & loc & glo & loc & glo & both \\
    \midrule
    \multicolumn{20}{l}{\textit{Latent dimension $Q=2$}} \\
    \midrule
    PCA & \ms{1.5}{0.0} & \ms{27.9}{0.0} & \ms{1.1}{0.0} & \ms{65.9}{0.0} & \ms{1.1}{0.0} & \textbf{\ms{88.1}{0.0}} & \ms{1.9}{0.0} & \ms{71.0}{0.0} & \ms{0.8}{0.0} & \ms{52.7}{0.0} & \ms{0.9}{0.0} & \textbf{\ms{92.3}{0.0}} & \ms{0.4}{0.0} & \textbf{\ms{77.2}{0.0}} & \ms{0.5}{0.0} & \textbf{\ms{94.5}{0.0}} & 9.00 & 3.35 & 6.18 \\
    UMAP~\cite{mcinnes2018umap} & \ms{5.3}{0.0} & \ms{12.4}{0.2} & \textbf{\ms{13.3}{0.1}} & \ms{67.2}{0.8} & \textbf{\ms{14.3}{0.1}} & \ms{49.1}{2.3} & \textbf{\ms{22.4}{0.1}} & \ms{60.6}{1.1} & \textbf{\ms{11.7}{0.1}} & \ms{35.7}{1.0} & \textbf{\ms{13.2}{0.1}} & \ms{79.3}{1.1} & \textbf{\ms{11.6}{0.1}} & \ms{53.1}{1.4} & \textbf{\ms{8.3}{0.1}} & \ms{63.4}{1.5} & \textbf{1.12} & 7.83 & 4.47 \\
    BGP-LVM~\cite{titsias2010bayesian} & \ms{2.8}{0.1} & \ms{71.8}{0.6} & \ms{3.3}{2.4} & \ms{72.4}{15.2} & \ms{4.8}{1.7} & \ms{73.3}{9.6} & \ms{2.5}{0.2} & \ms{47.8}{0.5} & \ms{0.9}{0.0} & \textbf{\ms{67.0}{0.0}} & \ms{0.2}{0.1} & \ms{10.5}{3.1} & - & - & - & - & - & - & - \\
    RFF-LVM~\cite{gundersen2021latent} & \ms{2.8}{0.0} & \ms{69.3}{0.9} & \ms{9.1}{0.1} & \ms{85.1}{0.4} & \ms{6.7}{0.2} & \ms{73.8}{1.2} & \ms{6.2}{0.1} & \ms{74.4}{1.9} & \ms{3.3}{0.2} & \ms{41.6}{0.9} & \ms{3.0}{0.1} & \ms{84.6}{0.2} & \ms{1.7}{0.0} & \ms{72.5}{1.6} & \ms{0.9}{0.0} & \ms{86.2}{0.7} & 4.30 & 4.93 & 4.61 \\
    \addlinespace[2pt]
    Poincar\'e Map~\cite{klimovskaia2020poincare} & \ms{4.1}{0.1} & \ms{13.5}{0.2} & \ms{10.9}{0.4} & \ms{78.8}{1.6} & \ms{13.9}{0.1} & \ms{47.6}{1.5} & - & - & - & - & - & - & - & - & - & - & - & - & - \\
    BHGP-LVM~\cite{pmlr-v258-watanabe25a} & \ms{2.5}{0.0} & \ms{23.2}{2.7} & \ms{3.0}{0.5} & \ms{73.9}{8.4} & \ms{3.2}{0.9} & \ms{56.9}{15.2} & - & - & - & - & - & - & - & - & - & - & - & - & - \\
    Ht-SNE~\cite{skrodzki2024accelerating} & \textbf{\ms{5.9}{0.2}} & \ms{27.4}{1.1} & \ms{4.7}{0.3} & \ms{17.0}{5.7} & \ms{2.7}{0.3} & \ms{14.7}{6.6} & \ms{6.8}{0.9} & \ms{44.6}{5.5} & \ms{8.1}{1.2} & \ms{21.7}{7.9} & \ms{4.5}{0.4} & \ms{36.5}{6.2} & \ms{2.3}{0.4} & \ms{36.0}{4.6} & \ms{3.6}{0.2} & \ms{21.7}{7.6} & 3.77 & 8.85 & 6.31 \\
    \addlinespace[2pt]
    RHFF-LVM (HyLa)~\cite{yu2023random} & \ms{2.1}{0.0} & \ms{64.1}{0.6} & \ms{6.2}{0.6} & \textbf{\ms{90.6}{1.0}} & \ms{4.6}{0.2} & \ms{85.3}{1.0} & \ms{4.9}{0.2} & \ms{77.9}{0.3} & \ms{2.9}{0.1} & \ms{44.2}{0.8} & \ms{2.5}{0.1} & \ms{89.8}{0.4} & \ms{1.4}{0.0} & \ms{75.1}{0.8} & \ms{0.8}{0.0} & \ms{89.8}{0.6} & 6.62 & 3.33 & 4.97 \\
    RHFF-LVM ($\nu=0.5$) & \cellcolor{lightblue}\ms{2.9}{0.1} & \cellcolor{lightblue}\textbf{\ms{72.0}{0.5}} & \cellcolor{lightblue}\ms{10.1}{0.4} & \ms{85.3}{1.1} & \cellcolor{lightblue}\ms{8.3}{0.4} & \ms{69.9}{3.9} & \cellcolor{lightblue}\ms{9.3}{0.4} & \ms{68.0}{2.1} & \cellcolor{lightblue}\ms{3.8}{0.2} & \ms{45.1}{3.7} & \cellcolor{lightblue}\ms{4.6}{0.2} & \ms{76.8}{1.2} & \cellcolor{lightblue}\ms{3.5}{0.2} & \ms{68.0}{2.1} & \cellcolor{lightblue}\ms{1.7}{0.1} & \ms{69.1}{1.6} & 2.45 & 5.70 & \textbf{4.08} \\
    RHFF-LVM ($\nu=1.5$) & \ms{2.5}{0.2} & \ms{65.7}{0.9} & \ms{7.5}{0.2} & \ms{89.0}{0.4} & \ms{6.0}{0.3} & \ms{78.7}{2.4} & \ms{6.3}{0.4} & \ms{73.9}{2.0} & \ms{3.4}{0.0} & \ms{43.9}{2.0} & \ms{3.2}{0.2} & \ms{81.1}{3.5} & \ms{1.9}{0.1} & \ms{72.6}{1.5} & \ms{1.0}{0.1} & \ms{82.7}{1.3} & 4.15 & 4.73 & 4.44 \\
    RHFF-LVM ($\nu=2.5$) & \ms{2.2}{0.1} & \ms{64.9}{2.1} & \ms{6.4}{0.5} & \ms{89.6}{1.8} & \ms{4.8}{0.3} & \ms{82.0}{1.0} & \ms{5.1}{0.2} & \ms{77.3}{1.2} & \ms{2.9}{0.1} & \ms{44.8}{4.4} & \ms{2.6}{0.1} & \ms{85.6}{5.2} & \ms{1.5}{0.2} & \ms{74.6}{2.5} & \ms{0.9}{0.0} & \ms{88.3}{2.8} & 5.92 & 3.75 & 4.84 \\
    RHFF-LVM (heat) & \ms{2.1}{0.0} & \ms{64.2}{0.4} & \ms{5.6}{0.4} & \cellcolor{lightblue}\ms{90.1}{0.8} & \ms{4.2}{0.2} & \cellcolor{lightblue}\ms{87.9}{0.6} & \ms{4.7}{0.0} & \cellcolor{lightblue}\textbf{\ms{78.8}{0.3}} & \ms{2.8}{0.0} & \cellcolor{lightblue}\ms{45.3}{3.1} & \ms{2.4}{0.0} & \cellcolor{lightblue}\ms{90.9}{0.2} & \ms{1.3}{0.0} & \cellcolor{lightblue}\ms{76.4}{0.3} & \ms{0.8}{0.0} & \cellcolor{lightblue}\ms{91.1}{0.4} & 7.65 & \textbf{2.55} & 5.10 \\
    \midrule
    \multicolumn{20}{l}{\textit{Latent dimension $Q=3$}} \\
    \midrule
    PCA & \ms{2.0}{0.0} & \ms{42.3}{0.0} & \ms{2.2}{0.0} & \ms{84.0}{0.0} & \ms{2.5}{0.0} & \textbf{\ms{90.7}{0.0}} & \ms{6.0}{0.0} & \textbf{\ms{88.4}{0.0}} & \ms{2.6}{0.0} & \ms{64.4}{0.0} & \ms{2.6}{0.0} & \textbf{\ms{95.6}{0.0}} & \ms{1.9}{0.0} & \textbf{\ms{89.9}{0.0}} & \ms{1.3}{0.0} & \textbf{\ms{96.3}{0.0}} & 8.00 & \textbf{2.48} & 5.24 \\
    UMAP~\cite{mcinnes2018umap} & \textbf{\ms{6.6}{0.1}} & \ms{12.1}{0.3} & \textbf{\ms{16.4}{0.1}} & \ms{68.9}{0.3} & \ms{16.6}{0.1} & \ms{46.5}{2.5} & \textbf{\ms{29.0}{0.0}} & \ms{66.4}{0.3} & \textbf{\ms{23.0}{0.1}} & \ms{36.2}{0.2} & \textbf{\ms{20.4}{0.2}} & \ms{80.9}{0.9} & \textbf{\ms{17.2}{0.0}} & \ms{66.1}{1.9} & \textbf{\ms{14.3}{0.0}} & \ms{63.6}{2.5} & \textbf{1.00} & 7.62 & 4.31 \\
    BGP-LVM~\cite{titsias2010bayesian} & \ms{3.4}{0.0} & \ms{74.5}{0.2} & \ms{8.9}{0.2} & \ms{86.5}{0.7} & \ms{6.5}{2.5} & \ms{71.9}{10.6} & \ms{6.9}{1.0} & \ms{55.0}{1.5} & \ms{4.9}{1.0} & \textbf{\ms{67.9}{3.2}} & \ms{1.9}{0.8} & \ms{43.6}{3.2} & - & - & - & - & - & - & - \\
    RFF-LVM~\cite{gundersen2021latent} & \ms{3.3}{0.1} & \ms{77.9}{0.1} & \ms{15.0}{0.1} & \ms{85.6}{0.9} & \ms{12.8}{0.2} & \ms{70.1}{0.8} & \ms{17.1}{0.1} & \ms{81.4}{0.4} & \ms{9.8}{0.3} & \ms{54.3}{0.9} & \ms{8.3}{0.1} & \ms{85.2}{1.4} & \ms{5.7}{0.1} & \ms{78.9}{0.6} & \ms{2.9}{0.0} & \ms{83.5}{0.5} & 3.92 & 4.67 & 4.30 \\
    \addlinespace[2pt]
    Poincar\'e Map~\cite{klimovskaia2020poincare} & \ms{5.2}{0.1} & \ms{14.3}{0.0} & \ms{15.4}{0.3} & \ms{83.7}{0.9} & \textbf{\ms{19.1}{0.0}} & \ms{42.3}{1.5} & - & - & - & - & - & - & - & - & - & - & - & - & - \\
    BHGP-LVM~\cite{pmlr-v258-watanabe25a} & - & - & - & - & - & - & - & - & - & - & - & - & - & - & - & - & - & - & - \\
    Ht-SNE~\cite{skrodzki2024accelerating} & - & - & - & - & - & - & - & - & - & - & - & - & - & - & - & - & - & - & - \\
    \addlinespace[2pt]
    RHFF-LVM (HyLa)~\cite{yu2023random} & \ms{2.9}{0.1} & \ms{71.8}{0.3} & \ms{12.0}{0.2} & \textbf{\ms{93.4}{0.3}} & \ms{9.4}{0.2} & \ms{80.6}{0.6} & \ms{14.1}{0.3} & \ms{85.5}{0.3} & \ms{9.3}{0.2} & \ms{54.8}{0.9} & \ms{6.6}{0.1} & \ms{89.3}{0.6} & \ms{4.9}{0.1} & \ms{83.5}{0.4} & \ms{2.6}{0.0} & \ms{87.1}{0.5} & 6.72 & 2.72 & 4.72 \\
    RHFF-LVM ($\nu=0.5$) & \cellcolor{lightblue}\ms{3.8}{0.1} & \cellcolor{lightblue}\textbf{\ms{83.8}{0.6}} & \cellcolor{lightblue}\ms{15.5}{0.2} & \ms{87.4}{0.6} & \cellcolor{lightblue}\ms{14.4}{0.6} & \ms{71.9}{0.9} & \cellcolor{lightblue}\ms{21.0}{1.1} & \ms{72.7}{2.6} & \cellcolor{lightblue}\ms{11.7}{0.3} & \ms{53.8}{1.5} & \cellcolor{lightblue}\ms{12.4}{0.4} & \ms{74.7}{2.6} & \cellcolor{lightblue}\ms{9.5}{0.6} & \ms{65.7}{1.4} & \cellcolor{lightblue}\ms{5.3}{0.3} & \ms{59.6}{1.5} & 2.00 & 5.80 & \textbf{3.90} \\
    RHFF-LVM ($\nu=1.5$) & \ms{3.1}{0.1} & \ms{76.9}{1.1} & \ms{14.1}{0.3} & \ms{91.0}{0.1} & \ms{12.6}{0.3} & \ms{71.8}{2.0} & \ms{17.2}{0.6} & \ms{79.5}{1.2} & \ms{10.1}{0.4} & \ms{52.3}{2.1} & \ms{9.5}{0.5} & \ms{79.4}{2.1} & \ms{7.0}{0.2} & \ms{74.9}{2.0} & \ms{3.6}{0.1} & \ms{71.6}{2.7} & 3.48 & 5.38 & 4.42 \\
    RHFF-LVM ($\nu=2.5$) & \ms{3.1}{0.1} & \ms{73.7}{0.5} & \ms{13.2}{0.3} & \ms{91.7}{0.4} & \ms{11.1}{0.1} & \ms{74.9}{1.5} & \ms{15.8}{0.2} & \ms{82.5}{0.8} & \ms{9.9}{0.4} & \ms{52.9}{2.8} & \ms{7.8}{0.2} & \ms{86.6}{1.0} & \ms{5.7}{0.2} & \ms{79.6}{0.8} & \ms{3.0}{0.1} & \ms{82.9}{0.8} & 4.62 & 4.15 & 4.39 \\
    RHFF-LVM (heat) & \ms{3.0}{0.1} & \ms{72.3}{0.4} & \ms{12.3}{0.2} & \cellcolor{lightblue}\ms{93.1}{0.3} & \ms{9.5}{0.1} & \cellcolor{lightblue}\ms{79.7}{0.6} & \ms{14.1}{0.1} & \cellcolor{lightblue}\ms{85.2}{0.2} & \ms{9.5}{0.2} & \cellcolor{lightblue}\ms{54.0}{1.0} & \ms{6.6}{0.1} & \cellcolor{lightblue}\ms{89.3}{0.3} & \ms{4.9}{0.0} & \cellcolor{lightblue}\ms{83.2}{0.3} & \ms{2.6}{0.0} & \cellcolor{lightblue}\ms{86.7}{0.2} & 6.25 & 3.18 & 4.71 \\
    \midrule
    \multicolumn{20}{l}{\textit{Latent dimension $Q=5$}} \\
    \midrule
    PCA & \ms{2.6}{0.0} & \ms{48.1}{0.0} & \ms{4.5}{0.0} & \ms{86.7}{0.0} & \ms{6.2}{0.0} & \textbf{\ms{91.2}{0.0}} & \ms{12.7}{0.0} & \textbf{\ms{93.4}{0.0}} & \ms{10.7}{0.0} & \textbf{\ms{78.0}{0.0}} & \ms{7.0}{0.0} & \textbf{\ms{97.1}{0.0}} & \ms{6.3}{0.0} & \textbf{\ms{93.7}{0.0}} & \ms{4.7}{0.0} & \textbf{\ms{98.1}{0.0}} & 8.00 & \textbf{2.45} & 5.22 \\
    UMAP~\cite{mcinnes2018umap} & \textbf{\ms{8.6}{0.1}} & \ms{11.9}{0.1} & \ms{18.6}{0.1} & \ms{69.5}{0.5} & \ms{17.6}{0.1} & \ms{43.8}{1.7} & \ms{30.7}{0.0} & \ms{68.1}{0.2} & \textbf{\ms{28.7}{0.0}} & \ms{40.9}{0.2} & \ms{24.2}{0.0} & \ms{78.7}{0.3} & \ms{19.5}{0.0} & \ms{70.7}{0.6} & \textbf{\ms{18.5}{0.0}} & \ms{67.3}{0.2} & 3.10 & 6.95 & 5.03 \\
    BGP-LVM~\cite{titsias2010bayesian} & \ms{3.8}{0.0} & \ms{76.2}{0.3} & \ms{14.4}{0.6} & \ms{90.2}{0.3} & \ms{9.3}{0.4} & \ms{75.8}{4.3} & \ms{12.1}{0.8} & \ms{54.6}{0.4} & \ms{18.9}{1.1} & \ms{68.8}{3.1} & \ms{9.5}{0.8} & \ms{53.4}{1.3} & - & - & - & - & - & - & - \\
    RFF-LVM~\cite{gundersen2021latent} & \ms{4.2}{0.2} & \ms{83.1}{0.2} & \ms{19.7}{0.1} & \ms{86.7}{0.4} & \ms{20.4}{0.2} & \ms{65.4}{1.0} & \ms{33.9}{0.3} & \ms{77.3}{0.7} & \ms{25.6}{0.3} & \ms{60.4}{0.5} & \ms{21.9}{0.2} & \ms{80.3}{0.5} & \ms{15.8}{0.1} & \ms{76.7}{0.4} & \ms{9.6}{0.2} & \ms{74.8}{1.2} & 4.03 & 4.72 & 4.38 \\
    \addlinespace[2pt]
    Poincar\'e Map~\cite{klimovskaia2020poincare} & \ms{7.0}{0.2} & \ms{15.5}{0.0} & \ms{19.5}{0.1} & \ms{84.8}{0.5} & \textbf{\ms{25.4}{0.1}} & \ms{45.3}{3.6} & - & - & - & - & - & - & - & - & - & - & - & - & - \\
    BHGP-LVM~\cite{pmlr-v258-watanabe25a} & - & - & - & - & - & - & - & - & - & - & - & - & - & - & - & - & - & - & - \\
    Ht-SNE~\cite{skrodzki2024accelerating} & - & - & - & - & - & - & - & - & - & - & - & - & - & - & - & - & - & - & - \\
    \addlinespace[2pt]
    RHFF-LVM (HyLa)~\cite{yu2023random} & \ms{3.5}{0.1} & \ms{79.7}{0.3} & \ms{18.4}{0.1} & \ms{90.3}{0.4} & \ms{19.2}{0.2} & \ms{68.4}{0.6} & \ms{32.2}{0.2} & \ms{80.8}{0.4} & \ms{26.2}{0.2} & \ms{62.3}{0.0} & \ms{19.5}{0.3} & \ms{82.0}{0.5} & \ms{14.8}{0.1} & \ms{79.6}{0.3} & \ms{9.0}{0.1} & \ms{77.5}{0.8} & 6.14 & 2.70 & 4.42 \\
    RHFF-LVM ($\nu=0.5$) & \cellcolor{lightblue}\ms{5.4}{0.1} & \cellcolor{lightblue}\textbf{\ms{87.9}{0.1}} & \cellcolor{lightblue}\textbf{\ms{20.5}{0.2}} & \ms{88.3}{0.4} & \cellcolor{lightblue}\ms{21.9}{0.2} & \cellcolor{lightblue}\ms{69.0}{1.0} & \cellcolor{lightblue}\textbf{\ms{35.4}{0.3}} & \ms{69.2}{1.2} & \cellcolor{lightblue}\ms{26.7}{0.9} & \ms{59.8}{0.3} & \cellcolor{lightblue}\textbf{\ms{24.3}{0.3}} & \ms{74.7}{1.1} & \cellcolor{lightblue}\textbf{\ms{19.8}{0.1}} & \ms{65.9}{1.3} & \cellcolor{lightblue}\ms{14.2}{0.7} & \ms{54.7}{1.6} & \textbf{1.60} & 5.58 & \textbf{3.59} \\
    RHFF-LVM ($\nu=1.5$) & \ms{4.5}{0.1} & \ms{85.0}{0.3} & \ms{18.4}{0.5} & \ms{90.1}{1.5} & \ms{20.2}{0.6} & \ms{67.6}{2.2} & \ms{31.5}{2.7} & \ms{73.2}{3.0} & \ms{24.4}{1.3} & \ms{60.7}{1.0} & \ms{21.4}{2.0} & \ms{73.6}{2.0} & \ms{17.9}{0.6} & \ms{67.0}{1.5} & \ms{12.5}{0.5} & \ms{59.7}{3.2} & 4.30 & 5.25 & 4.78 \\
    RHFF-LVM ($\nu=2.5$) & \ms{4.0}{0.1} & \ms{83.3}{0.3} & \ms{18.8}{0.1} & \ms{88.4}{0.7} & \ms{20.4}{0.3} & \ms{66.9}{1.3} & \ms{34.4}{0.5} & \ms{73.2}{0.9} & \ms{25.8}{0.5} & \ms{61.1}{0.7} & \ms{23.3}{0.2} & \ms{76.7}{1.4} & \ms{17.8}{0.4} & \ms{69.6}{1.7} & \ms{11.0}{0.5} & \ms{66.9}{2.0} & 3.72 & 4.95 & 4.34 \\
    RHFF-LVM (heat) & \ms{3.5}{0.0} & \ms{80.3}{0.2} & \ms{18.6}{0.1} & \cellcolor{lightblue}\textbf{\ms{90.8}{0.7}} & \ms{20.1}{0.2} & \ms{66.9}{0.7} & \ms{32.9}{0.2} & \cellcolor{lightblue}\ms{79.3}{0.2} & \ms{26.5}{0.2} & \cellcolor{lightblue}\ms{62.0}{0.2} & \ms{20.3}{0.3} & \cellcolor{lightblue}\ms{81.0}{0.2} & \ms{15.3}{0.2} & \cellcolor{lightblue}\ms{78.5}{0.4} & \ms{9.3}{0.1} & \cellcolor{lightblue}\ms{75.6}{1.0} & 5.11 & 3.40 & 4.26 \\
    \midrule
    \multicolumn{20}{l}{\textit{Dimension-average}} \\
    \midrule
    PCA & \ms{2.0}{0.0} & \ms{39.4}{0.0} & \ms{2.6}{0.0} & \ms{78.9}{0.0} & \ms{3.2}{0.0} & \textbf{\ms{90.0}{0.0}} & \ms{6.8}{0.0} & \textbf{\ms{84.3}{0.0}} & \ms{4.7}{0.0} & \ms{65.1}{0.0} & \ms{3.5}{0.0} & \textbf{\ms{95.0}{0.0}} & \ms{2.9}{0.0} & \textbf{\ms{86.9}{0.0}} & \ms{2.1}{0.0} & \textbf{\ms{96.3}{0.0}} & 8.00 & \textbf{2.75} & 5.38 \\
    UMAP~\cite{mcinnes2018umap} & \textbf{\ms{6.8}{0.0}} & \ms{12.2}{0.1} & \textbf{\ms{16.1}{0.0}} & \ms{68.5}{0.2} & \ms{16.2}{0.1} & \ms{46.5}{1.0} & \textbf{\ms{27.4}{0.0}} & \ms{65.0}{0.4} & \textbf{\ms{21.2}{0.0}} & \ms{37.6}{0.3} & \textbf{\ms{19.3}{0.1}} & \ms{79.6}{0.2} & \textbf{\ms{16.1}{0.0}} & \ms{63.3}{0.7} & \textbf{\ms{13.7}{0.0}} & \ms{64.8}{1.1} & \textbf{1.70} & 7.43 & 4.56 \\
    BGP-LVM~\cite{titsias2010bayesian} & \ms{3.4}{0.0} & \ms{74.2}{0.1} & \ms{8.9}{0.8} & \ms{83.0}{4.8} & \ms{6.9}{1.2} & \ms{73.7}{5.2} & \ms{7.2}{0.5} & \ms{52.5}{0.4} & \ms{8.2}{0.6} & \textbf{\ms{67.9}{1.5}} & \ms{3.9}{0.4} & \ms{35.8}{1.5} & - & - & - & - & - & - & - \\
    RFF-LVM~\cite{gundersen2021latent} & \ms{3.4}{0.0} & \ms{76.8}{0.2} & \ms{14.6}{0.0} & \ms{85.8}{0.2} & \ms{13.3}{0.1} & \ms{69.8}{0.4} & \ms{19.1}{0.1} & \ms{77.7}{0.7} & \ms{12.9}{0.1} & \ms{52.1}{0.5} & \ms{11.1}{0.1} & \ms{83.4}{0.4} & \ms{7.7}{0.0} & \ms{76.1}{0.4} & \ms{4.5}{0.1} & \ms{81.5}{0.6} & 3.85 & 4.78 & 4.31 \\
    \addlinespace[2pt]
    Poincar\'eMap~\cite{klimovskaia2020poincare} & \ms{5.4}{0.1} & \ms{14.4}{0.1} & \ms{15.3}{0.2} & \ms{82.5}{0.6} & \textbf{\ms{19.4}{0.1}} & \ms{45.1}{0.6} & - & - & - & - & - & - & - & - & - & - & - & - & - \\
    BHGP-LVM~\cite{pmlr-v258-watanabe25a} & \ms{2.5}{0.0} & \ms{23.2}{2.4} & \ms{3.0}{0.5} & \ms{73.9}{7.5} & \ms{3.2}{0.8} & \ms{56.9}{13.6} & - & - & - & - & - & - & - & - & - & - & - & - & - \\
    Ht-SNE~\cite{skrodzki2024accelerating} & \ms{5.9}{0.2} & \ms{27.4}{0.9} & \ms{4.7}{0.3} & \ms{17.0}{5.1} & \ms{2.7}{0.2} & \ms{14.7}{5.9} & \ms{6.8}{0.8} & \ms{44.6}{4.9} & \ms{8.1}{1.1} & \ms{21.7}{7.1} & \ms{4.5}{0.4} & \ms{36.5}{5.5} & \ms{2.3}{0.4} & \ms{36.0}{4.1} & \ms{3.6}{0.2} & \ms{21.7}{6.8} & - & - & - \\
    RHFF-LVM (HyLa)~\cite{yu2023random} & \ms{2.9}{0.0} & \ms{71.9}{0.2} & \ms{12.2}{0.2} & \textbf{\ms{91.4}{0.3}} & \ms{11.1}{0.1} & \ms{78.1}{0.5} & \ms{17.0}{0.1} & \ms{81.4}{0.2} & \ms{12.8}{0.1} & \ms{53.8}{0.4} & \ms{9.5}{0.1} & \ms{87.0}{0.3} & \ms{7.0}{0.0} & \ms{79.4}{0.4} & \ms{4.1}{0.0} & \ms{84.8}{0.2} & 6.25 & 2.92 & 4.58 \\
    \addlinespace[2pt]
    RHFF-LVM ($\nu=0.5$) & \cellcolor{lightblue}\ms{4.0}{0.1} & \cellcolor{lightblue}\textbf{\ms{81.2}{0.1}} & \cellcolor{lightblue}\ms{15.4}{0.2} & \ms{87.0}{0.6} & \cellcolor{lightblue}\ms{14.9}{0.2} & \ms{70.3}{1.5} & \cellcolor{lightblue}\ms{21.9}{0.4} & \ms{70.0}{1.3} & \cellcolor{lightblue}\ms{14.1}{0.3} & \ms{52.9}{1.4} & \cellcolor{lightblue}\ms{13.8}{0.2} & \ms{75.4}{0.8} & \cellcolor{lightblue}\ms{10.9}{0.2} & \ms{66.5}{1.0} & \cellcolor{lightblue}\ms{7.1}{0.3} & \ms{61.1}{0.7} & 1.87 & 5.69 & \textbf{3.78} \\
    RHFF-LVM ($\nu=1.5$) & \ms{3.4}{0.1} & \ms{75.9}{0.3} & \ms{13.3}{0.2} & \ms{90.0}{0.5} & \ms{12.9}{0.2} & \ms{72.7}{1.5} & \ms{18.3}{0.7} & \ms{75.5}{0.5} & \ms{12.7}{0.5} & \ms{52.3}{1.1} & \ms{11.4}{0.6} & \ms{78.0}{1.8} & \ms{8.9}{0.2} & \ms{71.5}{1.1} & \ms{5.7}{0.2} & \ms{71.3}{0.4} & 3.74 & 5.12 & 4.43 \\
    RHFF-LVM ($\nu=2.5$) & \ms{3.1}{0.0} & \ms{74.0}{0.6} & \ms{12.8}{0.2} & \ms{89.9}{0.5} & \ms{12.1}{0.1} & \ms{74.6}{0.6} & \ms{18.4}{0.1} & \ms{77.7}{0.5} & \ms{12.9}{0.2} & \ms{52.9}{1.8} & \ms{11.3}{0.1} & \ms{83.0}{1.1} & \ms{8.4}{0.2} & \ms{74.6}{0.8} & \ms{4.9}{0.2} & \ms{79.4}{1.4} & 4.51 & 4.28 & 4.40 \\
    RHFF-LVM (heat) & \ms{2.9}{0.0} & \ms{72.2}{0.1} & \ms{12.2}{0.2} & \cellcolor{lightblue}\ms{91.3}{0.3} & \ms{11.3}{0.1} & \cellcolor{lightblue}\ms{78.2}{0.5} & \ms{17.2}{0.1} & \cellcolor{lightblue}\ms{81.1}{0.1} & \ms{12.9}{0.1} & \cellcolor{lightblue}\ms{53.8}{0.8} & \ms{9.7}{0.1} & \cellcolor{lightblue}\ms{87.1}{0.2} & \ms{7.2}{0.1} & \cellcolor{lightblue}\ms{79.3}{0.2} & \ms{4.2}{0.0} & \cellcolor{lightblue}\ms{84.5}{0.3} & 6.09 & 3.04 & 4.56 \\
    \bottomrule
  \end{tabular}%
  }
\end{sidewaystable*}

\begin{figure}[p]
    \centering
    \includegraphics[height=0.93\textheight,keepaspectratio]{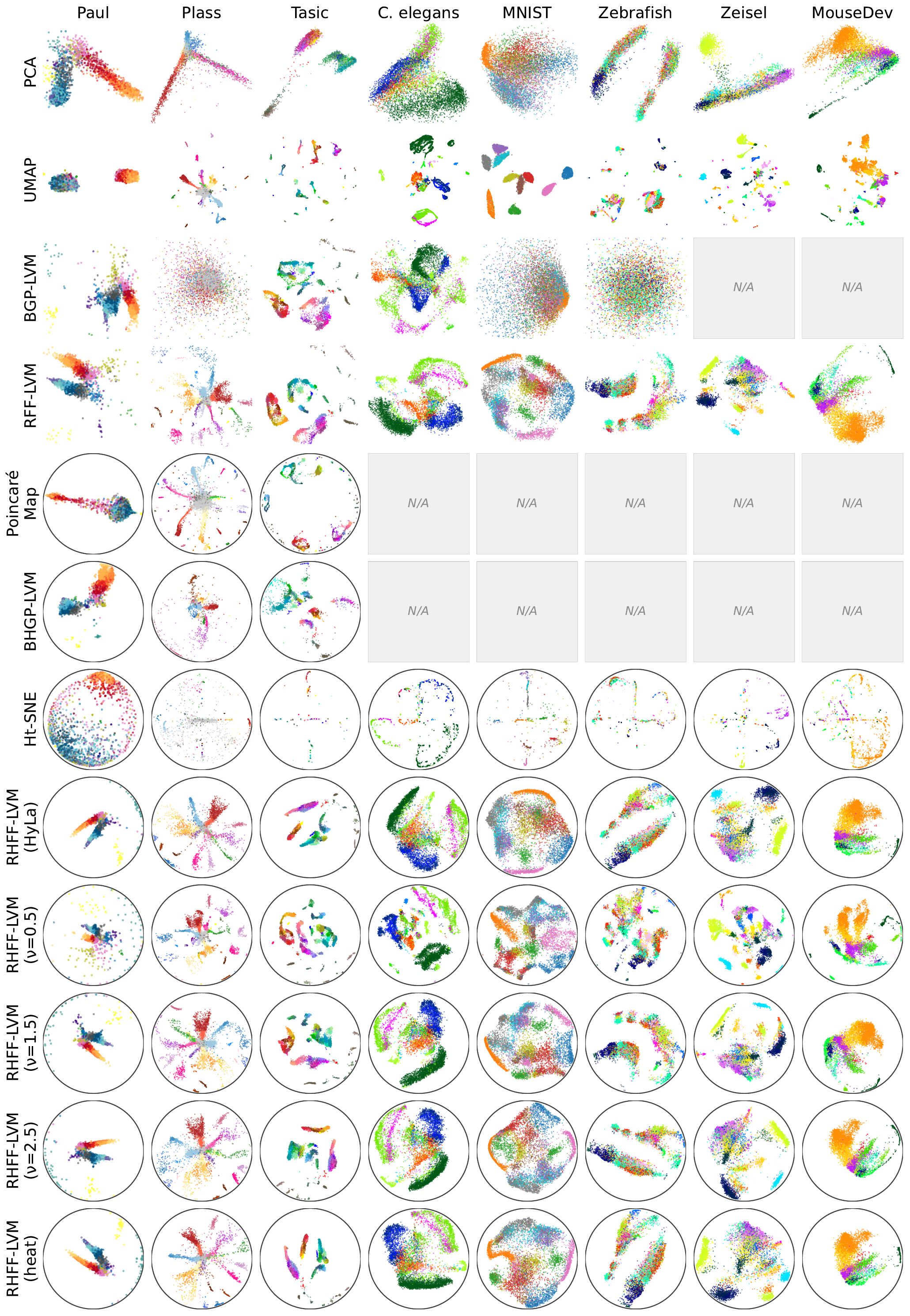}
    \caption{Embedding visualizations at $Q=2$ for all methods (rows) and datasets (columns), colored by cell type. Hyperbolic methods are shown on the Poincar\'e ball, Euclidean methods directly; N/A marks method--dataset pairs that exceeded the time or memory budget or are unsupported.}
    \label{Fig:Embedding-Visualization}
\end{figure}

\Cref{Tab:Real-Data-All} reports the full results at $Q\in\{2,3,5\}$ and their dimension-average; \Cref{Fig:Embedding-Visualization} shows the corresponding embeddings.
Ranks within each block are computed over the methods available on all datasets, so they differ from contrasts restricted to fewer methods.

\paragraph{Dimension dependence of the global metric.}
On \textbf{glo}, RHFF-LVM (heat) leads all methods at $Q=2$ ($2.55$ vs.\ $3.35$ for PCA) but cedes the top position to PCA from $Q=3$ onward ($3.18$ vs.\ $2.48$ at $Q=3$; $3.40$ vs.\ $2.45$ at $Q=5$).
This crossover is consistent with the low-dimensional nature of the hyperbolic advantage~\cite{sarkar2011low}: as $Q$ grows, Euclidean space regains enough capacity that the linear optimum becomes hard to beat on raw distance preservation, while the hyperbolic approximation error simultaneously grows with $Q$ (\Cref{Prop:Error-Bound}).
The hyperbolic advantage is thus largest exactly where dimensionality reduction is hardest---at $Q=2$, the regime that motivates this work.

\paragraph{The trade-off softens, and $\nu=0.5$ takes the lead everywhere.}
The local--global tension relaxes with dimension on both ends.
The heat spectrum recovers local structure as $Q$ grows (its local rank improving from $7.65$ at $Q=2$ to $5.11$ at $Q=5$), and UMAP's local dominance fades ($1.12 \to 1.00 \to 3.10$), overtaken at $Q=5$ by RHFF-LVM ($\nu=0.5$) ($1.60$), which attains the best raw \textbf{loc} on four of the eight datasets.
As a result, RHFF-LVM ($\nu=0.5$) holds the best combined rank at every dimension ($4.08$, $3.90$, and $3.59$ at $Q=2,3,5$; $3.78$ on the dimension-average), and its combined rank improves monotonically with $Q$.
The heavy-tailed spectrum thus offers the most robust default when both criteria matter, while the heat spectrum remains the choice when global fidelity alone is the target.

\paragraph{Where each baseline fails.}
The full table also clarifies the failure modes summarized in the main text.
PCA is bimodal: first or near-first on \textbf{glo} at every dimension, last on \textbf{loc} at every dimension ($8.00$--$9.00$).
UMAP mirrors this at $Q=2$--$3$ but loses its local edge at $Q=5$, where its \textbf{glo} remains poor ($6.95$).
BGP-LVM and the hyperbolic baselines are truncated by scale: BGP-LVM covers six datasets at best, Poincar\'e Map three, and BHGP-LVM and Ht-SNE only the $Q=2$ setting, which is why they are excluded from the rank comparison.
Among RHFF-LVM variants, the interpolation is orderly at every dimension: $\nu=1.5$ and $\nu=2.5$ sit between the two extremes on both metrics, confirming that the spectrum acts as a single, monotone control knob independently of $Q$.

\thispagestyle{empty}


\end{document}